\documentclass[twoside,11pt]{article}

\usepackage{amsmath, amsfonts}
\usepackage{amsthm}
\usepackage{mathrsfs}

\usepackage[abbrvbib, preprint]{jmlr2e}

\usepackage[utf8]{inputenc}
\usepackage[T1]{fontenc}

\usepackage{booktabs}
\usepackage{multirow}
\usepackage{subcaption}
\usepackage{tikz}
\usetikzlibrary{arrows.meta, bending}
\usepackage{algorithm}
\usepackage{algpseudocode}

\usepackage{microtype}

\usepackage{xcolor}
\usepackage{float}

\usepackage{aliascnt}
\usepackage[nameinlink]{cleveref}
\usepackage[margin=1in, marginparwidth=2cm, marginparsep=0.3cm]{geometry}
\usepackage[textwidth=2cm, textsize=footnotesize]{todonotes} 

\usepackage{lastpage}
\jmlrheading{23}{2026}{1-\pageref{LastPage}}{1/21; Revised 5/22}{9/22}{21-0000}{Ryan Liu, Chuwei Wang, Ricardo Baptista, Andrew Stuart and Anima Anandkumar}

\ShortHeadings{WaSP: Wasserstein Data Synthesis Pipeline}{Liu, Wang, Baptista, Stuart, and Anandkumar}
\firstpageno{1}

\newcommand{\cn}{\mathcal{N}}
\newcommand{\cg}{\mathcal{G}}
\newcommand{\ch}{\mathcal{H}}
\newcommand{\cu}{\mathcal{U}}
\newcommand{\cv}{\mathcal{V}}
\newcommand{\cl}{\mathcal{L}}
\newcommand{\cb}{\mathcal{B}}
\newcommand{\ci}{\mathcal{I}}
\newcommand{\ct}{\mathcal{T}}
\newcommand{\eee}{\mathop{{\mathbb{E}}}\nolimits}
\newcommand{\spp}{\mathscr{P}}
\newcommand{\id}{\mathrm{Id}}
\newcommand{\T}{\mathbb{T}}
\newcommand{\Z}{\mathbb{Z}}
\newcommand{\R}{\mathcal{R}}
\newcommand{\norm}[1]{\left\|#1\right\|}
\newcommand{\jp}[1]{\langle #1\rangle}
\newcommand{\ltH}{L^2_t H^\alpha_x}
\newcommand{\kap}{\kappa_\nu}

\definecolor{eqblue}{RGB}{20,90,235}
\definecolor{solred}{RGB}{225,30,30}
\definecolor{cRed}{RGB}{228, 59, 44}
\definecolor{cBlue}{RGB}{59, 130, 246}
\definecolor{cOrange}{RGB}{245, 142, 46}
\definecolor{cGreen}{RGB}{116, 196, 84}

\makeatletter
\newcommand{\jmlrretheorem}[3]{%
  \expandafter\let\csname #1\endcsname\relax
  \expandafter\let\csname end#1\endcsname\relax
  \expandafter\let\csname c@#1\endcsname\relax
  \expandafter\let\csname the#1\endcsname\relax
  \expandafter\let\csname p@#1\endcsname\relax
  \newaliascnt{#1}{theorem}%
  \newtheorem{#1}[#1]{#2}%
  \aliascntresetthe{#1}%
  \Crefname{#1}{#2}{#3}%
}
\makeatother

\jmlrretheorem{lemma}{Lemma}{Lemmas}
\jmlrretheorem{proposition}{Proposition}{Propositions}
\jmlrretheorem{corollary}{Corollary}{Corollaries}
\jmlrretheorem{definition}{Definition}{Definitions}
\jmlrretheorem{remark}{Remark}{Remarks}
\jmlrretheorem{conjecture}{Conjecture}{Conjectures}
\jmlrretheorem{axiom}{Axiom}{Axioms}
\jmlrretheorem{example}{Example}{Examples}
\jmlrretheorem{assumption}{Assumption}{Assumptions}

\title{WaSP: Wasserstein Data Synthesis Pipeline for\\Solver-Free Operator Learning}

\author{%
\name Ryan Liu\textsuperscript{*} \email ryanliu@caltech.edu\\
\addr California Institute of Technology\\
Pasadena, CA, USA
\AND
\name Chuwei Wang\textsuperscript{*} \email chuweiw@caltech.edu\\
\addr California Institute of Technology\\
Pasadena, CA, USA
\AND
\name Ricardo Baptista \email r.baptista@utoronto.ca\\
\addr University of Toronto\\
Toronto, ON, Canada
\AND
\name Andrew Stuart \email astuart@caltech.edu\\
\addr California Institute of Technology\\
Pasadena, CA, USA
\AND
\name Anima Anandkumar \email anima@caltech.edu\\
\addr California Institute of Technology\\
Pasadena, CA, USA
}

\editor{My editor}

\date{}

\begin{document}
\maketitle

\renewcommand{\thefootnote}{\fnsymbol{footnote}}
\footnotetext[1]{Equal contribution.}
\renewcommand{\thefootnote}{\arabic{footnote}}

\begin{abstract}%
Neural operators offer fast surrogates for solving PDEs, but training them typically requires expensive input--solution pairs. Two solver-free alternatives include physics-informed learning and training on manufactured data. Physics-informed learning approximates solutions by minimizing PDE residuals. While the residual norm must be strong enough to control solution error, stronger norms increase mini-batch variance and hinder optimization. Manufactured data constructs input--solution pairs for learning by applying the PDE operator to artificial solutions. However, the resulting training distribution can be mutually singular with the target distribution, yielding poor accuracy on the target distribution. We introduce the Wasserstein Data Synthesis Pipeline (WaSP), which trains on manufactured input--solution pairs while gradually mitigating the distribution shift. WaSP synthesizes new pairs by applying the learned operator to a path of intermediate distributions converging to the target input distribution. Training proceeds in stages along this path. At each stage, the previous surrogate generates solutions, and the exact PDE operator derives corresponding inputs, yielding training pairs for the next surrogate. Theoretically, we prove a stepwise error-propagation bound for the distribution shift introduced during synthesis. Empirically, WaSP reduces solution error by $3.9\times$ to $9.5\times$ compared to existing solver-free methods on PDEs such as Allen--Cahn, Kuramoto--Sivashinsky, and Ginzburg--Landau, and better reproduces the two chaotic systems' statistical observables.

\end{abstract}    


\section{Introduction}

Operator learning constructs mappings between function spaces, such as solution operators that map input conditions to the corresponding solutions of partial differential equations (PDEs). Neural operators parametrize these maps and, once trained, replace numerical solvers at inference time \citep{li2021fourier, lu2021learning, kovachki2023neural}. They are now applied where repeated high-fidelity simulation is expensive or infeasible, including fluid dynamics \citep{2025SpectralRefiner}, weather forecasting \citep{pathak2022fourcastnet}, quantum many-body dynamics \citep{waugh2026toward}, astrophysical accretion and feedback \citep{bhojwani2025black}, and chaotic-system surrogates \citep{wang2024coarse}.

However, the standard training pipeline remains data-intensive. Mainstream methods are trained by supervised regression on input--solution pairs, often obtained from numerical solvers, laboratory measurements, or observational data. Acquiring them at scale can be challenging; simulating a chaotic or multiscale system often requires fine spatial--temporal resolution and expensive implicit solves, and laboratory and observational data are sparse, noisy, and constrained by the measurement setup. Large curated datasets are available for selected benchmark and operational settings \citep{li2008public, hersbach2020era5, takamoto2022pdebench, ohana2024well}, but not for all equations, parameter regimes, geometries, or physical systems. We therefore study \emph{solver-free} operator learning, that is, training a neural operator without any solver-generated solution label for the target inputs.

Physics-informed learning is a mainstream solver-free operator learning paradigm, which replaces solution labels with an equation-derived loss. Physics-informed neural networks (PINNs) instantiate the idea for learning a single PDE instance; it directly parametrizes the solution function with a neural network that maps points in its domain to the corresponding solution values and learns from the PDE residual and boundary constraints \citep{raissi2019physics}. Physics-informed neural operators (PINOs) extend the principle from single PDE instances to operator learning \citep{li2024physics, goswami2023physics}. However, training with physics-informed objectives encounters two obstructions. (1) A small residual loss is not automatically a reliable certificate of solution quality. Its validity depends on whether the residual norm aligns with the stability properties of the PDE \citep{wang20222}. (2) Nonlinear terms and high-order derivatives in the PDE operator enter the computation graph directly, and the resulting composite objective is difficult to optimize \citep{wang2021understanding, krishnapriyan2021characterizing}. We survey the proposed mitigations in \Cref{sec:related-works}.

Another solver-free route reverses the usual data-generation direction. Instead of solving the PDE from inputs to solutions, it samples artificial solutions and applies the PDE operator to obtain the corresponding input conditions\citep{hasani2024generating, sojitra2025method, bischof2026hypino}. Applying the PDE operator is far cheaper and more parallelizable than a numerical solve, so we can efficiently generate large volumes of labeled input--solution pairs. Given the resemblance of this approach to the method of manufactured solutions used to verify numerical solvers \citep{roache2002code}, we refer to these generated pairs as \emph{manufactured data} (MD) in this work.\footnote{We follow the terminology of \citet{sojitra2025method}; \citet{hasani2024generating} call the same construction ``synthetic data,'' which we reserve for data generated by numerical solvers or by our proposed pipeline.} However, since the training pairs are sampled and constructed from the artificial solution prior, their training distribution diverges from the target distribution and provides limited learning signal that is relevant for improving test-time performance.

\paragraph{Our Contribution.} We analyze both routes and develop an algorithm that avoids their obstructions. For physics-informed neural operators, we identify a regularity--variance tradeoff (\Cref{sec:residual-stability,sec:residual-variance}). On the Kuramoto--Sivashinsky equation, we construct a sequence of solutions whose residual and initial-condition errors both vanish in a weak norm while the solution error does not, and we show that even when the norm is strong enough to admit a stability estimate, it carries an amplification constant growing exponentially in the inverse dissipation. On the other hand, while stronger norms restore control, we show with a linear model that such norms amplify the mini-batch variance of the residual learning objective, posing significant challenges for optimizing stability-compatible norms in practice. For manufactured data methods, we prove a complementary obstruction (\Cref{sec:manufactured-shift}). In Gaussian function-space models, the induced input distribution is mutually singular with the target input distribution unless its variance tracks the target's mode by mode. Without further assumptions on the neural operator used, the mutual singularity implies that learning on the exact manufactured pairs provides no performance guarantees at the target distribution. Calibrating the artificial solution prior mode-by-mode to avoid the obstruction, we argue, requires distributional knowledge that is itself as hard as solving the equation.

\begin{figure}
    \centering
    \includegraphics[width=\textwidth]{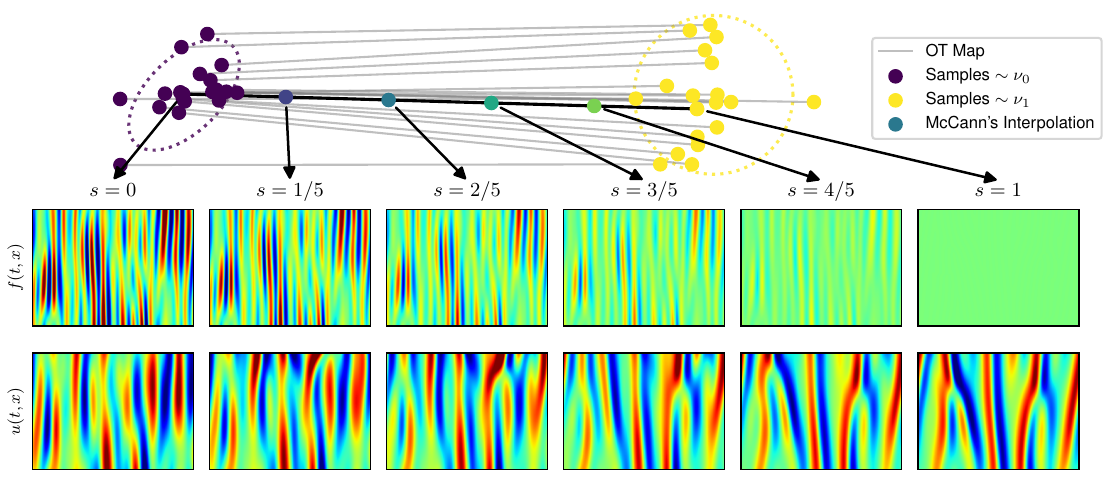}
    \caption{Illustration of how WaSP training distributions are constructed. Top: the McCann interpolation computed for 2D Gaussian point clouds. The gray segments are the estimated optimal transport map $\ct$ between samples drawn from a reference distribution $\nu_0$ and the target $\nu_1$. Each intermediate distribution is defined by the sampling rule in \eqref{eq:mccann-path}, where a sample from the reference is drawn, mapped by $\ct$, and linearly interpolated. Bottom: an input--solution pair for the Kuramoto--Sivashinsky equation interpolated between the manufactured input distribution ($s=0$, left) and the target input distribution ($s=1$, right). The top row is the forcing component of the input, which anneals to near zero as $s$ increases; the bottom row is the corresponding solution field $u(t,x)$, obtained from an ETD-RK4 solver. As the forcing decreases in magnitude and the initial state moves onto the target attractor, the chaotic dynamics emerge, so the interpolation presents the operator with a sequence of progressively harder problems, forming a training curriculum.}
    \label{fig:ks_interpolation}
\end{figure}

This diagnosis motivates the Wasserstein Data Synthesis Pipeline (WaSP), a solver-free training method that preserves the optimization advantage of manufactured-data supervision while mitigating its distribution shift by training along a transport path (\Cref{sec:method}). WaSP samples artificial solutions and applies the PDE operator to obtain the manufactured input distribution, estimates a transport map that maps the manufactured inputs to the target input distribution, and applies McCann interpolation to define a sequence of intermediate training distributions \citep{mccann1997convexity, villani2009optimal}. Training thus becomes a supervised curriculum ordered from easy to hard \citep{bengio2009curriculum}. At each stage, the current operator synthesizes solutions for inputs drawn from the next intermediate distribution, expecting that the neural operator generalizes locally and synthesizes well-approximated solutions. WaSP applies the PDE operator to the solutions and returns corresponding inputs, forming exact input--solution pairs for training at the next stage. The primary learning signal is the solution-space regression loss on the pairs, which avoids the optimization pathologies of physics-informed methods. WaSP breaks the large structural distribution shift between manufactured data and target down into small and manageable local generalization steps; therefore, it also avoids the obstruction identified for manufactured data. \Cref{fig:ks_interpolation} visualizes the transport path for the Kuramoto--Sivashinsky equation, showing how it gradually bridges the distribution shift by defining the intermediate distributions.

We deliver three primary results to support WaSP. First, theoretically, we prove a stepwise error-propagation bound under strong Lipschitzness assumptions (\Cref{thm:wasp-informal}). We illustrate how, without the local generalization assumptions, the error propagation bound \emph{worsens} when the transport path is refined, and with local generalization, the error propagation bound can be exponentially improved and \emph{tightens} instead. Second, implementation-wise, we argue in \Cref{sec:analysis} how, in practice, the strong and unrealistic Lipschitzness assumptions can be removed, which motivates two important designs, the local PDE residual regularization and the adaptive scheduling function. We also discuss efficient implementation of WaSP and methods to improve its numerical stability in \Cref{sec:implementation}. Finally, empirically, with no solver-generated labeled pair from the target distribution used in training, WaSP reduces relative solution error by factors of $3.9$, $9.5$, and $4.6$ relative to the strongest baseline on the Allen--Cahn, Kuramoto--Sivashinsky, and complex Ginzburg--Landau equations. On the two chaotic systems, it also tracks statistical observables more closely, including the spatial and temporal autocorrelations, energy spectrum, and invariant measures. We also conduct two ablations: one removes the self-generated targets while keeping the transport curriculum, and the other adds physics-informed objectives to the manufactured-data training. The ablation results are consistent with the obstructions identified in \Cref{sec:negative-results}.

\paragraph{Paper Organization.}
\Cref{sec:setup} defines the PDE and operator-learning notation, collects the regularity and stability tools used in the analysis, and states the training objectives we compare. \Cref{sec:negative-results} formalizes the two obstructions: the regularity--variance tradeoff for residual training (\Cref{sec:residual-stability,sec:residual-variance}) and the structural distribution shift for manufactured data (\Cref{sec:manufactured-shift}). \Cref{sec:method} introduces WaSP, proves the stepwise error-propagation bound, and shows how the local residual regularizer and the adaptive schedule relax its unrealistic assumptions (\Cref{sec:analysis}), and states the algorithm with its implementation (\Cref{sec:implementation}). \Cref{sec:experiments} describes the three equations, the experiment and evaluation setup, and reports pointwise errors and long-time observables against the physics-informed and manufactured-data baselines. \Cref{sec:discussion} discusses construction of the manufactured solution distribution (\Cref{sec:mu0-construction}) and the cost of the transport estimate (\Cref{sec:ot-cost}). \Cref{sec:related-works} places WaSP in context, and \Cref{sec:conclusion} concludes. Proofs and hyperparameters are in the appendix.

\section{Preliminaries}
\label{sec:setup}
his section first introduces the PDE and operator-learning setting (\Cref{sec:operator-learning}), then defines the regularity and stability notions used in our analysis (\Cref{sec:regularity-stability}), and finally formalizes the training objectives considered in this work (\Cref{sec:training-objectives}).

\subsection{PDE Setting and Notation}
We consider a general PDE setting in which the unknown solution field $u$ is related to the prescribed input $v$ through an equation
\[
    \cn(u)=v .
\]
Here $\cn:\cu\to\cv$ denotes the PDE operator written in a unified form, $\cu$ is a normed space of solution fields, and $\cv$ is a normed space of PDE inputs, which may include forcing terms, boundary data, initial data, or tuples of these quantities. Throughout, we assume $\cu$ and $\cv$ are chosen so that $\cn$ is a bijection whose inverse is the solution operator $\cg:\cv\to\cu$, giving $\cg\circ\cn=\id$ on $\cu$ and $\cn\circ\cg=\id$ on $\cv$; this is an assumption of existence and uniqueness only. Continuous dependence of $u$ on $v$ is a separate property, formalized in \Cref{def:stability}, and it is precisely the property that degrades for the equations we study. \Cref{sec:equations} describes how including the forcing field in the input realizes this assumption for the three benchmarks.

\begin{example}[Evolution problem]
\label{exmp:evolution}
    Consider an evolution problem on the space--time domain $[0,T]\times\Omega$, where $\Omega\subset\mathbb R^d$ is the spatial domain:
    \[
    \begin{cases}
        \mathcal{D}_{t,x}(u)(t,x)=f(t,x), & (t,x)\in(0,T]\times\Omega,\\
        \ci(u)(x)=u_0(x), & x\in\Omega,\\
        \cb(u)(t,x)=g(t,x), & (t,x)\in(0,T]\times\partial\Omega,
    \end{cases}
    \]
    where $\mathcal{D}_{t,x}$ is the space--time PDE operator, $\ci$ is the initial-condition operator, $\cb$ is the boundary operator, and all three may be nonlinear. In the unified notation,
    \[
        \cn(u)
        =
        \bigl(\ci(u),\mathcal{D}_{t,x}(u),\cb(u)\bigr),
        \qquad
        v=(u_0,f,g),
    \]
    so $\cv$ is the product of the initial, interior, and boundary condition spaces. The benchmarks of \Cref{sec:equations} are of this form with periodic boundaries, giving $v=(u_0,f)$.
\end{example}

\subsection{Operator Learning}
\label{sec:operator-learning}
Operator learning seeks a parametric approximation $\cg_\theta:\cv\to\cu$ of the solution operator $\cg$. In realistic applications, the inputs of interest are governed by an underlying distribution over forcing fields, ensembles of initial conditions, or observational states encountered in deployment. We denote this target input distribution by $\nu_1\in\spp(\cv)$, where for a normed space $\mathcal Z$ we write $\spp(\mathcal Z)$ for the set of Borel probability distributions on $\mathcal Z$ with the moments required by the Wasserstein distances used below. Unless otherwise specified, $\nu_1$ denotes the target input distribution and $\nu$ denotes a generic input distribution. The subscript anticipates \Cref{sec:method}, where $\nu_1$ is the endpoint of a transport path that starts at a chosen reference distribution $\nu_0$, such as the input distribution induced by manufactured solutions. For a normed evaluation space $\mathcal X\subseteq\cu$, the target risk of $\cg_\theta$ under a generic input distribution $\nu$ is
\begin{equation}
\label{eq:target-risk}
    \mathcal E(\theta;\nu;\mathcal X)
    :=
    \eee_{v\sim\nu}
    \left[
        \|\cg_\theta(v)-\cg(v)\|_{\mathcal X}^2
    \right].
\end{equation}
Thus, the definition applies to any $\nu$, and evaluating it at $\nu=\nu_1$ gives the target-distribution risk.

A neural operator parametrizes $\cg_\theta$ by composing pointwise lifting and projection maps with a sequence of layers that combine a local linear map with a learnable integral operator on the physical domain \citep{kovachki2023neural}. We use the Fourier Neural Operator \citep{li2021fourier} for $\cg_\theta$ in all of our experiments, which parametrizes the integral operator as a spectral convolution {operator.} The training procedures developed in this work are independent of the architectural choices.

\subsection{Regularity and Stability}
\label{sec:regularity-stability}

Our analysis compares losses and errors measured in norms of different regularity. We therefore recall the relevant Sobolev spaces and formalize what it means for one norm to be stronger than another.

Let $\Omega\subset\mathbb R^d$ be a domain and let $m\in\mathbb N_0$. We write $H^m(\Omega):=W^{m,2}(\Omega)$ for the Sobolev space of functions whose weak derivatives up to order $m$ belong to $L^2(\Omega)$. For a real index $s\in\mathbb R$, the periodic Sobolev space $H^s(\mathbb T^d)$ is defined by
\begin{equation}
\label{eq:sobolev-hs}
    \|f\|_{H^s(\mathbb T^d)}^2
    :=
    \sum_{k\in\mathbb Z^d}
    (1+|k|^2)^s|\widehat f(k)|^2,
\end{equation}
where $\widehat f(k)$ is the $k$th Fourier coefficient of $f$.

\begin{definition}[PDE stability]
\label{def:stability}
    Let $\mathcal A\subseteq\cu$ be an admissible set, let $\mathcal X$ be an evaluation space with $\mathcal A\subseteq\mathcal X$, and let $\mathcal Y$ be a residual space with $\cn(\mathcal A)\subseteq\mathcal Y$. The solution operator $\cg$ is \emph{stable from $\mathcal Y$ to $\mathcal X$ on $\mathcal A$} if there exists a constant $C>0$ such that
    \begin{equation}
    \label{eq:stability-definition}
        \|u-\widetilde u\|_{\mathcal X}
        \leq C\,\|\cn(u)-\cn(\widetilde u)\|_{\mathcal Y}
        \qquad \text{for all } u,\widetilde u\in\mathcal A.
    \end{equation}
    Equivalently, the restriction of $\cg$ to $\cn(\mathcal A)$ is $C$-Lipschitz from $\|\cdot\|_{\mathcal Y}$ to $\|\cdot\|_{\mathcal X}$.
\end{definition}

Stability converts a bound on the PDE residual into a bound on the solution error. If $v\in\cv$ satisfies $\cg(v)\in\mathcal A$, then taking $u=\cg(v)$ in \eqref{eq:stability-definition} and using $\cn\circ\cg=\id$ gives
\begin{equation}
\label{eq:stability-residual}
    \|\cg(v)-\widetilde u\|_{\mathcal X}
    \leq
    C\,\|\cn(\widetilde u)-v\|_{\mathcal Y}
    \qquad\text{for every trial solution }\widetilde u\in\mathcal A.
\end{equation}
Thus, the residual measured in $\mathcal Y$ certifies the solution error measured in $\mathcal X$, with a factor of $C$. Whether such a $C$ exists depends on both norms, which motivates comparing them.

\begin{definition}[Norm comparison]
\label{def:stronger-norm-main}
Let $\|\cdot\|_a$ and $\|\cdot\|_b$ be norms on a common vector space $\mathcal Z$. We say that $\|\cdot\|_b$ is \emph{stronger} than $\|\cdot\|_a$ on $\mathcal Z$ if there exists a constant $C>0$ such that
\begin{equation}
    \label{eq:norm-comparison}
    \|f\|_a
    \leq
    C\|f\|_b
    \qquad
    \text{for every }f\in\mathcal Z,
\end{equation}
and \emph{strictly stronger} if no converse inequality holds uniformly on $\mathcal Z$.
\end{definition}

On the Sobolev scale, $H^{s_2}$ is stronger than $H^{s_1}$ when $s_2\geq s_1$, so a stronger norm controls more derivatives and imposes higher regularity. Combining \eqref{eq:norm-comparison} and \eqref{eq:stability-definition} show that stability is preserved when the residual norm is strengthened or the solution norm is weakened and may fail in either of the opposite directions. 

\subsection{Training Neural Operators}
\label{sec:training-objectives}

We now introduce the three training objectives considered in this work. We call a training procedure \emph{solver-free} if it never uses a solver-generated pair $(v,\cg(v))$ as a training label; it may sample manufactured solutions $u\sim\mu$ from a distribution $\mu \in \spp(\mathcal{U})$, evaluate the PDE operator $\cn$, and draw unpaired inputs $v\sim\nu_1$ from the target distribution. The distribution $\mu$ may itself be produced by any procedure other than running a solver, which is detailed in \Cref{sec:mu0-construction}.

\paragraph{Data Loss.}
The supervised data loss minimizes the expected squared error between the predicted and the true solution under an input distribution $\nu$. Writing $\ell_{\mathrm{data}}(w; v; \mathcal X):=\|w - \cg(v)\|^2_{\mathcal X}$ for the per-sample loss at an input $v\in\cv$ and a predicted solution $w\in\cu$, the operator-level loss evaluated on the target distribution $\nu_1$ is
\begin{equation}
    \label{eq:data-loss}
    \cl_{\mathrm{data}}(\theta;\nu_1;\mathcal X)
        := \eee_{v\sim\nu_1}
        \ell_{\mathrm{data}}(\cg_\theta(v); v; \mathcal X)
        =
        \eee_{v\sim\nu_1}
        \left[
            \|\cg_\theta(v)-\cg(v)\|_{\mathcal X}^2
        \right],
\end{equation}
where $\mathcal{X}$ is the normed space used to quantify the error and $\cg(v)$ is obtained from a numerical solver or from experimental measurements. This objective requires labeled pairs $(v,\cg(v))$ with $v\sim\nu_1$, which can be expensive or unavailable for new equations or environments.

\paragraph{PDE Residual Loss.}
Physics-informed training replaces labeled solution pairs with supervision from the governing equation \citep{raissi2019physics}. Specifically, the per-sample residual loss is
\begin{equation}
\label{eq:pinn-residual-loss}
    \ell_{\mathrm{PDE}}(w; v; \mathcal Y)
    :=
    \|\cn(w)-v\|_{\mathcal Y}^2,
\end{equation}
where $v\in\cv$ is the input and $w\in\cu$ is the predicted solution. In the PINN setting, where the learning target is a single PDE instance (i.e., a fixed $v\in\cv$ and a neural-network-parametrized $w$), the infinite-dimensional norm $\|\cdot\|_{\mathcal Y}^2$ is approximated by a Monte-Carlo estimate of the mean-squared residual at a set of collocation points, corresponding to an $L^2$-type norm over the interior residual and, when present, the initial- and boundary-condition mismatches~\citep{chen2021solving}.

In operator learning, physics-informed neural operators extend this idea by applying the residual loss to the predicted solution field $\cg_\theta(v)$ and averaging over inputs $v\sim\nu$ \citep{li2024physics,goswami2023physics}. This gives
\begin{equation}
\label{eq:pino-residual-loss}
    \cl_{\mathrm{PDE}}(\theta;\nu;\mathcal Y)
    :=\eee_{v\sim\nu} \ell_{\mathrm{PDE}}(\cg_\theta(v); v; \mathcal Y)
    =
    \eee_{v\sim\nu}
    \left[
        \|\cn(\cg_\theta(v))-v\|_{\mathcal Y}^2
    \right].
\end{equation}
Thus, unlike the supervised data loss in \eqref{eq:data-loss}, the PDE residual loss does not require access to the labeled solution $\cg(v)$. It requires only samples $v\sim\nu$ and evaluation of the PDE operator $\cn$ on the neural-operator output.

Lastly, the PDE residual loss certifies the target risk for the solution error through PDE stability in \Cref{def:stability}. If $\cg(v)$ and $\cg_\theta(v)$ both lie in $\mathcal A$ almost everywhere, squaring \eqref{eq:stability-residual} with $\widetilde u=\cg_\theta(v)$ and averaging over $v\sim\nu$ yields the following target risk \eqref{eq:target-risk} evaluated at $\nu$:
\begin{equation}
\label{eq:stability-risk-bound}
    \mathcal E(\theta;\nu;\mathcal X)
    \leq
    C^2\,\cl_{\mathrm{PDE}}(\theta;\nu;\mathcal Y).
\end{equation}
Hence, the amplification is set by the stability constant $C$, which can be uninformatively large or infinite in practice, as we show in \Cref{thm:ks-residual-stability} for the Kuramoto--Sivashinsky equation.

\paragraph{Supervised Learning with Manufactured Data (MD).}
An alternative solver-free approach reverses the usual data-generation direction. Instead of solving the PDE from prescribed inputs to obtain solutions, one samples artificial solutions and applies the PDE operator to obtain the corresponding inputs \citep{hasani2024generating}. Applying $\cn$ is typically far less expensive than solving the PDE forward, especially for evolution equations that require sequential time stepping, stability-restricted step sizes, or repeated implicit solves. Let $\mu\in\spp(\cu)$ be a distribution over solution fields. Sampling $u\sim\mu$ and evaluating $v=\cn(u)$ produces an exact labeled pair $(v,u)$ without solving the PDE; the induced distribution of $v$ is the pushforward distribution $\cn_{\#}\mu \in \spp(\cv)$. By using the change-of-variables from $v$ to $u$, the corresponding training objective is given by
\begin{equation}
    \label{eq:manufactured-loss}
        \cl_{\mathrm{MD}}(\theta;\mu;\mathcal X)
        :=
        \cl_{\mathrm{data}}(\theta;\cn_{\#}\mu;\mathcal X)
        =
        \eee_{u\sim\mu}
        \left[
            \|\cg_\theta(\cn(u))-u\|_{\mathcal X}^2
        \right],
\end{equation}
where the last equality uses $\cg\circ\cn=\id$ to remove the dependence on $\cg$. We write $\ell_{\mathrm{MD}}(w; u; \mathcal X):=\|w - u\|^2_{\mathcal X}$ for the per-sample manufactured-data loss, whose second argument is a solution rather than an input. Then,~\eqref{eq:manufactured-loss} also reads as
\begin{equation}
    \label{eq:manufactured-loss-per-sample}
    \cl_{\mathrm{MD}}(\theta;\mu;\mathcal X)
    =
    \eee_{u\sim\mu}\,
    \ell_{\mathrm{MD}}\bigl(\cg_\theta(\cn(u)); u; \mathcal X\bigr).
\end{equation}
We keep $\ell_{\mathrm{MD}}$ distinct from $\ell_{\mathrm{data}}$ to emphasize that it carries no solver dependency as the regression target is the sampled field $u$ itself.

\begin{remark}[Data and PDE losses are complementary constraints]
    The two solver-free objectives constrain complementary inverses of $\cn$. The PDE residual loss \eqref{eq:pino-residual-loss} enforces $\cn\circ\cg_\theta\approx\id$ on $\cv$, an approximate right-inverse constraint. The manufactured-data loss \eqref{eq:manufactured-loss} enforces $\cg_\theta\circ\cn\approx\id$ on $\cu$, an approximate left-inverse constraint.
\end{remark} 

A shared characteristic of both solver-free formulations based on minimizing the PDE residual loss and using manufactured data is that neither controls the target risk~\eqref{eq:target-risk} at the target distribution $\nu_1$. In the following sections, we study the limitations of existing formulations and propose a novel approach to address them. \Cref{sec:negative-results} identifies a failure mode for each solver-free objective. Then, \Cref{sec:method} develops WaSP, which trains on both objectives and avoids their limitations by constructing an interpolation path that transports the training distribution to the target distribution.

\section{Failure Modes of Existing Solver-Free Objectives}
\label{sec:negative-results}

In this section, we formalize the obstructions that the two solver-free objectives introduced in \Cref{sec:training-objectives} face. For physics-informed learning, \Cref{sec:residual-stability,sec:residual-variance} develop a regularity--variance trade-off for the PDE residual objective, and for manufactured data, \Cref{sec:manufactured-shift} shows a structural distribution shift between the manufactured and the target input distribution.

\subsection{Regularity Requirements for the Residual Loss}\label{sec:residual-stability}

Although a physics-informed objective only minimizes the PDE residual, the residual can serve as a reliable proxy for the solution error, provided the residual norm is strong enough to control the solution error in the norm of interest. This control depends on the PDE's stability properties, which do not always hold. We illustrate conditions for stability in the solution with the Kuramoto--Sivashinsky equation, a nonlinear and weakly dissipative evolution equation that is widely used as the canonical example for chaotic PDE; the thresholds presented here are specific to that equation, but the mechanism they expose is common in nonlinear PDEs.

We consider the Kuramoto--Sivashinsky equation on the periodic spatial domain $\mathbb T$ and time domain $[0,T]$, which models flame-front and thin-film instabilities and is a standard testbed for spatiotemporal chaos. The PDE for the solution field $u$ is:
\begin{equation}
\label{eq:ks-main}
    \partial_t u
    +
    \nu \partial_x^4 u
    +
    \partial_x^2 u
    +
    u\partial_x u
    =
    0,
    \qquad
    (t,x)\in[0,T]\times\mathbb T,
\end{equation}
with initial condition
\[
    u(0,x)=u_0(x),
    \qquad x\in\mathbb T.
\]
Here $\nu>0$ is the dissipation parameter,\footnote{We reuse $\nu$ for the dissipation parameter to align with standard notation; as the dissipation parameter is a scalar and the input distribution is a measure, no confusion should arise.}. We state the instability result in Sobolev norms applied to the spatial components, extended to the Bochner norm $L^p(0,T;H^s)$ for space--time functions. We follow the definition of \citet{wang20222} that the equation is $(\alpha,r,\beta)$-stable if a small loss in the PDE residual and initial condition guarantees small solution error.

\begin{theorem}[Residual--solution stability thresholds for Kuramoto--Sivashinsky, informal]
\label{thm:ks-residual-stability}
Fix $\alpha,r,\beta\in\mathbb R$, and let $u$ be a fixed strong solution of the one-dimensional periodic Kuramoto--Sivashinsky equation \eqref{eq:ks-main} with $u\in L^\infty(0,T;H^m(\mathbb T))$ for some $m>\alpha+\tfrac32$. Define
\[
    \mathcal R(w)
    :=
    \partial_t w+\nu\partial_x^4 w+\partial_x^2 w+w\partial_x w ,
\]
for the PDE residual of the trial solution $w$. Measuring the residual in $L^2(0,T;H^\alpha)$, the initial-condition error in $H^r$, and the solution error in $H^\beta$, the following statements hold.

\begin{enumerate}
    \item If $r<\beta$ and
    $\alpha<\min\{\beta+1,{2\beta+1}\}$, there exists a sequence of smooth functions $w_n$ for which
    \[
        \|w_n(0)-u(0)\|_{H^r}
        +\|\mathcal R(w_n)\|_{L^2(0,T;H^\alpha)}
        \longrightarrow0,
    \]
    while $\|w_n-u\|_{L^\infty(0,T;H^\beta)}$ remains bounded away from zero.
    \item If $r<\beta-2$ and
    $\alpha<\min\{\beta-1,{2\beta-3}\}$, the same conclusion holds with the error in $L^2(0,T;H^\beta)$.
    \item With the exact initial condition $w_n(0)=u(0)$, the same conclusion in $L^\infty(0,T;H^\beta)$ holds if  $\alpha<\min\{\beta-2,{2\beta+1}\}$.
    \item With the exact initial condition, the same conclusion in $L^2(0,T;H^\beta)$ holds if $\alpha<\min\{\beta-4,{2\beta-3}\}$.
\end{enumerate}
Moreover, even if the norm indices $(\alpha,r,\beta)$ are chosen so that a stability estimate holds, there still exists a smooth $w$ for which the residual and initial-condition norms are small, but the solution error is amplified by a factor that grows exponentially in $\nu^{-1}$:
\[
\begin{aligned}
\|w-u\|_{L^2(0,T;H^\beta)}
&\gtrsim 
\operatorname{poly}(\nu) \times e^{T/(4\nu)} \big(\|w(0)-u(0)\|_{H^r} + \|\mathcal R(w)\|_{L^2(0,T;H^\alpha)}\big),\\[1ex]
\|w-u\|_{L^\infty(0,T;H^\beta)}
&\gtrsim 
\operatorname{poly}(\nu)\times e^{T/(4\nu)}\big(\|w(0)-u(0)\|_{H^r} + \|\mathcal R(w)\|_{L^2(0,T;H^\alpha)}\big) .
\end{aligned}
\]
\end{theorem}

A small residual loss is therefore not a certificate of solution accuracy. When the loss norm is too weak no uniform stability estimate is available, and when stability holds the amplification constant can be large, so a small solution error is guaranteed only below a residual threshold that shrinks exponentially in $\nu^{-1}$, which stochastic optimization rarely reaches in practice. The formal statement and proof are given in \Cref{app:ks-stability-proof}.

\subsection{Variance Amplification by Residual Regularity}
\label{sec:residual-variance}

\Cref{sec:residual-stability} proves a lower bound on the regularity of the residual norm required to achieve stability. In operator learning, however, increasing that regularity by using a stronger norm for the training loss raises the variance of the empirical loss and gradient estimates used in stochastic optimization. The variance of those estimates is then known to govern training stability and convergence speed \citep{ghadimi2013stochastic}. Training approximates objective \eqref{eq:pino-residual-loss} by an empirical mean over a mini-batch $v_1,\ldots,v_B\overset{\mathrm{i.i.d.}}{\sim}\nu$ of the per-sample loss $\ell(\theta;v):=\ell_{\mathrm{PDE}}(\cg_\theta(v);v;\mathcal Y)$. If $\ell(\theta;\cdot)$ has a finite second moment under $\nu$, the mini-batch loss and gradient estimates have variance $\operatorname{Var}_{v\sim\nu}[\ell(\theta;v)]/B$ and converge at the root-mean-square rate $B^{-1/2}$, so high per-sample variance translates into noisy updates.

\begin{theorem}[Variance amplification by residual regularity, informal]
\label{thm:pde-loss-variance}
Let $\Omega=\mathbb T^d$, and let $\cn=\cl$ be a linear differential operator of order $m>0$, with Fourier multiplier $p_\cl(k)$ defined by $\widehat{(\cl f)}(k)=p_\cl(k)\hat f(k)$ and satisfying $|p_\cl(k)|\asymp|k|^m$. Define
$\ell_{\mathrm{PDE}}^{(a)}(v):=\|\cl(\cg_\theta(v)-\cg(v))\|_{H^a}^2$ and
$\ell_{\mathrm{data}}^{(a)}(v):=\|\cg_\theta(v)-\cg(v)\|_{H^a}^2$.
Suppose the squared Fourier amplitudes of $\cg_\theta(v)-\cg(v)$ are pairwise uncorrelated under $v\sim\nu$. Then, for $\alpha'>\alpha$,
\begin{equation}
    \operatorname{Var}_\nu\!\left[\ell_{\mathrm{PDE}}^{(\alpha')}(v)\right]
    \gg
    \operatorname{Var}_\nu\!\left[\ell_{\mathrm{PDE}}^{(\alpha)}(v)\right]
    \asymp
    \operatorname{Var}_\nu\!\left[\ell_{\mathrm{data}}^{(\alpha+m)}(v)\right]
    \gg
    \operatorname{Var}_\nu\!\left[\ell_{\mathrm{data}}^{(\alpha)}(v)\right],
    \label{eq:pde-loss-variance-chain}
\end{equation}
where $A\asymp B$ means that each quantity is bounded by a constant multiple of the other, while $A\gg B$ means that $A$ is a strictly stronger variance functional, i.e., there exist admissible error distributions for which $A=\infty$ and $B<\infty$, and whenever $A<\infty$, $B$ is bounded by $A$ up to a constant independent of the error distribution.
\end{theorem}

\Cref{app:pino-variance-proof} gives the formal statement and proof. For nonlinear PDEs, the variance structure is more problem-dependent, and we do not claim a universal ordering beyond the linear setting. Nonlinear terms typically add further variability, since the residual depends on products and compositions of the predicted solution and its derivatives. The linear result nevertheless identifies the basic mechanism: increasing the regularity of a residual norm can increase the variance of the mini-batch objective by a factor polynomial in the frequency content of the error. For the formal version of this theorem, we refer to \Cref{app:pino-variance-proof} in the Appendix.

Together with \Cref{sec:residual-stability}, we identify the \emph{regularity--variance tradeoff}: weak residual norms may fail to control the target solution error, whereas stronger norms can make the objective for stochastic optimization noisier and hence slower to converge.

\subsection{Structural Distribution Shift of Manufactured Data}
\label{sec:manufactured-shift}

We now turn to the alternative solver-free strategy of training with manufactured data, that is, to the objective $\cl_{\mathrm{MD}}$ of \eqref{eq:manufactured-loss}. It removes label error and avoids numerical solves, but it replaces the target input distribution\footnote{For notational simplicity, we write $\nu$ rather than $\nu_1$ for the target distribution in this subsection, in which no generic input distribution involves in the argument.} $\nu$ with the manufactured distribution $\cn_\#\mu$. A structural distribution shift therefore accompanies the exact labeled pairs.

\begin{theorem}[Structural distribution shift under manufactured sampling, informal]
\label{thm:manufactured-shift-informal}
Let $\mu\in\spp(\cu)$ be a Gaussian distribution for PDE solutions, let $\nu\in\spp(\cv)$ be the target condition {distribution,} and let $d_H^2(\rho,\rho')$ denote the square of the Hellinger distance between two distributions $\rho,\rho'$. Let $\rho:=\cn_\#\mu$. Let $s_\mu$ and $s_\nu$ be their sharp Sobolev thresholds, so samples belong almost surely to $H^s$ below the threshold and not above it.

\begin{enumerate}
    \item Suppose $\cn=\cl+\cb$, where $\cl$ is elliptic of order $K$ and $\cb$ is a lower-order (possibly nonlinear) perturbation. Then $\cn_\#\mu$ has sharp threshold $s_\mu-K$. If $s_\nu\ne s_\mu-K$, then $\cn_\#\mu$ and $\nu$ are mutually singular, and $d_H(\cn_\#\mu,\nu)=1$.
    \item Suppose $\cn=\cl$ is linear and diagonal in an orthonormal basis $\{\phi_j\}_{j\ge1}$ such that $\cn\phi_j=\ell_j\phi_j$, and the coefficients of $\mu$ and $\nu$ along $\phi_j$ are independent centered Gaussians with variances $\sigma_{\mu,j}^2$ and $\sigma_{\nu,j}^2$. 
    Let $\rho_N$ and $\nu_N$ denote the distribution of the first $N$ coefficients in this basis.
    Define
    \begin{equation}
        \lambda_j
        :=
        \frac{\ell_j^2\sigma_{\mu,j}^2}{\sigma_{\nu,j}^2},
        \qquad
        b_j:=\lambda_j-1,
        \label{eq:manufactured-relative-variance}
    \end{equation}
    and let $d_{H, N}=d_H(\rho_N, \nu_N)$ be the Hellinger distance between $\rho_N$ and $\nu_N$. Then
    \begin{equation}
        d_{H, N}^2
        =
        1-
        \exp\!\left(
            -\frac12\sum_{j=1}^N
            \psi(b_j)
        \right),
        \qquad
        \psi(b):=-\log\!\left(\frac{2\sqrt{1+b}}{2+b}\right)
        =\frac{b^2}{8}+O(|b|^3).
        \label{eq:manufactured-hellinger-informal}
    \end{equation}
    Hence, by \eqref{eq:manufactured-hellinger-informal}, $d_H(\cn_\#\mu,\nu)<1$ if and only if $\sum_{j\ge1}b_j^2<\infty$. In particular, $b_j\asymp j^{-q}$ gives $d_H<1$ exactly when $q>1/2$.
    \item For an arbitrary $\cn$, fix an orthonormal basis $\{\phi_j\}_{j\geq 1}$ and let $\rho_N$ and $\nu_N$ denote the distributions of the first $N$ coefficients of $\rho=\cn_\# \mu$ and $\nu$, respectively. If there is a family of statistics $T_N$ for which
    \begin{equation}
        \frac{
            \sqrt{\operatorname{Var}_{\rho_N}(T_N)}
            +\sqrt{\operatorname{Var}_{\nu_N}(T_N)}
        }{
            |\eee_{\rho_N}T_N-\eee_{\nu_N}T_N|
        }
        \longrightarrow0 \quad \text{as}\quad N\rightarrow\infty,
        \label{eq:nonlinear-separation-informal}
    \end{equation}
    then $d_H(\rho_N,\nu_N)\to1$. Thus, an accumulated discrepancy in a nonlinear modal statistic is sufficient for asymptotic orthogonality.
\end{enumerate}
\end{theorem}

\begin{example}[A nonlinear example]
\label{rem:nonlinear-separation-example} {The third statement in the theorem gives a general criterion for detecting mutual singularity, which applies when regularity-based arguments are inconclusive. We illustrate this with a concrete example in which the two measures are indistinguishable by regularity.} 

{For mean-zero functions on $\T$, let $\{\phi_j\}_{j\ge1}$ be the real Fourier basis, with $k_j$ the wavenumber of $\phi_j$. Consider the stationary Kirchhoff operator
\[
    \cn(u)=-\bigl(1+\norm{\partial_x u}_{L^2}^2\bigr)\,\partial_x^2 u.
\]
Let $\mu$ be the centered Gaussian with $\sigma_{\mu,j}^2=k_j^{-2a}$ for some $a>5/2$, and the target $\nu=(c\cl)_\#\mu$ with $\cl=-\partial_x^2$ and $c>0$. The measures $\cn_\#\mu$ and $\nu$ have the same Sobolev thresholds $s_\mu-2$. However, the operator $\cn(u)=M\,\cl u$ multiplies the
nonlinearity $\cl u$ by the \emph{random} amplitude
$M=1+\norm{\partial_x u}_{L^2}^2$, whereas $\nu$ multiplies it by the \emph{constant} $c$. To detect this difference, we define

\[
    R_N(v)=\frac1N\sum_{j=1}^N\frac{\jp{v,\phi_j}^2}{k_j^4\sigma_{\mu,j}^2},
\]
which averages the squared Fourier coefficients of $v$ after normalizing them
by the variance they would have under $\cl_\#\mu$. The statistic $T_N=\mathbf 1\{|R_N/c^2-1|>N^{-1/4}\}$ satisfies \eqref{eq:nonlinear-separation-informal}, so $d_H(\cn_\#\mu,\nu)=1$ for every $c>0$. That is, $\cn_\#\mu$ and $\nu$ are
mutually singular for every $c$, even though they share the same Sobolev threshold. The complete statements and proofs are given in the Appendix~\ref{apdx_thm3:sec:kirchhoff-example}.}
\end{example}

Mutual singularity means the manufactured and target distributions concentrate on disjoint sets of inputs, and a Hellinger distance close to one means little effective overlap. Without further structural assumptions on the learned operator, optimizing the manufactured-data loss \eqref{eq:manufactured-loss} then places no constraint on the target risk \eqref{eq:target-risk} since exact prediction on manufactured inputs can coexist with arbitrarily poor performance on target inputs.

The second part also exposes why direct calibration is not a simple remedy. Achieving $d_H<1$ requires the relative deviations $b_j$ of \eqref{eq:manufactured-relative-variance} to be square summable, so $\sigma_{\mu,j}^2$ must track $\sigma_{\nu,j}^2/\ell_j^2$ with vanishing aggregate error. For an arbitrary $\cn$, part 3 gives a sufficient condition for mutual singularity. A family of statistics $T_N$ whose means separate faster than their standard deviations implies mutual singularity, as \eqref{eq:nonlinear-separation-informal} requires. Constructing a $\mu$ to either in the linear case, match the variances mode-by-mode, or in the generic case, prevent any family of statistics from exhibiting separation, requires mode-by-mode knowledge of the target distribution and of the action of $\cn$ on solution modes, which is often as hard as solving the PDE. Therefore, calibration alone cannot resolve the difficulty. The formal statement and proof are given in \Cref{app:synthetic-shift-proof}.

\section{WaSP: Wasserstein Data Synthesis Pipeline}
\label{sec:method}

In this section, we introduce WaSP, a solver-free training framework that avoids the obstructions identified in \Cref{sec:negative-results}. WaSP achieves the goal by iteratively applying the trained neural operator to synthesize the solutions on an input distribution that is transported closer to the target distribution, and thereby mitigating the structural distribution shift of manufactured data without facing optimization challenges of physics-informed objective. We organize the section as follows: \Cref{sec:transporting-distribution} introduces the high-level design of WaSP, \Cref{sec:analysis} proceeds to analyze the error propagation of the algorithm and motivate the design choices, and \Cref{sec:implementation} presents the formal algorithm and the implementation details for it.

\subsection{The High-Level Design of WaSP}
\label{sec:transporting-distribution}

Given a manufactured solution distribution $\mu_0\in\spp(\cu)$ that is easy to sample, we apply the PDE operator to obtain the initial condition distribution
\[
    \nu_0:=\cn_{\#}\mu_0.
\]
WaSP estimates a transport map $\ct:\cv\to\cv$ from $\nu_0$ to the target input distribution $\nu_1$, using entropy-regularized optimal transport under the squared-Euclidean cost \citep{cuturi2013sinkhorn, pooladian2021entropic}; \Cref{sec:ot-cost} describes the estimator and its cost.

\begin{figure}
    \centering
    \begin{tikzpicture}[
        point/.style={circle, fill=black, inner sep=2.5pt, outer sep=2pt},
        axis/.style={very thick, -{Latex[length=3mm, width=2.5mm]}},
        arr/.style={thick, -{Latex[length=2.5mm, width=2mm]}},
        dasharr/.style={thick, dashed, -{Latex[length=2.5mm, width=2mm]}},
        greenline/.style={thick, cGreen, {Bar[width=2.5mm, line width=1pt]}-{Bar[width=2.5mm, line width=1pt]}}
      ]

      \begin{scope}
          \draw[axis] (0,0) -- (6.0,0) node[below] {$s$};
          \node[left] at (-0.3, 0) {$\spp(\cv)$};
          \node[left] at (-0.3, 2) {$\spp(\cu)$};
          
          \node[point, label=below:{$\nu_0$}] (n0) at (0.2, 0) {};
          \node[point, label=below:{$\nu_s$}] (ns) at (1.8, 0) {};
          \node[point, label=below:{$\nu_{s'}$}] (nsp) at (3.4, 0) {};
          \node[point, label=below:{$\nu_{s''}$}] (nspp) at (5.0, 0) {};
          
          \node[point, label=above:{$\mu_0$}] (m0) at (0.2, 2) {};
          \node[point, label=above:{$\hat{\mu}_s$}] (mhs) at (1.7, 2.1) {};
          \node[point, label=above:{$\hat{\mu}_{s'}$}] (mhsp) at (3.2, 2.5) {};
          \node[point, label=above:{$\hat{\mu}_{s''}$}] (mhspp) at (4.6, 3.0) {}; 
          
          \draw[arr, cRed] (m0) to[bend right=25] node[left] {$\cn_\#$} (n0);
          
          \draw[arr, cBlue] (m0) -- (0.6, 1.3);
          \draw[arr, cBlue] (n0) -- (0.6, 0.7);
          \node[cBlue] at (1.0, 1.0) {$\cg^0$};
          
          \draw[dasharr, cOrange] (ns) to[bend right=15] (mhs);
          \draw[dasharr, cOrange] (nsp) to[bend right=15] (mhsp);
          \draw[dasharr, cOrange] (nspp) to[bend right=15] (mhspp);
          \node[cOrange, fill=white, inner sep=1pt] at (2.2, 1.0) {$(\cg^0)_\#$};
      \end{scope}

      \begin{scope}[xshift=6.5cm]
          \draw[axis] (0,0) -- (6.0,0) node[below] {$s$};
          
          \node[point, label=below:{$\nu_0$}] (rn0) at (0.2, 0) {};
          \node[point, label=below:{$\nu_s$}] (rns) at (1.8, 0) {};
          \node[point, label=below:{$\nu_{s'}$}] (rnsp) at (3.4, 0) {};
          \node[point, label=below:{$\nu_{s''}$}] (rnspp) at (5.0, 0) {};
          
          \node[point, label=above:{$\mu_0$}] (rm0) at (0.2, 2) {};
          \node[point, label=above:{$\hat{\mu}_s$}] (rmhs) at (1.7, 2.1) {};
          \node[point, label=above:{$\hat{\mu}_{s'}$}] (rmhsp) at (3.2, 2.5) {};
          \node[point, label=above:{$\hat{\mu}_{s''}$}] (rmhspp) at (4.6, 3.0) {};
          
          \node[point, label={[label distance=-1mm]right:{$\hat{\nu}_s$}}] (rnhs) at (1.7, 0.7) {};
          \node[point, label={[label distance=-1mm]right:{$\hat{\nu}_{s'}$}}] (rnhsp) at (3.1, 1.1) {};
          \node[point, label={[label distance=-1mm]right:{$\hat{\nu}_{s''}$}}] (rnhspp) at (4.4, 1.5) {};
          
          \draw[arr, cRed] (rmhs) -- (rnhs);
          \draw[arr, cRed] (rmhsp) -- (rnhsp);
          \draw[arr, cRed] (rmhspp) -- (rnhspp) node[midway, right=2pt] {$\cn_\#$};
          
          \draw[greenline] (rnhs) -- (rns);
          \draw[greenline] (rnhsp) -- (rnsp);
          \draw[greenline] (rnhspp) -- (rnspp) node[midway, right=4pt] {$W(\nu, \hat{\nu})$};
      \end{scope}
    \end{tikzpicture}
    \caption{Intuitive picture for local generalization of neural operator and its implications. Left: the operator $\cg^0$ is trained at $s=0$ on $\mu_0$ with the manufactured-data loss and on $\nu_0=\cn_\#\mu_0$ with the PDE loss, then applied at $\nu_s,\nu_{s'},\nu_{s''}$ farther along the McCann path (dashed orange); the resulting $\hat\mu_s,\hat\mu_{s'},\hat\mu_{s''}$ drift increasingly far from the solution distributions $\cg_\#\nu_s$. Right: mapping each $\hat\mu_s$ back through $\cn_\#$ (red) gives the realized input distribution $\hat\nu_s$, whose Wasserstein gap $W(\nu,\hat\nu)$ to the intended $\nu_s$ (green segments) grows with $s$. WaSP chooses each increment small enough that $\nu_{n+1}$ stays close to $\nu_n$ so that local generalization holds and updates the operator $\cg^{n+1}$ on $\hat\mu_{n+1}$ and $\nu_{n+1}$, repeating until $s=1$.}
    \label{fig:high-level}
\end{figure}
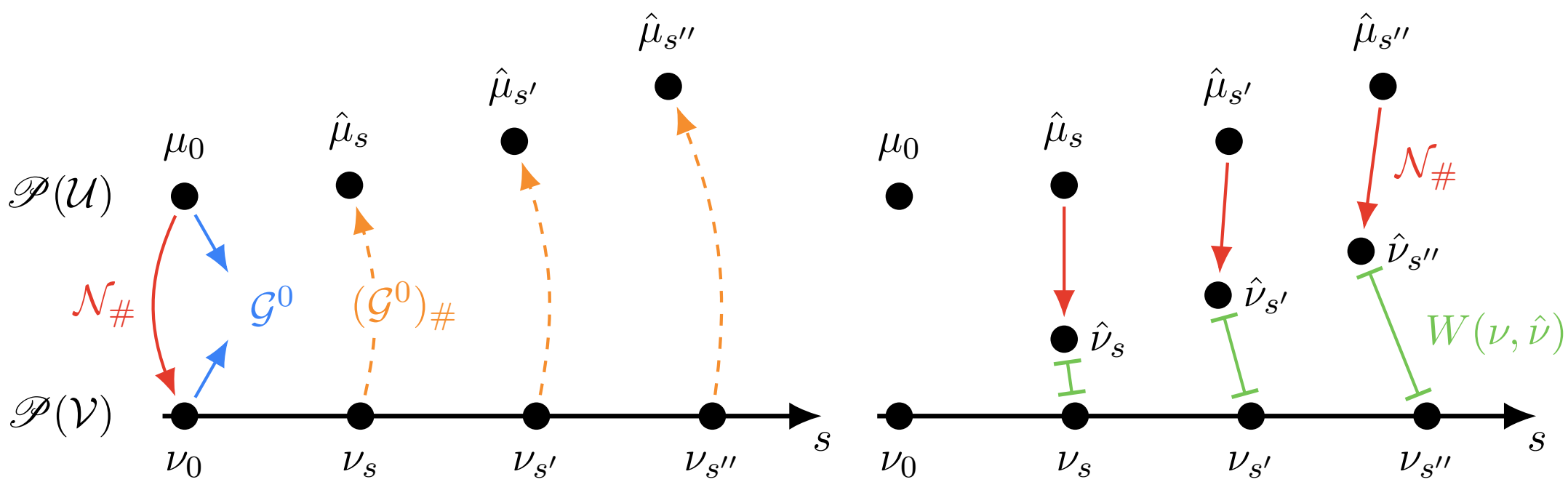

For $s\in[0,1]$, we define the interpolated input distribution
\begin{equation}
\label{eq:mccann-path}
    \nu_s:=((1-s)\id+s\ct)_{\#}\nu_0.
\end{equation}
This McCann interpolation starts at $\nu_0$ and ends at the transported target approximation \citep{mccann1997convexity, villani2009optimal}. We discretize the path by $0=s_0\leq s_1\leq\cdots\leq s_{N}=1$ and write $\nu_n:=\nu_{s_n}$ for convenience.\footnote{The subscript on $\nu$ should be read from context. An integer subscript is a step index, so that $\nu_n:=\nu_{s_n}$ is the $n$th distribution along the discretized path. A real-valued subscript is the interpolation parameter $s\in[0,1]$, so that $\nu_0$ and $\nu_1$ are the path endpoints and $\nu_1$ is the target of \Cref{sec:setup}. To avoid a clash at $s=1$ with the step $n=1$, when presented alone, $\nu_1$ always denotes the target distribution instead of $\nu_{s_1}$} The WaSP algorithm begins by training the neural operator on exact pairs $(\cn(u),u)$ with $u\sim\mu_0$, which corresponds to $s=0$. At a later step $n$ with $s_n>0$, it draws $v\sim\nu_n$ and synthesizes $u=\cg^{n-1}(v)$ using the neural operator trained at the preceding step. Here we write $\theta^n$ for the parameters at step $n$ and $\cg^n:=\cg_{\theta^n}$ for the corresponding neural operator. The resulting solution distribution and its realized input distribution are
\begin{equation}
    \hat\mu_n:=(\cg^{n-1})_\#\nu_n,
    \qquad
    \hat\nu_n:=\cn_\#\hat\mu_n,
    \qquad n\geq1,
    \label{eq:realized-distributions-main}
\end{equation}
with the convention $\hat\mu_0:=\mu_0$ and $\hat\nu_0:=\nu_0$ at the manufactured endpoint, where the pairs are exact and the realized and intended input distributions coincide. Training on $\hat\mu_{n-1}$ remains useful for synthesizing solutions for $\nu_n$ because neural networks generalize locally across nearby input distributions \citep{ben2010theory, benitez2024out}. \Cref{fig:high-level} illustrates the intuition; with a single model $\cg^0$ trained on $\mu_0$, the predicted solution distribution $\hat\mu_s$ and its realized input distribution $\hat\nu_s$ drift further from the intended $\nu_s$ as $s$ grows. Advancing in small transport steps and updating the model at each one therefore reduces the total distribution drift that the process incurs, which is formalized in \Cref{sec:analysis}.

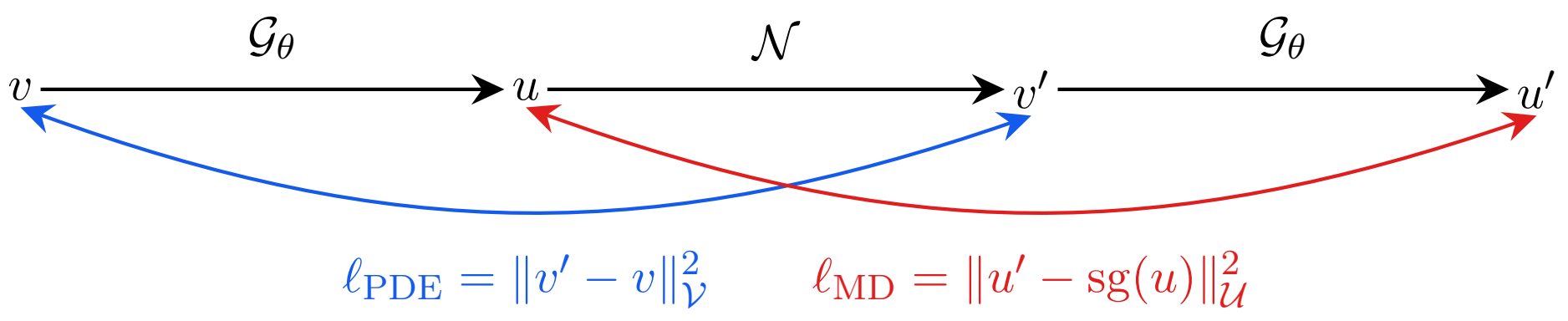
\begin{figure}[t]
    \centering
    \begin{tikzpicture}[
        >={Stealth[length=2.6mm,width=2.4mm]},
        every node/.style={inner sep=1.5pt},
    ]
        \node (v)  at (0,0)   {$v$};
        \node (u)  at (4,0) {$u$};
        \node (vp) at (8,0) {$v'$};
        \node (up) at (12,0) {$u'$};
        \draw[->,thick] (v)  -- node[above=5pt]{$\cg_\theta$} (u);
        \draw[->,thick] (u)  -- node[above=5pt]{$\cn$}        (vp);
        \draw[->,thick] (vp) -- node[above=5pt]{$\cg_\theta$} (up);
        \draw[<->,thick,eqblue] (v.south) to[bend right=20] (vp.south);
        \draw[<->,thick,solred] (u.south) to[bend right=20] (up.south);
        \node[eqblue,anchor=center] at (4,-1.5) {$\ell_{\mathrm{PDE}}=\Vert v'-v\Vert^2_{\cv}$};
        \node[solred,anchor=center] at (8,-1.5) {$\ell_{\mathrm{MD}}=\Vert u' - \mathrm{sg}(u)\Vert^2_{\cu}$};
    \end{tikzpicture}
    \caption{The data synthesis chain of WaSP described in the text. The input loss $\ell_{\mathrm{PDE}}$ (blue) is the residual on the sampled $v$ and reaches $\cg_\theta$ through the first application; the solution loss $\ell_{\mathrm{MD}}$ (red) regresses $u'$ given the realized input $v'=\cn(u)$ onto the synthesized solution $u$ and reaches $\cg_\theta$ only through the second, because $\mathrm{sg}(\cdot)$ is a stop-gradient that holds the synthesized pair fixed rather than making it self-referential.}
    \label{fig:pipeline}
\end{figure}

At each step, the two supervision signals are produced by the synthesis chain shown in \Cref{fig:pipeline}; from a sampled $v\sim\nu_n$, WaSP synthesizes $u=\cg_\theta(v)$, applies $\cn$ to recover the exact input $v'=\cn(u)$, so that $(v',u)$ is an exact pair, and predicts again $u'=\cg_\theta(v')$. The solution loss $\ell_{\mathrm{MD}}$ penalizes $\|\mathrm{sg}(u)-u'\|_{\cu}$ and carries the primary learning signal, regressing $u'$ onto the synthesized solution held fixed as a stop-gradient target; the input loss $\ell_{\mathrm{PDE}}$ penalizes $\|v-v'\|_{\cv}$ and regresses the regenerated input onto the fixed sampled $v$. These are the manufactured-data loss \eqref{eq:manufactured-loss} and the PDE residual loss \eqref{eq:pino-residual-loss} of \Cref{sec:training-objectives}, evaluated locally along the transport path; \Cref{sec:analysis} combines them into the WaSP objective.

\subsection{Theoretical Analysis}
\label{sec:analysis}

The transport path introduces a distinction between the interpolated input distribution $\nu_n$ and the realized distribution $\hat\nu_n$ of \eqref{eq:realized-distributions-main}. We prove a stepwise error propagation bound relating the training error to the target risk, and show that it improves exponentially when local generalization is available.

\paragraph{Stepwise Error Propagation.}
We defer the formal algorithm to \Cref{sec:implementation} to present the analysis that motivates specific designs of WaSP first. We analyze the following setup:
\begin{enumerate}
    \item The whole algorithm takes $N+1$ steps, running from $0$ to $N$, inclusive.
    \item The distributions $\mu_0$ and $\nu_1$ and the grid $s_0, \dots, s_N$ are given and fixed.
    \item At step $n$, the neural operator is $\cg^n$, and the error operator is $\ch^n:=\cg^n-\cg$.
    \item At step $0$, $\hat\mu_0=\mu_0$ and $\hat\nu_0=\nu_0$. At steps $n\geq1$, $\hat\mu_n$ and $\hat\nu_n$ are defined as \eqref{eq:realized-distributions-main}. 
    \item We are given $\epsilon_n:=\eee_{v\sim\hat\nu_n}\|\ch^n(v)\|_{\cu}$ and $e_n:=\eee_{v\sim\nu_n}\|\ch^n(v)\|_{\cu}$ at all $n$.
\end{enumerate}
While $\epsilon_n$ is analogous to the training error and $e_n$ is analogous to the testing error, we make no assumptions about the training process. We treat the operators at each step $\cg^n$ as given and bound $e_n$ by $\epsilon_n$ and other available quantities; whether $\epsilon_n$ is achievable in practice is out of the scope of the analysis.

\begin{assumption}[Operator regularity]
\label{ass:operator-regularity}
$\cn$ is $K_\cn$-Lipschitz, and $\ch^n$ is $K_\ch$-Lipschitz for every $0\leq n\leq N$.
\end{assumption}

\Cref{ass:operator-regularity} is unrealistic: $\cn$ typically differentiates, and \Cref{thm:ks-residual-stability} shows that $\cg$, and hence $\ch^n$, admits infinite or exponentially large Lipschitz constants on the norms of interest. We assume it deliberately, preferring a small set of standard strong hypotheses to the weaker nonstandard substitutes available, which keeps the analysis concise; the two paragraphs that close this subsection remove both constants.

Throughout the analysis, $W:=W_1$ denotes the $1$-Wasserstein distance induced by the norm on $\cv$. For convenience, we define the distance between input distributions:
\[
    d_n:=W(\nu_n,\hat\nu_n),
    \qquad
    \delta_n:=W(\nu_n,\nu_{n+1}),
\]
and show the following worst-case bound:
\begin{corollary}[Worst-case error propagation]
\label{cor:worstcase}
Under \Cref{ass:operator-regularity}, for every $0\leq n<N$,
\begin{equation}
    d_{n+1}
    \leq
    K_\cn K_\ch(d_n+\delta_n)+K_\cn\epsilon_n ,
    \label{eq:worstcase-step}
\end{equation}
and consequently
\begin{equation}
    e_N
    \leq
    \epsilon_N
    +\sum_{k=0}^{N-1}(K_\cn K_\ch)^{N-k}
    \left(K_\ch\delta_k+\epsilon_k\right).
    \label{eq:worstcase-total}
\end{equation}
\end{corollary}

Whenever $K_\cn K_\ch>1$, the factor in \eqref{eq:worstcase-total} is exponential in the number of steps $N$, while refining the path shrinks each $\delta_k$ linearly. Therefore, increasing $N$ \emph{accumulates} more error instead of reducing it, contradicting its purpose. The proof is given in \Cref{app:wasp-step-dependent-proof}. Fortunately, neural networks often generalize across nearby distributions substantially better than the worst-case Lipschitz analysis predicts, which can exponentially improve \Cref{cor:worstcase}

\begin{assumption}[Local distributional generalization]
\label{ass:local-generalization}
There exists $\eta>0$ such that for each step $n$ there exist
$C_n>0$ with
\begin{equation}
\label{eq:local-generalization-assumption}
    \eee_{v\sim\nu'}\|\ch^n(v)\|_{\cu}
    \leq \epsilon_n+C_n W(\nu',\hat\nu_n)^{1+\eta},
\end{equation}
for any input distribution $\nu'\in\spp(\cv)$.
\end{assumption}

Taking $\eta=0$ and $C_n=K_\ch$ recovers the linear transfer bound of \Cref{cor:worstcase}. The content of \Cref{ass:local-generalization} is the improvement near $\hat\nu_n$. Far from $\hat\nu_n$, the term $W(\nu',\hat\nu_n)^{1+\eta}$ need not improve on a linear bound, so the assumption is informative only in the local regime.

\begin{theorem}[WaSP error propagation]
\label{thm:wasp-informal}
Let
\[
    B_n:=2^\eta K_\cn C_n,
    \qquad
    a_n:=K_\cn\epsilon_n+B_n\delta_n^{1+\eta}.
\]
Suppose \Cref{ass:operator-regularity,ass:local-generalization} hold. For any fixed $\alpha\in(1,2)$, define
\[
K:=\min_{0\leq n\leq N} \alpha^nB_n^{-\frac 1 \eta},\quad R_n:= 2^{-\frac 1 \eta}B_n^{-\frac 1 {1+\eta}} (K\alpha^{-n})^{\frac 1 {1+\eta}},
\]
and suppose $\epsilon_n$ and $\delta_n$ are small enough that
\begin{equation}
a_n\leq \left(1 - \frac{\alpha}{2}\right)R_{n+1}\label{eq:smallness}
\end{equation}
for all $0\leq n<N$. Then $d_n\leq R_n$ for every $0\leq n\leq N$, and
\begin{equation}
    e_{N}
    \leq
    \epsilon_{N}
    +2K_\ch
    \max_{0\leq k<N}
    \left[
        K_\cn\epsilon_k
        +2^\eta K_\cn C_k\delta_k^{1+\eta}
    \right].
    \label{eq:wasp-error-propagation-main}
\end{equation}
\end{theorem}

For $\eta>0$, the recursion for $d_n$ contracts inside the trust radius $R_n$ that \eqref{eq:smallness} enforces, so the geometric sum of \eqref{eq:worstcase-total} collapses to a maximum with a single factor $2K_\ch$, and refining the path now reduces the bound instead of inflating it.

Provided the schedule terminates at $s_N=1$, we have $\nu_N=\nu_1$, so $e_N$ is the error of the final operator on the target input distribution, the first-moment analogue of the target risk \eqref{eq:target-risk}. Of the quantities in the bound, only $\epsilon_k$ and $\delta_k$ are quantifiable in practice; $\eta$, $K_\ch$ and $C_k$ are properties of the PDE and of the neural operator parametrization and are not directly measurable, so \Cref{thm:wasp-informal} is an account of which quantities the design controls rather than a sharp performance guarantee.

\begin{remark}[Smallness condition as a step-size constraint]
\label{rmk:schedule-bound}
Write $M:=\eee_{v\sim\nu_0}\|\ct v-v\|_{\cv}$ for the transport cost of the global map $\ct$ used in \Cref{alg:wasp}. Since the interpolation $\nu_s=((1-s)\id+s\ct)_\#\nu_0$ displaces each sample by $(s_{n+1}-s_n)(\ct v-v)$, the induced coupling gives $\delta_n\leq(s_{n+1}-s_n)M$, and \eqref{eq:smallness} is implied by
\begin{equation}
  s_{n+1}-s_n
  \leq
  \frac{1}{M}
  \left[
    \frac{\left(1-\frac{\alpha}{2}\right)R_{n+1}-K_\cn\epsilon_n}{2^\eta K_\cn C_n}
  \right]^{1/(1+\eta)} .
  \label{eq:schedule-bound}
\end{equation}
\end{remark}

\Cref{thm:wasp-informal} uses only the coupling property of $\ct$, so the entropic, PCA-projected map estimated in practice (\Cref{sec:ot-cost}) is still admissible; it pays for its suboptimality through a larger $M$, which tightens the bound in proportion to $1/M$. The trust radius scales as $R_n\propto C_n^{-1/(1+\eta)}$, degrading to $C_n^{-1/\eta}$ at the step attaining the worst constant, so a harder local problem (large $C_n$) or worse generalization (small $\eta$) shrinks both $R_n$ and the admissible increment, and no increment at all is admissible if $K_\cn\epsilon_n\geq(1-\frac{\alpha}{2})R_{n+1}$. Together these motivate a $\mathrm{Schedule}$ function that advances when the local training error is controlled, described in \Cref{sec:implementation}.

In practice, \Cref{ass:operator-regularity} is often not satisfiable by a PDE operator, and the next two paragraphs describe how to relax the assumptions and their practical implications.

\paragraph{The PDE Regularizer Removes $K_\cn$.}
The constant $K_\cn$ enters at one point, the step converting the solution error at step $n$ into the input-space gap at step $n+1$. Coupling $v\sim\nu_{n+1}$ with $\cn(\cg^n(v))\sim\hat\nu_{n+1}$ and using $\cn\circ\cg=\id$ gives
\begin{equation}
    d_{n+1}
    \leq
    \eee_{v\sim\nu_{n+1}}\|\cn(\cg^n(v))-\cn(\cg(v))\|_{\cv}
    \leq
    K_\cn\eee_{v\sim\nu_{n+1}}\|\ch^n(v)\|_{\cu} .
    \label{eq:coupling-with-KN}
\end{equation}
Because $\cn$ typically involves differentiation, closeness in solution space need not imply closeness in input space, so a self-generated $\hat\mu_n$ that looks accurate can still realize an input distribution $\hat\nu_n$ far from $\nu_n$; \Cref{thm:manufactured-shift-informal} treats the same sensitivity. For the Sobolev pairs of interest, $K_\cn$ is either infinite or large enough to make \eqref{eq:coupling-with-KN} uninformative, and a solution-space loss alone is blind to the resulting drift.

We can measure and control the gap instead of bounding it. Applying Cauchy--Schwarz to \eqref{eq:coupling-with-KN} gives
\begin{equation}
    d_{n+1}
    \leq
    \eee_{v\sim\nu_{n+1}}
    \|\cn(\cg^n(v))-v\|_{\cv}
    \leq
    \left(
        \eee_{v\sim\nu_{n+1}}
        \|\cn(\cg^n(v))-v\|_{\cv}^2
    \right)^{1/2},
    \label{eq:realized-gap-residual}
\end{equation}
whose right-hand side takes the same form as \eqref{eq:pino-residual-loss}, suggesting that controlling the PDE residual loss can control the distribution shift $d_{n+1}$. This motivates the stepwise objective:
\begin{equation}
\label{eq:loss_fn}
\begin{aligned}
    \cl_{\mathrm{WaSP}}(\theta;\mu,\nu)
    &:=\alpha_u \cl_{\mathrm{MD}}(\theta;\mu;\cu) + \alpha_v \cl_{\mathrm{PDE}}(\theta;\nu;\cv)
    \\
    &=\alpha_u
    \eee_{u\sim\mu}
    \left[
        \|\cg_\theta(\cn(u))-u\|_{\cu}^2
    \right]
    +
    \alpha_v
    \eee_{v\sim\nu}
    \left[
        \|\cn(\cg_\theta(v))-v\|_{\cv}^2
    \right],
\end{aligned}
\end{equation}
where $\alpha_u,\alpha_v>0$ are the loss weights and $\cu$, $\cv$ are the norms chosen for the manufactured-data loss and the PDE residual loss, respectively. The objective shares a similar functional form with the mixed manufactured-data and physics-informed objective of \citet{bischof2026hypino}, but is motivated differently.

Since \eqref{eq:coupling-with-KN} is the only place where $K_\cn$ enters the analysis, the conclusion of \Cref{thm:wasp-informal} stands without requiring $\cn$ to be Lipschitz, provided the PDE residual loss is controlled. Unlike residual-only training, WaSP never asks the PDE loss to certify the solution error or to reach its global minimum: it acts as a \emph{local regularizer} that keeps $d_n$ small, and can therefore be treated as a threshold for advancing the transport rather than an optimization proxy. WaSP thereby avoids the regularity--variance tradeoff identified in \Cref{sec:residual-stability,sec:residual-variance} since weaker norms suffice for its purpose.

The residual loss also prevents the confirmation bias of self-generated targets. Because WaSP regresses on the self-generated target $\hat\mu_n$ and the inputs it realizes, a degenerate prediction at one step becomes the regression target at the next, inducing a strong confirmation bias that collapses every input to the degenerate outputs \citep{arazo2020pseudo}. For example, an operator that always predicts all zeros in the interior and matches the boundary data would incur a very high MD loss at step $0$, and for $n\geq 1$, the MD loss is trivially optimized. By introducing a model-independent regression target, the neural operator cannot exploit degenerate predictions and thus prevents the training from collapsing.

\paragraph{Pathwise Amplification Removes $K_\ch$.}

The transfer between adjacent distributions exercises $\ch^n$ only along the displacement that a single transport step applies. \Cref{thm4:lem:transport} replaces $K_\ch$ by the \emph{pathwise amplification factor} $\overline K_n$, the displacement-weighted average of the local secant slopes of the current error map along that step, giving the one-step transfer bound
\begin{equation}
\label{eq:pathwise-transfer-bound}
    \eee_{v\sim\nu_{n+1}}\|\ch^n(v)\|_{\cu}
    \leq
    \eee_{v\sim\nu_n}\|\ch^n(v)\|_{\cu}
    +\overline K_n\,\delta_n.
\end{equation}
The product $\overline K_n\delta_n$ is finite whenever the error map has integrable increments along the path, with no Lipschitz hypothesis at all, and since $\overline K_n\leq K_\ch$ the substitution is never worse than the constant it replaces; it applies wherever a Lipschitz constant of $\ch^n$ is used, including the final transfer in \Cref{thm:wasp-informal}.

\subsection{The WaSP Algorithm and Implementation}
In this section, we formally introduce the WaSP in \Cref{alg:wasp} and its implementation, focusing on the schedule function motivated in \Cref{sec:analysis}. We then turn the idealized algorithm into a practical training scheme with two additions, including a replay buffer, which improves computational efficiency and training stability and a dynamic normalization scheme, which keeps the two loss terms balanced as the distributions drift along the path
\label{sec:implementation}
\paragraph{WaSP Algorithm.}
We present \Cref{alg:wasp}, an idealized version of WaSP. WaSP first pushes the manufactured solution distribution through the operator to obtain $\nu_0=\cn_\#\mu_0$ and estimates the transport map $\ct$ from $\nu_0$ to $\nu_1$. Each iteration then selects an interpolation parameter $s$ from the schedule and takes one gradient step on the WaSP loss \eqref{eq:loss_fn}. The conditional distinguishes the warm-start phase from the transport phase. During warm start, where $s=0$, it trains on the manufactured pairs directly sampled from $\mu_0$. When $s>0$, WaSP forms the intermediate distribution $\nu_s$ by McCann interpolation \eqref{eq:mccann-path} and invokes $\cg_\theta$ to synthesize the matching solutions, since the true solution distribution $\cg_\#\nu_s$ is unavailable. For simplicity, we write a plain SGD update with step size $\mathrm{lr}$, but any optimizer may be used instead. We also define the ``terminal optimizer step'' $N$ for notational convenience, which makes the total optimizer steps taken $N+1$ instead of $N$.

In place of a fixed grid $s_0,\dots,s_{N}$, the intermediate parameters $s_n$ are produced by a schedule function $\mathrm{Schedule}(\ell,n)$ where $\ell$ is the training PDE residual loss and $n$ is the optimizer step. The function satisfies $\mathrm{Schedule}(\cdot,0)=0$, $\mathrm{Schedule}(\cdot,n)\leq\mathrm{Schedule}(\cdot,n+1)$, and $\mathrm{Schedule}(\cdot,N)=1$. While \Cref{rmk:schedule-bound} argues the necessity of the schedule function based on the MD loss $\epsilon_n$, we choose to monitor PDE residual loss instead for the following reasons. First, \eqref{eq:schedule-bound} itself relies on unrealistic assumptions and depends on constants including $R_{n+1}$ and $C_n$ that we have no means to estimate. Therefore, monitoring MD loss $\epsilon_n$ is not more justified than PDE residual loss. Second, we can bound the total distribution shift as 
\[
W(\nu_{n+1},\hat\nu_n)\leq W(\nu_{n+1},\nu_n) + W(\nu_{n},\hat\nu_n)=\delta_n+d_n.
\]
By advancing $s$ only when $d_n\leq\cl_{\mathrm{PDE}}(\theta^{n-1};\nu_n, \cv)^{1/2}$ is smaller than a threshold $d_{\mathrm{max}}$, we can guarantee that $W(\nu_{n+1},\hat\nu_n)$ is always smaller than $d_{\mathrm{max}} + M\Delta s$, where $M$ is the transport cost $\eee_{v\sim\nu_0}\|\ct v-v\|_{\cv}$ and $\Delta s$ is the increment in $s$. 

\begin{algorithm}[htbp]
\caption{Wasserstein Data Synthesis Pipeline (WaSP)}
\label{alg:wasp}
\begin{algorithmic}[0]
    \Require Solution distribution $\mu_0$, target input distribution $\nu_1$, operator $\cn$, terminal optimizer step $N$, step size $\mathrm{lr}$, initialization function $\mathrm{Initialize}()$, schedule function $\mathrm{Schedule}(\ell, n)$.
    \State $\nu_0\gets \cn_{\#}\mu_0$
    \State Estimate $\ct$ from $\nu_0$ to $\nu_1$
    \State $\theta\gets \mathrm{Initialize()}$,\quad$s\gets 0$
    \For{$n=0,\ldots,N$}
        \State $s\gets \mathrm{Schedule}(\hat\cl_{\mathrm{PDE}}(\theta; \nu_s; \cv), n)$ \Comment{$\hat\cl$ denotes the Monte Carlo estimate of $\cl$.}
        \If{$s > 0$}
            \State $\nu_s\gets ((1-s)\id+s\ct)_{\#}\nu_0$
            \State $\hat\mu_s\gets (\cg_\theta)_\#\nu_s$
            \State $\theta\gets \theta- \mathrm{lr}\cdot\,\nabla_{\theta}\hat\cl_{\mathrm{WaSP}}(\theta; \hat\mu_s, \nu_s)$
        \Else
            \State $\theta\gets \theta- \mathrm{lr}\cdot\,\nabla_{\theta}\hat\cl_{\mathrm{WaSP}}(\theta; \mu_0, \nu_0)$
        \EndIf
    \EndFor
    \State \Return $\theta$
\end{algorithmic}
\end{algorithm}

\paragraph{Sample Reuse.} The redundancy in evaluating the two loss terms in \Cref{fig:pipeline} motivates us to reuse the samples. The manufactured-data term at step $n$ regresses on solution samples $u\sim\hat\mu_n=(\cg^{n-1})_\#\nu_n$, which appear to require a dedicated forward pass through $\cg_\theta$; however, the PDE term already performs this computation, since evaluating $\cl_{\mathrm{PDE}}$ on a batch $v\sim\nu_n$ produces $\cg_\theta(v)$ together with its exact input $\cn(\cg_\theta(v))$, yielding exact pairs as a by-product. An implementation of \Cref{alg:wasp} with no redundant computation therefore evaluates the PDE loss first (the PDE half-step), caches the forward pass, evaluates the solution loss on the cached pairs (the MD half-step), and takes a single optimizer step on the accumulated gradients. This gains computational efficiency at the expense of correlating the Monte Carlo estimates of the two loss terms in \eqref{eq:loss_fn}.

Our implementation departs from this half-step construction by introducing a \emph{replay buffer} \citep{lin1992self,mnih2015human}. We parallelize the two half-steps by using exact pairs generated at previous steps rather than those from the current half-step, thereby removing the serial dependency between the two evaluations. This allows the two half-steps to be combined into a single larger-batch forward pass, increasing arithmetic intensity and improving hardware utilization. The replay buffer also restores independence between the Monte Carlo estimates of the two terms in the WaSP objective \eqref{eq:loss_fn}, since the PDE loss is evaluated on fresh samples drawn from $\nu_n$ while the MD loss is evaluated on samples drawn from the buffer.

\paragraph{Dynamic Normalization.} Because the input and output statistics change along the path from $\nu_0$ to $\nu_1$, we apply dynamic normalization based on exponential moving averages (EMA) of the current statistics to control the input and output magnitudes of the neural operator; this also sets the relative scaling of $\alpha_u$ and $\alpha_v$ in \eqref{eq:loss_fn} and keeps gradients from both terms balanced as the path advances. Note that the dynamic normalization update must be frozen towards the final stage of training since the fluctuating normalizers create a noise floor that prevents the neural operator from converging to a fixed solution.

\begin{figure}
    \centering
    \begin{subfigure}{0.62\textwidth}
        \centering
        \includegraphics[width=\textwidth]{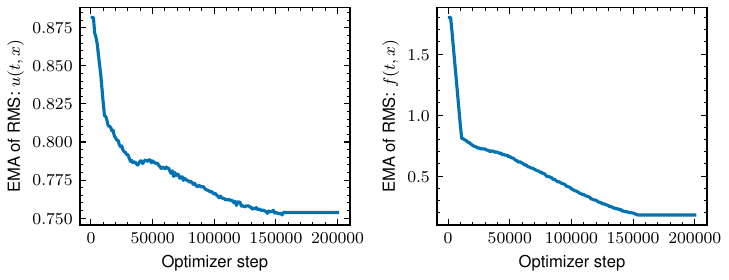}
        \caption{}
        \label{fig:dynamic-normalization}
    \end{subfigure}
    \begin{subfigure}{0.36\textwidth}
        \centering
        \includegraphics[width=\textwidth]{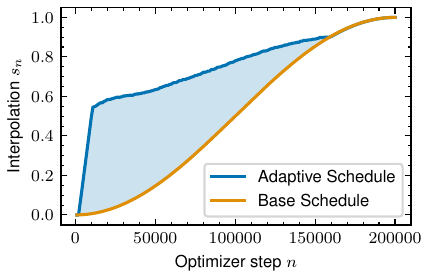}
        \caption{}
        \label{fig:adaptive}
    \end{subfigure}
    \caption{Training dynamics of the practical WaSP algorithm (\Cref{sec:implementation}), plotted against optimizer step, for the complex Ginzburg--Landau experiment of \Cref{sec:equations}, whose input is the pair $v=(u(0,x),f(t,x))$ of initial state and forcing. (a) Dynamic normalization: the EMA of the root-mean-square magnitude of the solution field, the output of $\cg_\theta$ (left), and of the forcing component $f$ of the input (right). (b) Adaptive transport schedule: the interpolation parameter $s$ on the vertical axis. The adaptive schedule advances to the next intermediate distribution only once $\cl_{\mathrm{PDE}}$ has fallen below a threshold, spending more steps on harder portions of the path, whereas the baseline advances on a fixed cosine schedule.
    }
    \label{fig:training-dynamics}
\end{figure}

\Cref{fig:dynamic-normalization} shows these running statistics drifting over the course of training, which is precisely what fixed normalization constants would fail to track. Finally, the EMA statistics also control the adaptive $\mathrm{Schedule}$ function: WaSP changes the value of $s_n$ at step $n$ only when the \emph{normalized} PDE loss $\alpha_v\cl_{\mathrm{PDE}}$ falls below a prescribed threshold, so that harder portions of the path receive smaller effective Wasserstein steps. \Cref{fig:adaptive} contrasts this adaptive schedule with the fixed cosine baseline, showing that it concentrates optimizer steps where the path is most difficult and converges faster overall.

\section{Experiments}
\label{sec:experiments}

We evaluate WaSP under a solver-free training setup: no solver-generated labeled pairs from the target distribution are used during training. We only assume that unpaired inputs from the target input distribution $v\sim\nu_1$ are available, as they are for every physics-informed method. We compare WaSP against two baselines and two ablations on the Allen--Cahn, Kuramoto--Sivashinsky (KS), and complex Ginzburg--Landau (CGLE) equations. The baselines are Physics-Informed Neural Operator (PINO) \citep{li2024physics}, which minimizes the residual loss \eqref{eq:pino-residual-loss} on the target distribution $\nu_1$, and Manufactured Data \citep[MD;][]{hasani2024generating, sojitra2025method}, which minimizes the supervised loss \eqref{eq:manufactured-loss} on manufactured pairs $(v,u)$ drawn from $\mu_0$. Each ablation study extends a baseline by introducing one WaSP component. PINO+OT extends PINO by training on the intermediate input distribution along the transport path, which is WaSP without self-generated data supervision. MD+PINO extends MD by training on the same manufactured distribution $\mu_0$ and introduces physics-informed supervision, which is \eqref{eq:loss_fn} evaluated on $\mu_0$ and $\nu_0$. 

\subsection{Equations}
\label{sec:equations}
We benchmark WaSP on three one-dimensional evolution equations with periodic spatial boundary conditions. To each equation we add a forcing term $f$. In the unified notation of \Cref{sec:setup}, the input is then $v=(u(0,x),f(t,x))$, comprising the initial state and the forcing field, and with the forcing included every sufficiently regular field $u$ is the unique solution of $\cn(u)=v$, which realizes the bijection assumed there. The target  {input} distribution $\nu_1$ corresponds to the unforced equations: $f\equiv 0$, with the initial condition $u(0,x)$ drawn from the distributions specified below. The forcing is therefore active only on the manufactured side of the transport path and anneals to zero as $s\to1$. Allen--Cahn and Kuramoto--Sivashinsky are established operator-learning benchmarks, while the chaotic complex Ginzburg--Landau regime we study is comparatively underexplored. Together the three span a range of difficulty, from relaxation to developed spatio-temporal chaos; \Cref{sec:related-works} places them in the operator-learning literature.

\paragraph{Allen--Cahn.} The Allen--Cahn equation is a reaction--diffusion PDE,
\begin{equation}
    \partial_t u - \nu \partial_x^2 u - u + u^3 = f.
\end{equation}
We take $\nu = 0.01$ on $\Omega=[0,5]_t\times[0,2\pi]_x$, with initial conditions drawn from a Gaussian random field with power spectrum proportional to $e^{-k_x}$, with RMS amplitude of $\sim0.15$. The bistable reaction term drives the field toward the phase-separated states $u=\pm1$, making this the mildest of the three benchmarks.
 
\paragraph{Kuramoto--Sivashinsky.} The Kuramoto--Sivashinsky equation is an evolution PDE that develops spatio-temporal chaos,
\begin{equation}
    \partial_t u + u \partial_x u + \partial_x^2 u + \nu \partial_x^4 u = f.
\end{equation}
We take $\nu = 0.01$ on $\Omega=[0,0.128]_t\times[0,5]_x$, with initial conditions drawn from the invariant measure. Linearizing about the zero state gives the growth rate $k^2-\nu k^4$, positive for $0<k<\nu^{-1/2}=10$ and maximized at $k=(2\nu)^{-1/2}\approx7.07$. On the domain $[0,5]$ the admissible wavenumbers are $k_n=2\pi n/5$, so the lowest seven Fourier modes are linearly unstable, and the sixth is the fastest growing. Rescaling space, time, and amplitude maps the equation to the standard Kuramoto--Sivashinsky equation at domain length $L/\sqrt{\nu}=50$, well above the conventional chaotic threshold $\gtrsim22$ \citep{cvitanovic2010state}.
 
\paragraph{Complex Ginzburg--Landau.} The complex Ginzburg--Landau equation governs a complex field $u$ and is an evolution PDE with chaotic dynamics,
\begin{equation}
    \partial_t u - u - (1 + i\alpha)\partial_x^2 u + (1 + i\beta)|u|^2u = f.
\end{equation}
We take $(\alpha,\beta) = (2,-2)$ on $[0,6.4]_t\times[0,50]_x$, with initial conditions drawn from the invariant measure. These parameters are in the \emph{defect turbulence} region \citep{shraiman1992spatiotemporal}, marked by persistent spatio-temporal disorder and phase singularities ($|u|=0$). Under other parameters, CGLE can instead exhibit phase turbulence, where the amplitude stays positive and slaved to the phase, whose equation then reduces asymptotically to the Kuramoto--Sivashinsky equation \citep{shraiman1992spatiotemporal, aranson2002world}; the defect-turbulence regime studied here therefore makes the two chaotic benchmarks complementary rather than redundant. The same parameter pair has been used as a defect-chaos benchmark for model-free prediction with reservoir and recurrent networks \citep{jiang2019model}; to our knowledge, however, the CGLE has not been adopted as an operator-learning benchmark, whereas the Kuramoto--Sivashinsky equation remains the standard chaotic test case for operator learning \citep{li2022learning}, with solver-generated training trajectories.
 
\begin{figure}[t]
    \centering
    \begin{subfigure}{0.32\textwidth}
        \centering
        \includegraphics[width=\textwidth]{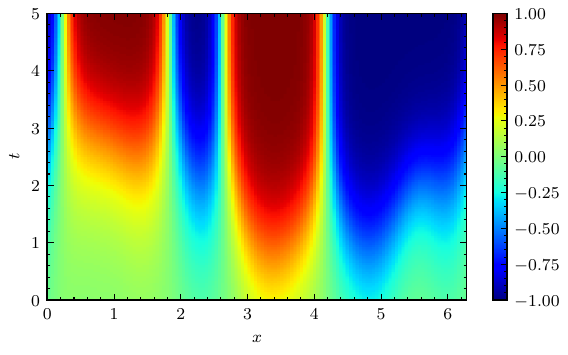}
        \caption{Allen--Cahn, reference}
        \label{fig:ac_illustration}
    \end{subfigure}
    \begin{subfigure}{0.32\textwidth}
        \centering
        \includegraphics[width=\textwidth]{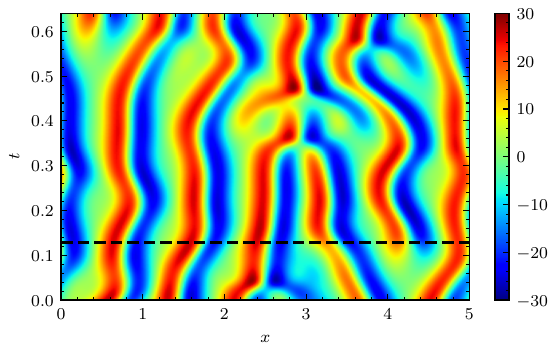}
        \caption{KS, reference}
        \label{fig:ks_illustration}
    \end{subfigure}
    \begin{subfigure}{0.32\textwidth}
        \centering
        \includegraphics[width=\textwidth]{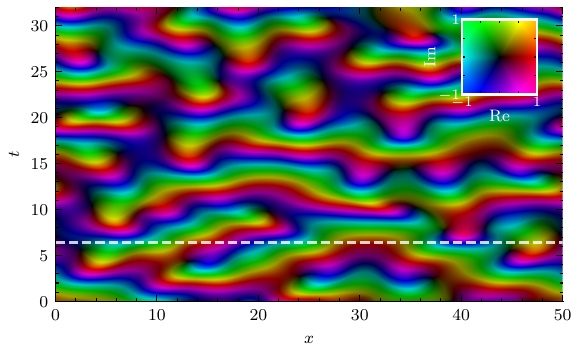}
        \caption{CGLE, reference}
        \label{fig:gl_illustration}
    \end{subfigure}
    \par\vspace{1.0em}
    \begin{subfigure}{0.32\textwidth}
        \centering
        \includegraphics[width=\textwidth]{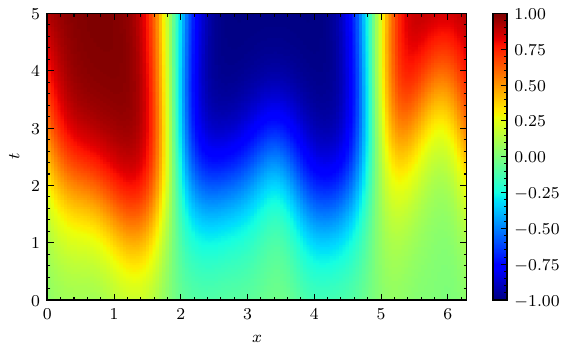}
        \caption{Allen--Cahn, $\mu_0$}
        \label{fig:ac_sample}
    \end{subfigure}
    \begin{subfigure}{0.32\textwidth}
        \centering
        \includegraphics[width=\textwidth]{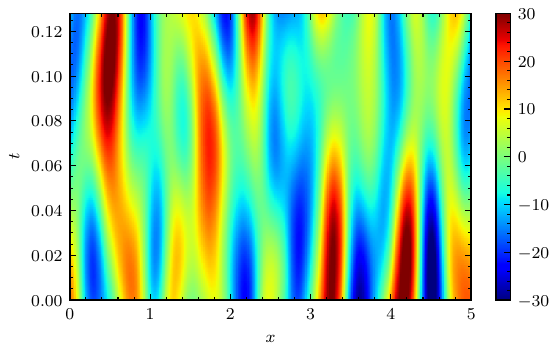}
        \caption{KS, $\mu_0$}
        \label{fig:ks_sample}
    \end{subfigure}
    \begin{subfigure}{0.32\textwidth}
        \centering
        \includegraphics[width=\textwidth]{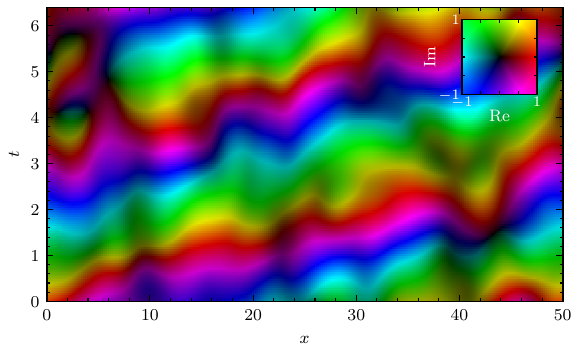}
        \caption{CGLE, $\mu_0$}
        \label{fig:gl_sample}
    \end{subfigure}
    \caption{Top row, representative solutions of the three benchmark equations: (a) Allen--Cahn on the training domain $[0,5]_t\times[0,2\pi]_x$ with $u\in[-1,1]$, showing phase separation into $\pm1$ domains, and (b) Kuramoto--Sivashinsky on $[0,0.64]_t\times[0,5]_x$ and (c) complex Ginzburg--Landau on $[0,32]_t\times[0,50]_x$, both with the time axis extended to $5\times$ the training horizon to expose the spatio-temporal chaos. Bottom row, representative samples $u\sim\mu_0$ from the manufactured solution distribution (\Cref{sec:mu0-construction}), on the training domains (d) $[0,5]_t\times[0,2\pi]_x$, (e) $[0,0.128]_t\times[0,5]_x$ and (f) $[0,6.4]_t\times[0,50]_x$. The complex Ginzburg--Landau fields in (c) and (f) are colored by phase and magnitude (inset color wheel); (c) shows the phase singularities ($|u|=0$) of the defect-turbulence regime.}
    \label{fig:dynamics_illustration}
\end{figure}

\Cref{fig:dynamics_illustration} shows representative solutions of all three systems, together with samples from the manufactured solution distribution.

\subsection{Experiment Setup}
\label{sec:experiments-setup}
We choose the Fourier Neural Operator (FNO) as the parametrization of $\cg_\theta$ in all experiments \citep{li2021fourier}.  {Inputs} and solutions are discretized on a uniform space--time grid of resolution $(n_t,n_x)$, namely $(128, 128)$ for Allen--Cahn and $(128, 256)$ for KS and the CGLE, with spacings $\delta t=T/n_t$ and $\delta x=L/n_x$; every model is trained and evaluated at the fixed resolutions. The FNO carries a strong periodic inductive bias, so to account for the non-periodic temporal boundary, we zero-pad the time dimension by $10\%$ and include a linear temporal encoding. Both terms of \eqref{eq:loss_fn} use $L^2$-type norms, the  {input} term being a weighted sum of an interior residual and an initial-condition error. All five methods share the optimizer, schedule, step budget, and total batch size, and differ only in how the batch is sampled and split between the solution and residual terms (\Cref{app:hparams-configs}), so the comparison is compute-matched.

For the KS equation, the predicted field is low-pass filtered to its leading $96$ spatial Fourier modes at the end of each forward pass, for all five methods, since the $\partial_x^4u$ term of the residual scales as $k^4$ and is numerically unstable in the highest modes. For the complex-valued CGLE field, the real and imaginary parts are treated as two real channels rather than using a complex-valued FNO.

All reference data, both the held-out test trajectories and the solver rollouts of \Cref{sec:evaluation}, are generated with a fourth-order exponential time-differencing Runge--Kutta scheme (ETD-RK4) \citep{cox2002exponential, kassam2005fourth} on the grids of \Cref{app:hparams-discretization}. For KS and the CGLE the initial conditions are snapshots of long ETD-RK4 trajectories recorded once the transient has decayed, so they are samples of the invariant measure.

\subsection{Evaluation}
\label{sec:evaluation}

As the primary evaluation metric, we evaluate the target risk \eqref{eq:target-risk} normalized by the solver-predicted RMS amplitude of the field. Since two of the benchmark equations are chaotic, a finite-horizon pointwise error cannot characterize the dynamics on its own: it grows over the rollout and eventually saturates. We therefore also evaluate statistics of the invariant measure and the spatial and temporal autocorrelations, defined below.

For the statistical observables, let $\mu^\infty$ and $\hat\mu^\infty$ denote the invariant measure of the solution field, estimated empirically from the reference solver and from an autoregressive rollout of the learned operator\footnote{Since $\cg_\theta$ predicts on the fixed window $[0,T]\times\Omega$ with $\Omega=[0,L]$, longer trajectories are generated autoregressively. The terminal slice $u(T,\cdot)$ of a predicted field becomes the initial condition of the next, with $f\equiv0$ throughout. We call each application an autoregressive window.}, respectively. We denote the spatial marginal of the invariant measures as $\mu^\infty_{\mathrm{spat}}$ and $\hat\mu^\infty_{\mathrm{spat}}$. Both are stationary in time, so $\eee_{u\sim\mu^\infty}[\,\cdot\,]$ averages over the ensemble and over the base time. Averaging over the periodic domain of length $L$, the correlation function is
\[
    R_{\mu^\infty}(\Delta t, \Delta x):=
    \eee_{u\sim\mu^\infty}\!\left[\frac{1}{L}\int_0^L
    \left(u(t,x) -  \bar u\right)\left( u(t+\Delta t,x+\Delta x)-\bar u\right)^{*}dx\right],
\]
which is independent of the base time $t$ by stationarity of $\mu^\infty$; here $(\cdot)^{*}$ is complex conjugation, reducing to the identity for the real-valued Allen--Cahn and KS fields. The scalar mean $\bar u$ is taken over the same ensemble and averaged over space--time.

\begin{enumerate}
    \item \emph{Relative solution error.} The aggregate relative $L^2$ error of the predicted solution under the target distribution, normalized over the test set rather than per sample,
    \[
        \frac{\eee_{v\sim\nu_1}\|\cg_\theta(v) - \cg(v)\|_2}{\eee_{v\sim\nu_1}\|\cg(v)\|_2},
    \]
    evaluated on the prediction domain $\Omega$.
    \item \emph{Invariant-measure distance.} The $2$-Wasserstein distance $W_2\left(\mu^\infty_{\mathrm{spat}},\hat\mu^\infty_{\mathrm{spat}}\right)$ between the reference and predicted invariant measures marginalized over time.
    \item \emph{Spatial autocorrelation.} The normalized spatial autocorrelation at separation $\Delta x$,
    \[
        R_{\Delta x}:=\frac{R_{\hat\mu^\infty}(0, \Delta x)}{R_{\hat\mu^\infty}(0, 0)} .
    \]
    \item \emph{Temporal autocorrelation.} The analogous normalized autocorrelation at separation $\Delta t$,
    \[
        R_{\Delta t}:=\frac{R_{\hat\mu^\infty}(\Delta t, 0)}{R_{\hat\mu^\infty}(0, 0)} .
    \]
    \item \emph{Energy spectrum.} The spatial energy spectrum as a function of wavenumber $k$,
    \[
        \varepsilon_k=\eee_{u\sim\hat\mu^\infty_{\mathrm{spat}}}|\mathcal F_k(u)|^2,
    \]
    where $\mathcal F_k(u)$ is the $k$-th spatial Fourier mode.
\end{enumerate}
 
Metric~1 is reported for all three equations; metrics~2--5 characterize a statistical steady state and are reported only for the chaotic KS and CGLE systems. To reach that steady state, we draw $v\sim\nu_1$ from test data obtained by running a numerical solver and discard a burn-in of $10000\,\delta t$ so that the statistics are independent of the initialization (\Cref{app:hparams-eval}).
 
For the invariant-measure distance (metric~2), the Sinkhorn estimator is biased upward by finite sampling and by the entropic regularization, so even two independent solver rollouts are separated by a strictly positive expected distance. To calibrate this offset, we report a solver-only reference from two independent ETD-RK4 rollouts, with sampling uncertainty estimated from the empirical variance of the Sinkhorn dual potentials, following the entropic-OT central limit theorem \citep{mena2019statistical, gonzalez2022weak}; a model whose distance lies within $2\sigma$ of this reference is statistically indistinguishable from the solver at the available sample size, and such entries are boldfaced along with the best model.

Metrics~3--5 are functions of $\Delta x$, $\Delta t$, and $k$, so we plot them against the ETD-RK4 reference and summarize each by its relative root-mean-square error (RMSE). The spatial separation is periodic and conjugate-symmetric, so the spatial correlation is shown over the full independent range $\Delta x\in[0,L/2]$. The energy spectrum is shown over the leading $40$ modes, whose spectral power already spans ten orders of magnitude. The temporal separation is unbounded; we therefore evaluate $\Delta t$ up to $127\,\delta t$, corresponding to one autoregressive window.

\subsection{Results}
\label{sec:results}

\begin{table}[t]
    \centering
    \caption{Evaluation of WaSP against two baselines and two ablations on the Allen--Cahn, Kuramoto--Sivashinsky (KS), and complex Ginzburg--Landau (CGLE) equations. Every method follows the solver-free training protocol. The baselines are PINO, which minimizes the PDE residual and initial-condition losses, and MD, which regresses on the fixed manufactured pairs $(\cn(u),u)$ with $u\sim\mu_0$. Each ablation is listed under the baseline it extends: PINO+OT trains the residual loss along the McCann interpolation, and MD+PINO is the WaSP objective \eqref{eq:loss_fn} frozen at $s=0$. Columns are metrics~1--5 of \Cref{sec:evaluation}; dashes mark the metrics reported only for the chaotic systems. WaSP attains the best or statistically indistinguishable value on every metric and equation.}

    \label{tab:evaluation_metrics}
    \resizebox{\textwidth}{!}{%
    \begin{tabular}{llccccc}
        \toprule
        Equation & Method & Sol. Rel. $L^2$ & Inv. Meas. Dist. & Spat. Corr. RMSE & Temp. Corr. RMSE & Spectrum RMSE \\
        \midrule
        \multirow{5}{*}{\shortstack{Allen--Cahn}} & WaSP (ours) & $\mathbf{8.52 \times 10^{-3}}$ & --- & --- & --- & --- \\
        & PINO & $3.29 \times 10^{-2}$ & --- & --- & --- & --- \\
        & PINO+OT & $4.73 \times 10^{-2}$ & --- & --- & --- & --- \\
        & MD & $1.10 \times 10^{-1}$ & --- & --- & --- & --- \\
        & MD+PINO & $1.06 \times 10^{-1}$ & --- & --- & --- & --- \\
        \midrule
        \multirow{6}{*}{\shortstack{KS equation}} & WaSP (ours) & $\mathbf{2.19 \times 10^{-2}}$ & $\mathbf{8.69 \times 10^{0}}$ & $\mathbf{1.02 \times 10^{-1}}$ & $\mathbf{8.45 \times 10^{-2}}$ & $\mathbf{1.04 \times 10^{-1}}$ \\
        & PINO & $2.08 \times 10^{-1}$ & $9.93 \times 10^{0}$ & $3.87 \times 10^{-1}$ & $1.46 \times 10^{-1}$ & $4.70 \times 10^{-1}$ \\
        & PINO+OT & $1.02 \times 10^{-1}$ & $1.48 \times 10^{1}$ & $1.16 \times 10^{0}$ & $2.67 \times 10^{-1}$ & $1.80 \times 10^{0}$ \\
        & MD & $3.38 \times 10^{-1}$ & $1.67 \times 10^{1}$ & $1.46 \times 10^{0}$ & $6.27 \times 10^{-1}$ & $2.48 \times 10^{0}$ \\
        & MD+PINO & $3.64 \times 10^{-1}$ & $1.68 \times 10^{1}$ & $1.44 \times 10^{0}$ & $4.57 \times 10^{-1}$ & $2.40 \times 10^{0}$ \\
        & \emph{Solver reference} & --- & $(7.87 \pm 0.05)\times 10^0$ & --- & --- & --- \\
        \midrule
        \multirow{6}{*}{\shortstack{CGLE}} & WaSP (ours) & $\mathbf{7.08 \times 10^{-2}}$ & $\mathbf{4.38 \times 10^{-1}}$ & $\mathbf{9.10 \times 10^{-3}}$ & $\mathbf{2.90 \times 10^{-2}}$ & $\mathbf{1.23 \times 10^{-2}}$ \\
         & PINO & $3.23 \times 10^{-1}$ & $\mathbf{4.34 \times 10^{-1}}$ & $3.98 \times 10^{-2}$ & $1.01 \times 10^{-1}$ & $5.20 \times 10^{-2}$ \\
         & PINO+OT & $1.16 \times 10^{-1}$ & $4.38 \times 10^{-1}$ & $1.23 \times 10^{-2}$ & $5.68 \times 10^{-2}$ & $1.36 \times 10^{-2}$ \\
         & MD & $6.65 \times 10^{-1}$ & $5.77 \times 10^{-1}$ & $1.46 \times 10^{0}$ & $5.10 \times 10^{-1}$ & $1.76 \times 10^{0}$ \\
         & MD+PINO & $7.54 \times 10^{-1}$ & $5.89 \times 10^{-1}$ & $9.79 \times 10^{-1}$ & $5.17 \times 10^{-1}$ & $1.63 \times 10^{0}$ \\
        & \emph{Solver reference} & --- & $(4.37 \pm 0.003)\times 10^{-1}$ & --- & --- & --- \\
        \bottomrule
    \end{tabular}%
    }
\end{table}

\Cref{tab:evaluation_metrics} reports all metrics. WaSP attains the lowest solution relative $L^2$ error on every equation, improving on PINO, the strongest baseline, by factors of $3.9$ on Allen--Cahn, $9.5$ on KS, and $4.6$ on the CGLE, and on MD by $9.4\times$ to $15.4\times$. Against the PINO+OT ablation the margin is narrower on the two chaotic equations, $4.7\times$ on KS and $1.6\times$ on the CGLE, and wider on Allen--Cahn at $5.6\times$. The other ablation, MD+PINO, offers no consistent improvement over MD. 

\begin{figure}
    \centering
    \begin{subfigure}{0.8\textwidth}
        \centering
        \includegraphics[width=\textwidth]{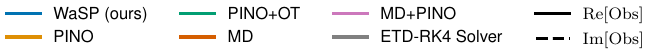}
    \end{subfigure}
    \begin{subfigure}{0.48\textwidth}
        \centering
        \includegraphics[width=\textwidth]{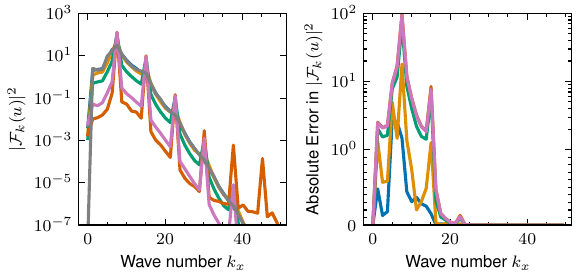}
        \caption{KS, energy spectrum}
        \label{fig:ks_spectrum}
    \end{subfigure}
    \begin{subfigure}{0.48\textwidth}
        \centering
        \includegraphics[width=\textwidth]{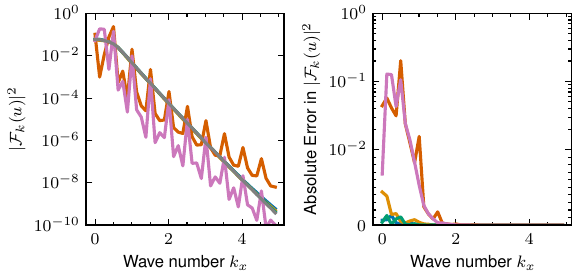}
        \caption{CGLE, energy spectrum}
        \label{fig:gl_spectrum}
    \end{subfigure}
    \par\vspace{0.6em}
    \begin{subfigure}{0.48\textwidth}
        \centering
        \includegraphics[width=\textwidth]{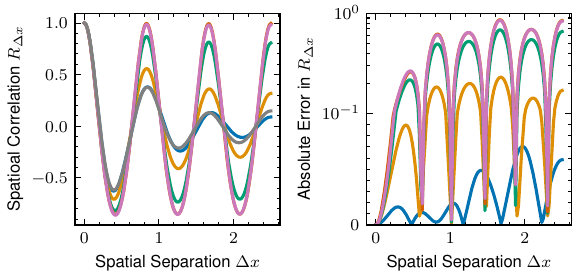}
        \caption{KS, spatial autocorrelation}
        \label{fig:ks_spatial_corr}
    \end{subfigure}
    \begin{subfigure}{0.48\textwidth}
        \centering
        \includegraphics[width=\textwidth]{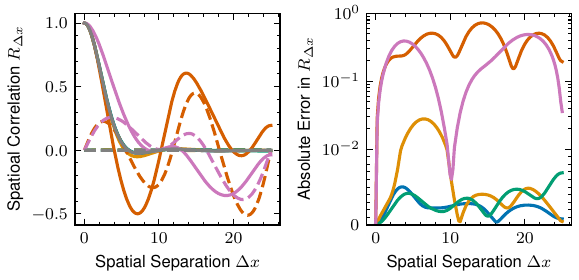}
        \caption{CGLE, spatial autocorrelation}
        \label{fig:gl_spatial_corr}
    \end{subfigure}
    \par\vspace{0.6em}
    \begin{subfigure}{0.48\textwidth}
        \centering
        \includegraphics[width=\textwidth]{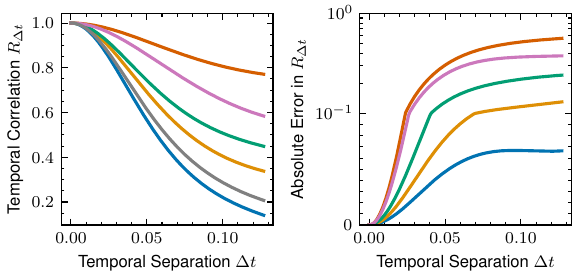}
        \caption{KS, temporal autocorrelation}
        \label{fig:ks_temporal_corr}
    \end{subfigure}
    \begin{subfigure}{0.48\textwidth}
        \centering
        \includegraphics[width=\textwidth]{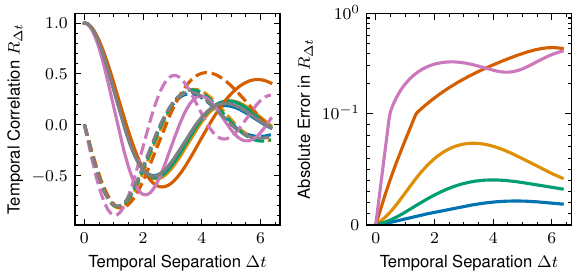}
        \caption{CGLE, temporal autocorrelation}
        \label{fig:gl_temporal_corr}
    \end{subfigure}
 
    \caption{Function-valued statistical observables for the Kuramoto--Sivashinsky and complex Ginzburg--Landau equations, plotted against the wavenumber $k$ and the separations $\Delta x$ and $\Delta t$. Within each panel, the left plot shows the raw curves against the ETD-RK4 reference and the right plot the absolute deviation from that reference on a symmetric-log scale, whose linear region is marked by evenly spaced minor ticks. The legend at the top applies to all six panels and distinguishes the five methods, the ETD-RK4 reference, and the real and imaginary parts of each complex observable.}
    \label{fig:functional_obs}
\end{figure}

For the statistical observables, WaSP attains the best or a statistically indistinguishable value throughout. \Cref{fig:functional_obs} plots all three function-valued observables against the ETD-RK4 reference. On KS, WaSP improves on every other method on all four metrics, by $1.1\times$ to $1.9\times$ on the invariant measure and by $1.7\times$ to $24\times$ on the three function-valued observables. On the CGLE the invariant measure saturates: WaSP, PINO and PINO+OT all lie within $2\sigma$ of the solver reference, so that metric separates none of the three, and all three also reach small spatial-correlation and spectrum errors. What separates them is how the field is organized in time. PINO matches the invariant measure to within the solver reference ($4.3\times10^{-1}$) yet its temporal-correlation error exceeds WaSP's by more than a factor of three ($1.0\times10^{-1}$ versus $2.9\times10^{-2}$), recovering the static distribution of states but not their temporal structure. \Cref{fig:gl_temporal_corr} shows the failure mode. PINO matches the reference at small temporal separation and then drifts steadily out of phase, the signature of a small error in the nonlinear frequency of the rotating field, which accumulates linearly in the separation.

\begin{figure}[t]
    \centering
    \begin{subfigure}{0.8\textwidth}
        \centering
        \includegraphics[width=\textwidth]{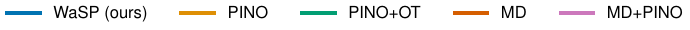}
    \end{subfigure}
    \begin{subfigure}{0.32\textwidth}
        \centering
        \includegraphics[width=\textwidth]{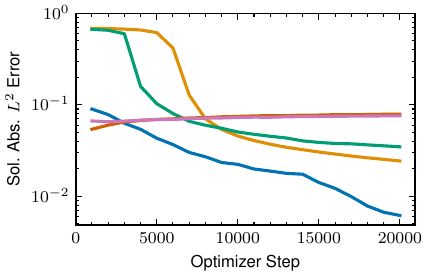}
        \caption{Allen--Cahn}
        \label{fig:ac_sol_loss}
    \end{subfigure}
    \begin{subfigure}{0.32\textwidth}
        \centering
        \includegraphics[width=\textwidth]{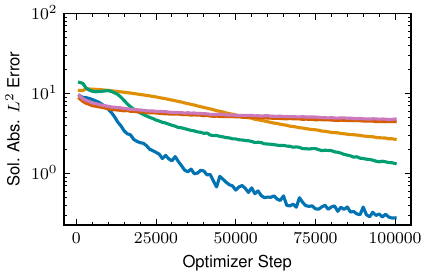}
        \caption{KS equation}
        \label{fig:ks_sol_loss}
    \end{subfigure}
    \begin{subfigure}{0.32\textwidth}
        \centering
        \includegraphics[width=\textwidth]{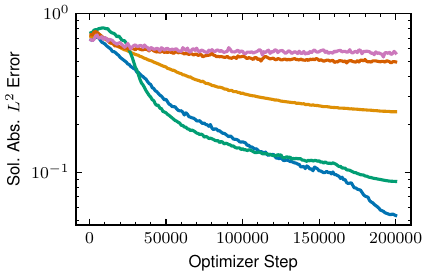}
        \caption{CGLE}
        \label{fig:gl_sol_loss}
    \end{subfigure}
    \caption{The panels plot the validation absolute $L^2$ error of the predicted solution on the target distribution $\nu_1$ (log scale) against optimizer step, for WaSP, the two baselines (PINO and MD), and the two ablations (PINO+OT and MD+PINO).}
    \label{fig:sol_loss_curves}
\end{figure}

\begin{figure}
    \centering
    \begin{subfigure}{0.8\textwidth}
        \centering
        \includegraphics[width=\textwidth]{images/legend.pdf}
    \end{subfigure}
    \begin{subfigure}{0.32\textwidth}
        \centering
        \includegraphics[width=\textwidth]{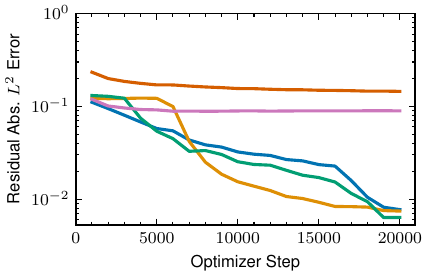}
        \caption{Allen--Cahn}
        \label{fig:ac_cond_loss}
    \end{subfigure}
    \begin{subfigure}{0.32\textwidth}
        \centering
        \includegraphics[width=\textwidth]{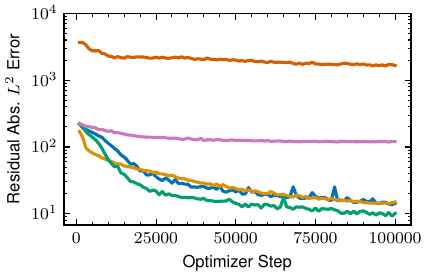}
        \caption{KS equation}
        \label{fig:ks_cond_loss}
    \end{subfigure}
    \begin{subfigure}{0.32\textwidth}
        \centering
        \includegraphics[width=\textwidth]{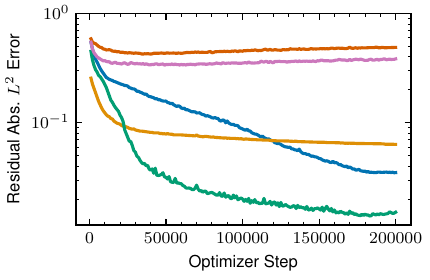}
        \caption{CGLE}
        \label{fig:gl_cond_loss}
    \end{subfigure}
    \caption{The panels plot the validation $L^2$ norm of the predicted PDE residual on the target distribution $\nu_1$ (log scale) against optimizer step, for WaSP, the two baselines (PINO and MD), and the two ablations (PINO+OT and MD+PINO). Note that smaller PDE residual does \emph{not} imply better accuracy as argued in \Cref{sec:setup}.}
    \label{fig:cond_loss_curves}
\end{figure}

\paragraph{Physics-Informed Track: PINO and PINO+OT.} \Cref{fig:sol_loss_curves,fig:cond_loss_curves} plot the validation solution and residual error on the target distribution $\nu_1$ against optimizer step; no checkpoint or hyperparameter is selected on these curves, and every reported number is from the final step (\Cref{app:hparams-eval}). Two readings follow. First, curriculum learning is known to mitigate part of the optimization difficulty of physics-informed methods \citep{krishnapriyan2021characterizing, wang2024respecting}, and \Cref{fig:cond_loss_curves} shows the terminal PDE residual of PINO+OT below that of PINO on all three equations, so the interpolation path defines a curriculum that helps even without self-generated supervision. Second, as \Cref{sec:residual-stability} shows, a residual measured in too weak a norm cannot certify the solution error, and the curves exhibit exactly the obstruction \Cref{thm:ks-residual-stability} proves. PINO+OT reaches a lower residual than WaSP on every equation and a higher solution error on every equation, and on Allen--Cahn it reaches a lower residual than PINO together with a higher solution error.

\paragraph{Manufactured-Data Track: MD and MD+PINO.} MD and MD+PINO are the weakest methods throughout, and they inherit the structural distribution shift of \Cref{sec:manufactured-shift}. Two observations support this. First, both are trained on the same $\mu_0$ as WaSP, constructed for each equation to lie as close to the target solution measure as we could make it (\Cref{sec:mu0-construction}), yet neither recovers the target dynamics; a well-matched starting distribution is thus insufficient on its own, consistent with the obstruction of \Cref{thm:manufactured-shift-informal}. Second, \Cref{fig:ks_spectrum,fig:ks_spatial_corr} show that both fail to capture the spatial structure and collapse onto the dominant sixth harmonic, the fastest-growing linearly unstable mode, which shows as evenly spaced spectral spikes. The imaginary spatial correlation in \Cref{fig:gl_spatial_corr} is the sharper case: the reference has none and $\mu_0$ has none by construction, yet MD and MD+PINO produce one and WaSP and PINO do not, so the artifact is generated during the autoregressive rollout rather than inherited from the training distribution. We read both as the same mechanism, where a test-time  {input} drawn from $\nu_1$ is slightly out of distribution for a method trained on $\mu_0$, the bias of the learned operator carries it further out, and the rollout settles on a spurious attractor.
 
\section{Discussion}
\label{sec:discussion}
This section examines the two design inputs WaSP depends on, the manufactured solution measure $\mu_0$ (\Cref{sec:mu0-construction}) and the transport map $\ct$ (\Cref{sec:ot-cost}).

\subsection{Construction of the Manufactured Solution Distribution}
\label{sec:mu0-construction}
WaSP imposes no structural constraints on $\mu_0$, but selecting it close to the target solution measure $\mu_1 := \cg_{\#}\nu_1$ is essential for fast convergence, as it shortens the transport path WaSP must traverse. By \Cref{thm:manufactured-shift-informal}, however, no tractable calibration of $\mu_0$ can remove the distribution gap entirely. We describe the construction of $\mu_0$ for each benchmark equation below; representative samples are shown in \Cref{fig:dynamics_illustration}.

\paragraph{Allen--Cahn.} We sample $u(0,x)$ from the target initial-condition distribution and obtain each output time $t$ by a \emph{single} Lie--Trotter splitting step of size $t$ applied to $u(0,\cdot)$. The reaction ODE $u_t=u-u^3$ is a Bernoulli equation, solved in closed form at each grid point,
\begin{equation*}
    \Phi(t; u_0)=\frac{u_0}{\sqrt{e^{-2t}\left(1-u_0^2\right)+u_0^2}},
\end{equation*}
and the diffusion operator $e^{\nu t\partial_x^2}$, with $\nu$ the dissipation parameter of \Cref{sec:equations}, is then applied exactly in Fourier space as the multiplier $e^{-\nu t k_x^2}$, so that $u(t,\cdot)=e^{\nu t\partial_x^2}\Phi\!\left(t;u(0,\cdot)\right)$. A single step of size $t$ is not a convergent splitting solver but a cheap sampler whose time slices are independent and can therefore be evaluated in parallel across $t$. The target initial condition for Allen--Cahn is itself a Gaussian random field, so the initial-condition marginal of $\nu_0$ coincides with that of $\nu_1$ by construction. The optimal map therefore reduces to the identity on the initial condition and the zero map on the forcing: the path still anneals $f\mapsto(1-s)f$ under the schedule of \Cref{app:hparams-transport}, but no Sinkhorn problem is solved.

\paragraph{Kuramoto--Sivashinsky.} We model $\mu_0$ as a space--time Gaussian random field (GRF) whose power spectrum is separable in the temporal and spatial frequencies $(k_t, k_x)$. For the spatial factor we use the power spectrum $P_x(k_x):=\eee\big[\,|\widehat{u_0}(k_x)|^2\,\big]$ of the target initial-condition distribution, estimated from the same unpaired target   {inputs} that are used to fit $\ct$ (\Cref{sec:ot-cost}), so that samples inherit its spatial statistics. We extend it to a 2D space--time spectrum by assuming a separable $P(k_t, k_x) = P_t(k_t)\,P_x(k_x)$ with a normalized Gaussian temporal factor $P_t(k_t) \propto \exp\left(-k_t^2 / 2\sigma_t^2\right)$, which gives $\eee\|\partial_t u\|^2/\eee\|u\|^2=\sigma_t^2$. The temporal factor is calibrated to the rate at which the true dynamics evolve the field at $t=0$. Setting $f\equiv0$ and solving the equation for $\partial_t u$ gives $\widehat{\partial_t u}(k_x)\big\vert_{t=0}$ in $u_0$, and therefore
\begin{equation}
\label{eq:ks-sigmat}
    \sigma_t^2
    := \frac{\eee\big[\,\sum_{k_x} |-\tfrac{1}{2}\, i k_x\, \widehat{u_0^2}(k_x)
      + \big(k_x^2 - \nu k_x^4\big)\, \widehat{u_0}(k_x)|^2\,\big]}
            {\eee\big[\,\sum_{k_x} |\widehat{u_0}(k_x)|^2\,\big]},
\end{equation}
calibrates the temporal factor. Because an FFT-sampled GRF is periodic along each axis, we sample on a doubled temporal grid of length $2n_t$ and retain the first half, removing the spurious temporal periodicity that would otherwise be imposed on $\mu_0$.

\paragraph{Complex Ginzburg--Landau.} As shown in \Cref{fig:gl_illustration}, the CGLE's dynamics combine locally plane-wave patches, separated by phase defects, with spatio-temporal chaos appearing at larger scale. We therefore build each sample around an unstable plane wave and perturb it with a GRF. We first draw a spatial wavenumber $k_0$ with probability proportional to $P_x$, now recomputed from the target CGLE initial-condition distribution, and form the corresponding plane wave, whose temporal frequency follows the CGLE dispersion relation
\begin{equation*}
    \omega(k_0) = -\beta - (\alpha - \beta)\, k_0^2.
\end{equation*}
Here $\omega(k_0)$ is the frequency of the exact unstable plane wave of amplitude $\sqrt{1-k_0^2}$. We then modulate this wave by a complex GRF $\xi$ drawn from the same separable space--time spectrum $P(k_t, k_x)$ used in the KS construction, recalibrated with unforced CGLE. The doubled-grid sampling that removes temporal periodicity is also in place, giving the sample
\begin{equation*}
    u(t,x) = \tfrac{1}{\sqrt 2}\,\big(1 + \xi(t,x)\big)\, e^{\,i\left(\omega(k_0)\, t + k_0 x\right)}.
\end{equation*}
The multiplicative perturbation preserves the plane-wave backbone while injecting the small-scale fluctuations characteristic of the defect-turbulence regime. The amplitude is fixed at $1/\sqrt2$ rather than the exact $\sqrt{1-k_0^2}$; the pair $(\cn(u),u)$ is exact by construction for any sampled $u$, and a constant preserves the output norm.

\subsection{Cost of the Transport Estimate}
\label{sec:ot-cost}
The transport map $\ct$ in \eqref{eq:mccann-path} is needed only to define the interpolation path, and estimating it is a one-time preprocessing step, separate from neural-operator training. Because the target forcing distribution is the point mass at $f=0$, any transport to $\nu_1$ must send the forcing component to zero, so the map factors exactly: on the forcing component $\ct$ is the zero map and \eqref{eq:mccann-path} reduces to linear annealing $f\mapsto(1-s)f$. Optimal transport is therefore required only for the spatial initial condition, whose dimension is the spatial grid size rather than the full space--time grid. For that component, we estimate $\ct$ with the entropy-regularized Sinkhorn algorithm under the squared-Euclidean cost \citep{cuturi2013sinkhorn}, run in the log domain on top of matrix-free GPU kernel reductions \citep{JMLR:v22:20-275}, so that the $n\times m$ coupling between the $n$ source and $m$ target samples is never materialized in memory. To reduce dimension, we project both clouds onto their leading principal components before solving the transport problem, and recover the map on new samples by barycentric projection of the entropic plan \citep{pooladian2021entropic}. Its settings are listed in \Cref{tab:hp-transport}.

\paragraph{Cost Analysis.} Because the path endpoints are fixed, $\ct$ is estimated once and reused at every optimizer step, so even a comparatively expensive estimate is amortized over the full training run. In our experiments, the Sinkhorn iterations account for $<10\,\mathrm{sec}$ on one RTX 5090. Furthermore, each Sinkhorn iteration scales as $O(nmq)$ in the sample counts $n,m$ and the PCA rank $q$, independently of the space--time resolution (\Cref{app:hparams-transport}). However, applying the map recurs rather than amortizes, since the barycentric projection has no closed form: each freshly drawn   {input} costs one reduction over the $m$ target fitting samples, $O(mq)$ in the projected space plus $O(md)$ for the barycentric average, a few MFLOPs at typical settings and well below one forward pass of $\cg_\theta$ on the same field. Forming $\nu_s$ therefore adds a negligible per-step overhead on top of the one-time estimate.
\section{Related Work}
\label{sec:related-works}
Having presented the method and its evaluation, we now place WaSP in context.

\paragraph{Neural Operators.} Neural operators learn mappings between infinite-dimensional function spaces and are now a leading approach to data-driven PDE surrogate modeling \citep{kovachki2023neural}. The main families parametrize the integral kernel with a graph neural network \citep{li2020neural}, in the spectral domain via the $O(n\log n)$ FFT \citep{li2021fourier}, or, in the Galerkin Transformer, through a Petrov--Galerkin projection in place of softmax attention \citep{cao2021choose}; DeepONet instead encodes the input function and the output query coordinates in separate branch--trunk networks \citep{lu2021learning}. Pretrained foundation models generalize across PDE families \citep{herde2024poseidon}, amortizing the cost of labeled data rather than removing it, since the pretraining dataset is itself solver-generated. WaSP is architecture-agnostic and can be used to train any of these neural operator architectures; our experiments use the Fourier Neural Operator throughout (\Cref{sec:experiments-setup}).

\paragraph{Physics-Informed Neural Operators.}
Physics-informed neural operators incorporate the PDE residual objective \citep{raissi2019physics} into operator training to reduce reliance on solver-generated data \citep{li2024physics, goswami2023physics}. The resulting objective is difficult to optimize. For PINNs, the documented mechanisms include gradient imbalance across loss terms \citep{wang2021understanding}, a loss landscape that becomes ill-conditioned as the PDE coefficients grow \citep{krishnapriyan2021characterizing}, and a spectral bias that fits low-frequency components far faster than high-frequency ones \citep{wang2022and}. A large body of work mitigates these pathologies by annealing the PDE coefficient toward its target value \citep{krishnapriyan2021characterizing}, weighting the residual causally in time \citep{wang2024respecting}, replacing first-order optimizers with quasi-second-order ones \citep{wang2026gradient}, and computing the spectral derivatives of the residual more accurately on non-periodic domains \citep{ganeshram2022fc}. A separate issue is that the residual is not a certificate of solution quality on its own, and \citet{wang20222} show that for the Hamilton--Jacobi--Bellman equation a small $L^2$ residual loss coexists with a large error in the solution metric, so the residual norm must be strong enough to control that metric. WaSP reformulates training as a sequence of supervised problems, and uses the residual as a local regularizer that holds the realized    {input} distribution close to the interpolated one rather than as its certificate for target risk (\Cref{sec:analysis,sec:implementation}).

\paragraph{Manufactured Data for Operator Learning.}
Several recent methods reuse the method of manufactured solutions \citep{roache2002code} as a data source for operator learning, evaluating the PDE operator on sampled solutions to obtain training pairs with no numerical solve. \citet{hasani2024generating} manufacture random solutions as random combinations of eigenfunctions of the leading-order differential operator and evaluate the approach on elliptic boundary-value problems; \citet{sojitra2025method} extend the idea to evolution PDEs; and \citet{bischof2026hypino} combine manufactured pairs with a physics-informed objective on unlabeled samples to train a hypernetwork for PINNs. However, the manufactured solution prior induces a distribution shift, and addressing it remains an open problem, existing methods either rely on calibrating the solution prior to align with the equation and geometry \citep{hasani2024generating} or use physics-informed learning as a test-time adaptation framework \citep{bischof2026hypino}. WaSP starts from the same data-generation mechanism and bridges the remaining gap with an explicit transport path. A well-chosen manufactured distribution still shortens that path (\Cref{sec:mu0-construction}); what WaSP removes is the requirement that the two distributions be close enough for a single generalization step to succeed.

\paragraph{Optimal Transport.}
Optimal transport (OT) provides a principled framework for comparing and interpolating between probability measures \citep{villani2009optimal, gabriel2019computational}. Brenier's theorem guarantees the existence and uniqueness of the optimal transport map between absolutely continuous measures as the gradient of a convex potential \citep{brenier1991polar}, and McCann's interpolation defines a constant-speed geodesic in the space of probability measures equipped with the Wasserstein metric \citep{mccann1997convexity}. Entropy-regularized OT enables tractable estimation of transport maps and Wasserstein distances at scale via the Sinkhorn algorithm \citep{cuturi2013sinkhorn}, and \citet{pooladian2021entropic} show how to estimate the barycentric projection of the entropic plan. Matrix-free and IO-aware GPU kernels lower the cost of the Sinkhorn iteration itself \citep{JMLR:v22:20-275, ye2026flashsinkhorn}. \Cref{sec:ot-cost} reports what the estimate costs in our setting.

\paragraph{Gradual Domain Adaptation.} In the domain adaptation literature, \citet{wang2022understanding} argue, based on their error bound, that optimal intermediate domains are the ones along the Wasserstein geodesic. \citet{he2024gradual} state this placement as an explicit design principle and show that generating the intermediate domains by McCann's interpolation and self-training on them improves gradual domain adaptation when real intermediate domains are scarce; \citet{el2025mccann} adopt the same construction. WaSP follows the same outline but changes the source of supervision. Gradual domain adaptation pseudo-labels each intermediate sample with the current model, so label error accumulates along the path; WaSP instead passes the model's output back through the PDE operator, so every intermediate pair satisfies the equation exactly (\Cref{sec:transporting-distribution}).

\paragraph{Curriculum Learning.} Curriculum learning trains models by presenting examples in a meaningful order, typically from easy to hard \citep{bengio2009curriculum}, which can improve convergence speed and final accuracy \citep{soviany2022curriculum}. Physics-informed curricula are built over the equation or over the time domain, as in the coefficient annealing and causal weighting discussed above; WaSP builds one over the    {input} distribution, interpolating along the Wasserstein path to present a sequence of problems that begins with well-conditioned supervised learning on manufactured data and moves toward the target.
\section{Conclusion}
\label{sec:conclusion}

We studied solver-free operator learning, where no paired input--solution data from the target distribution is available, and showed that both dominant routes face fundamental obstructions. Physics-informed training faces a regularity--variance tradeoff (\Cref{sec:residual-stability,sec:residual-variance}). A residual norm weak enough to keep the variance of mini-batch estimates manageable may fail to control the solution error, and a norm strong enough to control it amplifies the variance of gradients and hinders optimization. Manufactured-data training induces an input distribution that can be mutually singular with the target (\Cref{thm:manufactured-shift-informal}). In this case, even supervision with input--solution pairs does not limit the target error without further structural assumptions on the learned operator.

Motivated by this diagnosis, we introduced WaSP, which preserves the optimization advantage of supervised regression on manufactured data while mitigating the structural distribution shift identified for a fixed solution distribution. WaSP estimates an optimal transport map from the manufactured input distribution to the target and trains along the resulting Wasserstein interpolation, replacing one large manufactured-to-target shift with a curriculum of nearby supervised problems. At each step, the current operator predicts approximate solutions for the next intermediate distribution and manufactures exact input--solution pairs as the main learning signal; a local PDE residual term regularizes this self-generated supervision so that it cannot drift away from the transport path. A stepwise error-propagation bound under strong assumptions clarifies the intuition behind the algorithm: its success can be attributed to the local generalization property of neural operators. It also motivates an adaptive schedule that spends more steps on the parts of the path where the per-step error is largest.

Empirically, under a fully solver-free protocol that uses no labels from the target-distribution, WaSP attained the best or statistically indistinguishable values on every metric and test case across the Allen--Cahn, Kuramoto--Sivashinsky, and complex Ginzburg--Landau benchmark problems. It reduced the relative solution error by roughly $4\times$ to $10\times$ over the strongest baseline (PINO), and, on the two chaotic systems, more faithfully reproduced statistical observables including the invariant measure, spatial and temporal autocorrelations, and the energy spectrum. The results of the two ablations of PINO+OT and MD+PINO are consistent with \Cref{sec:negative-results}: PINO+OT mitigates the optimization challenges of PINO, but still fails to certify its solution error, and MD+PINO inherits the same structural distribution shift of MD.

Several directions remain open: the method requires a hand-constructed initial solution measure, its space--time memory footprint matches that of  physics-informed operators\footnote{This statement concerns only the GPU scaling. WaSP requires more CPU infrastructure and has worse scaling than PINO on a CPU. We note that CPU is not the practical bottleneck.} and the focused scope of study in this work (e.g., one-dimensional periodic evolution equations and the Fourier Neural Operator) invite extensions to higher dimensions, complex geometries, and other architectures. More broadly, our results suggest that  solver-free operator learning can be recast as transport along a curriculum of distributions, rather than as a single end-to-end objective. This approach combines the reliability of supervised regression with the physical grounding of the governing equation, a combination that may prove useful beyond operator learning.

\paragraph{Code Availability}
The project's code is available at \href{https://github.com/neuraloperator/WaSP.git}{https://github.com/neuraloperator/WaSP.git}.

\acks{Ryan Liu is supported in part by the National Science Foundation Graduate Research Fellowship Program under Grant No.~2139433. Anima Anandkumar is supported by the Bren named chair professorship, Schmidt AI2050 senior fellowship, and ONR (MURI grant N00014-18-1-2624). Any opinions, findings, and conclusions or recommendations expressed in this material are those of the author(s) and do not necessarily reflect the views of the National Science Foundation.}

\bibliography{ref}
\newpage
\appendix

The structure of the appendix is as follows.

\begin{itemize}
    \item \Cref{app:notation} collects notation and auxiliary definitions used in the theoretical analysis.
    \item \Cref{app:ks-stability-proof} provides the formal statement and proof of the KS residual-stability result.
    \item \Cref{app:pino-variance-proof} provides the formal analysis of residual-loss variance.
    \item \Cref{app:synthetic-shift-proof} provides the formal statement and proof of the manufactured-data distribution-shift result.
    \item \Cref{app:wasp-step-dependent-proof} provides the formal statement and proof of WaSP error propagation.
    \item \Cref{sec:hparams} documents the discretization, architecture, optimization, loss, replay-buffer, transport, baseline, and evaluation settings used in the experiments.
\end{itemize}

\section{Notation and Auxiliary Definitions}
\label{app:notation}

\begin{table}[h]
\centering
\caption{Notation used in the theoretical sections.}
\label{tab:notation}
\begin{tabular}{lll}
\toprule
Symbol & Definition & First used \\
\midrule
$\cu$ & Solution function space & \Cref{sec:setup} \\
$\cv$ &    {Input} function space & \Cref{sec:setup} \\
$\cn:\cu\to\cv$ & Differential operator mapping solution to    {input} & \Cref{sec:setup} \\
$\cg:\cv\to\cu$ & True solution operator & \Cref{sec:setup} \\
$\cg_\theta$ & Neural solution operator & \Cref{sec:setup} \\
$\spp(\mathcal X)$ & Probability distributions on $\mathcal X$ & \Cref{sec:setup} \\
$\ct_{\#}\rho$ & Pushforward of $\rho$ by $\ct$ & \Cref{sec:setup} \\
$\mu,\nu$ & Generic solution and    {input} distributions & \Cref{sec:setup} \\
$\mu_0$ & Initial manufactured solution distribution & \Cref{sec:transporting-distribution} \\
$\nu_0=\cn_{\#}\mu_0$ & Initial manufactured    {input} distribution & \Cref{sec:transporting-distribution} \\
$\nu_1$ & Target    {input} distribution & \Cref{sec:setup} \\
$\ct$ & Transport map from $\nu_0$ to $\nu_1$ & \Cref{sec:transporting-distribution} \\
$s_n$ & Discrete transport time & \Cref{sec:transporting-distribution} \\
$\nu_n$ & Intermediate    {input} distribution & \Cref{sec:transporting-distribution} \\
$\hat{\mu}_n$ & Proxy solution distribution generated at step $n$ & \Cref{sec:transporting-distribution} \\
$\hat{\nu}_n$ & Realized    {input} distribution generated at step $n$ & \Cref{sec:transporting-distribution} \\
$W$ & $1$-Wasserstein distance in the theory & \Cref{sec:setup} \\
$\ch^n=\cg^n-\cg$ & Step-$n$ solution-operator error & \Cref{sec:analysis} \\
$\epsilon_n$ & Error on the realized training distribution $\hat{\nu}_n$ & \Cref{sec:analysis} \\
$e_n$ & Error on the intended distribution $\nu_n$ & \Cref{sec:analysis} \\
$d_n$ & Mismatch $W(\nu_n,\hat{\nu}_n)$ & \Cref{sec:analysis} \\
$\delta_n$ & Step size $W(\nu_n,\nu_{n+1})$ & \Cref{sec:analysis} \\
$C_n,\eta$ & Local-generalization constants and exponent & \Cref{sec:analysis} \\
$R_n$ & Step-dependent trust radius & \Cref{sec:analysis} \\
$K_\cn,K_\ch$ & Lipschitz constants of $\cn$ and $\ch^n$ & \Cref{sec:analysis} \\
\bottomrule
\end{tabular}
\end{table}

\begin{definition}[Mutual singularity]
\label{def:mutual-singularity}
Two probability distributions $\rho$ and $\nu$ on a measurable space are mutually singular, written $\rho\perp\nu$, if there exists a measurable set $A$ such that $\rho(A)=1$ and $\nu(A)=0$.
\end{definition}

\section{Formal statement and proof of \Cref{thm:ks-residual-stability}}
\label{app:ks-stability-proof}

\begingroup

\noindent
This section provides the formal statement and full proof of \Cref{thm:ks-residual-stability}, using the Kuramoto-Sivashinsky (KS) equation as an example to illustrate the instability of PDE loss in controlling solution errors.

We first recall the notations needed for the statement. KS equation reads
\begin{equation}
  u_t+\nu\partial_x^4 u+\partial_x^2 u+u\partial_x u=0, \qquad (t, x)\in(0,T)\times\Omega,
  \label{apdx_eq:ks}
  \end{equation}
where $\nu$ is the viscosity and the smaller it is, the more chaotic the system is. In this section, we consider the case with $\Omega=\T$, i.e. 1-dimensional periodic domain. Analysis for higher-dimensional setting can be carried out similarly. 

We denote the linear part of the KS operator as $L=-\nu\partial_x^4-\partial_x^2$, with Fourier symbol $\lambda_k=|k|^2-\nu|k|^4$ and maximal growth rate $\kappa_\nu:=\max_k\lambda_k\approx\tfrac1{4\nu}$. We denote $\langle k\rangle=(1+|k|^2)^{1/2}$ and $\phi_N(x)=\cos(Nx)$ is the high-frequency probe mode. Throughout, $A\lesssim B$ means $A\le CB$ for a global constant $C$ independent of $(T,\nu)$ unless a subscript records an explicit dependence (e.g.\ $\lesssim_u$, which means $C$ depends on the true solution $u$).

\subsection{Formal statement for \Cref{thm:ks-residual-stability}}
\label{sec:statement}
Throughout the proof, we abbreviate $L^q_tH^\sigma_x:=L^q(0,T;H^\sigma(\Omega))$ when used as a subscript for notational simplicity.
\begin{theorem}
\label{apdx_thm:ks-instability}
Let $\Omega=\T$, let $u$ be a fixed strong solution of \cref{apdx_eq:ks} on $[0,T]$ satisfying $u\in L^\infty(0,T;H^m(\Omega))$ for some $m>\tfrac d2+\alpha+1$, where $d=1$ is the spatial dimensionality of the domain $\Omega$. For a trial solution $w(t, x)$, write $\R(w)=w_t+\nu\partial_x^4 w+\partial_x^2 w+w\partial_x w$, and define the best stability constants $C^{(2)}_{\alpha,r,\beta}(u)$, $C^{(\infty)}_{\alpha,r,\beta}(u)$ as the smallest constants for which
\begin{equation}
\|w-u\|_{X} \;\le\; C\Big(\|w(0)-u(0)\|_{H^r}+\|\R(w)\|_{L^2_tH^\alpha_x}\Big)
\qquad \text{for all admissible } w,
\label{apdx_eq:stability}
\end{equation}
holds with $X=L^2_tH^\beta_x$ (resp.\ $X=L^\infty_tH^\beta_x$). Here we use the convention $u(0):=u\big{|}_{t=0}$.

\medskip
\noindent\textbf{(i) Non-stability thresholds.} The following four statements hold:

\begin{enumerate}\renewcommand{\labelenumi}{\textbf{(\alph{enumi})}}
\item \textbf{Approximate initial conditions (IC), $L^\infty_tH^\beta_x$ target.} If $r<\beta$ \emph{and}
$\alpha<\min\{\beta+1,{2\beta+1}\}$, there is a sequence $w_N$ with
$\|w_N(0)-u(0)\|_{H^r}\to0$, $\|\R(w_N)\|_{L^2_tH^\alpha_x}\to0$, but
$\|w_N-u\|_{L^\infty_tH^\beta_x}\ge c>0$. In other word, $C^{(\infty)}_{\alpha,r,\beta}=\infty$.

\item \textbf{Approximate IC, $L^2_tH^\beta_x$ target.} If $r<\beta-2$ and
$\alpha<\min\{\beta-1,{2\beta-3}\}$, the analogous counterexample holds for
$\|\cdot\|_{L^2_tH^\beta_x}$ and therefore $C^{(2)}_{\alpha,r,\beta}=\infty$. 

\item \textbf{Exact IC $(w_N(0)=u(0))$, $L^\infty_tH^\beta_x$ target.} If $\alpha<\min\{\beta-2,{2\beta+1}\}$, there there exits a sequence $w_N$ with $w_N(0)=u(0)$, $\|\R(w_N)\|_{L^2_tH^\alpha_x}\to0$, but $\|w_N-u\|_{L^\infty_tH^\beta_x}\ge c>0$.

\item \textbf{Exact IC, $L^2_tH^\beta_x$ target.} If $\alpha<\min\{\beta-4,{2\beta-3}\}$, then there exists a sequence $w_N$ with $w_N(0)=u(0)$, $\|\R(w_N)\|_{L^2_tH^\alpha_x}\to0$, but $\|w_N-u\|_{L^2_tH^\beta_x}\ge c>0$. 
\end{enumerate}

\medskip
\noindent\textbf{(ii) Exponential lower bound in the stable regime.} Suppose $\nu$ is small enough such that the lattice $\Z$ contains $k_*$ with $\lambda_{k_*}=\kappa_\nu:=\max_k(|k|^2-\nu|k|^4)>0$ and $|k_*|\approx\nu^{-1/2}$ (so $\kappa_\nu\approx\tfrac1{4\nu}$). Then for every choice of indices for which $C^{(2)}_{\alpha,r,\beta}$, $C^{(\infty)}_{\alpha,r,\beta}$ are finite,
\begin{equation}
\begin{aligned}
C^{(2)}_{\alpha,r,\beta} &\;\gtrsim\; 
 \operatorname{poly}(\nu)\times\exp\Big(\frac{T}{4\nu}\Big), \\
C^{(\infty)}_{\alpha,r,\beta} &\;\gtrsim\; 
 \operatorname{poly}(\nu)\times\exp\Big(\frac{T}{4\nu}\Big).
\end{aligned}
\label{apdx_eq:exp-lower}
\end{equation}
\end{theorem}

\begin{remark}[Background regularity]
\label{assum:regularity} We assume $u\in L^\infty(0,T;H^m(\T))$ for some $m>\tfrac d2+\alpha+1$, so that $H^m(\T)\hookrightarrow W^{\alpha+1,\infty}(\T)$ and multiplication by $u(t,\cdot)$ or $u_x(t,\cdot)$ is bounded on $H^\alpha$, uniformly in $t\in[0,T]$, with constant $C_u$. This is guranteed by the well-posedness theory of KS equation as long as the initial condition is sufficiently regular.
\end{remark}

\subsection{Auxiliary results}
\label{sec:strategy}

The proof uses the following Fourier and time-amplitude estimates.

\begin{lemma}[Probe-mode Fourier estimates]
\label{lem:probe}
Let $\phi_N(x)=\cos(Nx)$. For every $\sigma\in\mathbb{R}$,
\begin{equation}
\|\phi_N\|_{H^\sigma(\T)}\sim N^\sigma, \qquad
\|\partial_x\phi_N\|_{H^\sigma}\sim N^{\sigma+1}, \qquad
\|\phi_N\partial_x\phi_N\|_{H^\sigma}\sim N^{\sigma+1}.
\label{apdx_eq:lemma1}
\end{equation}
\end{lemma}

\begin{proof}
$\phi_N$ has Fourier support at $k=\pm N$, so
\begin{equation}
\|\phi_N\|_{H^\sigma}^2=\sum_k(1+|k|^2)^\sigma|\widehat\phi_N(k)|^2\sim(1+N^2)^\sigma\sim N^{2\sigma},
\end{equation}
giving the first estimate. The second follows from $\partial_x\phi_N=-N\sin(Nx)$,
whose Fourier support is unchanged. For the third, $\phi_N\partial_x\phi_N=-\tfrac
N2\sin(2Nx)$ is supported at $k=\pm2N$, so
$\|\phi_N\partial_x\phi_N\|_{H^\sigma}\sim N\cdot(2N)^\sigma\sim N^{\sigma+1}$.
\end{proof}

\begin{lemma}[Time-amplitude functions]
\label{lem:time-amplitude}
For $\lambda<0$, $T>0$, set $I_T(\lambda)=\big(\int_0^Te^{2\lambda t}\,dt\big)^{1/2}$ and let
$V_{\lambda,T}g(t)=\int_0^te^{\lambda(t-s)}g(s)\,ds$. Then
\begin{equation}
I_T(\lambda)\sim |\lambda|^{-1/2}, \qquad
\|V_{\lambda,T}\|_{L^2\to L^\infty}:=J_T(\lambda)\sim|\lambda|^{-1/2}, \qquad
\|V_{\lambda,T}\|_{L^2\to L^2}\sim |\lambda|^{-1}
\label{apdx_eq:lemma2}
\end{equation}
whenever $|\lambda|T\gg1$ (so the finite-horizon corrections are negligible); for
$\lambda>0$ with $\lambda T\gg1$, $I_T(\lambda)\sim |\lambda|^{-1/2}e^{\lambda T}$ and
$J_T(\lambda)\sim|\lambda|^{-1/2}e^{\lambda T}$,
$\|V_{\lambda,T}\|_{L^2\to L^2}\gtrsim \lambda^{-1}e^{\lambda T}$.
\end{lemma}

\begin{proof}
Direct evaluation of the Gaussian-type integrals
$\int_0^Te^{2\lambda t}\,dt=\frac{1-e^{2\lambda T}}{-2\lambda}$ (for $\lambda<0$) or
$\frac{e^{2\lambda T}-1}{2\lambda}$ (for $\lambda>0$), and Young's convolution inequality
$\|V_{\lambda,T}g\|_{L^\infty}\le\|e^{\lambda t}\mathbf 1_{t\ge0}\|_{L^2}\|g\|_{L^2}$,
$\|V_{\lambda,T}g\|_{L^2}\le\|e^{\lambda t}\mathbf 1_{t\ge 0}\|_{L^1}\|g\|_{L^2}$ for the
operator-norm bounds.
\end{proof}

\subsection{Proof of \Cref{apdx_thm:ks-instability}}
\label{sec:proof}

\subsubsection{Proof of part (i)}
\begin{proof}
We construct $z_N=w_N-u$ explicitly. The proof first derives the residual identity, then treats approximate and exact initial conditions separately.

\medskip\noindent\textbf{Step 1. Residual identity.}
For $w_N=u+z_N$ with $z_N$ solving the \emph{linear} KS equation $\partial_tz_N=Lz_N$ exactly (so $\partial_tz_N-Lz_N=0$) and $u$ is a solution to the KS equation (so $\R(u)=0$), expanding $\R(u+z_N)$ and cancelling the linear parts gives the key identity
\begin{equation}
\R(w_N) = u\,\partial_x z_N + z_N u_x + z_N\partial_x z_N.
\label{apdx_eq:residual-identity}
\end{equation}
If instead $z_N$ solves a forced linear equation $\partial_tz_N-Lz_N=f_N$, then $f_N$ appears additively in \eqref{apdx_eq:residual-identity}.

\medskip\noindent\textbf{Step 2. Approximate initial conditions.}
Let $z_N(t,x)=A_Ne^{\lambda_Nt}\phi_N(x)$ with $\lambda_N=N^2-\nu N^4\to-\infty$, where $A_N$ will be determined later in the proof. By
\Cref{lem:time-amplitude} and the fact of $|\lambda_N|\sim\nu N^4$, we have
$\|e^{\lambda_Nt}\|_{L^2(0,T)}\sim N^{-2}$. Combining this with the multiplier bound indicated by the regulairity assumption of $u$ in
\Cref{assum:regularity} ($\|u(t)\eta\|_{H^\alpha}\le C_u\|\eta\|_{H^\alpha}$ for any $\eta\in H^\alpha$) and
\Cref{lem:probe} term-by-term in \eqref{apdx_eq:residual-identity} gives
\begin{equation}
\|\R(w_N)\|_{L^2_tH^\alpha_x} \;\lesssim_u\; A_N N^{\alpha-1} + A_N^2 N^{\alpha-1}.
\label{apdx_eq:residual-bound-weak}
\end{equation}
Also $\|z_N(0)\|_{H^r}\sim A_NN^r$ and $\|z_N\|_{L^\infty_tH^\beta_x}\sim A_NN^\beta$, $\|z_N\|_{L^2_tH^\beta_x}\sim A_N N^{\beta-2}$ (the last using $\|e^{\lambda_N t}\|_{L^2_t}\sim N^{-2}$).

\smallskip
\noindent\emph{Claim (a).} Choose $A_N=N^{-\beta}$, normalizing
$\|z_N\|_{L^\infty_tH^\beta_x}\sim1$. Then 
\begin{equation}
\|z_N(0)\|_{H^r}\sim N^{r-\beta}\to0 \iff r<\beta,
\end{equation}
 and substituting into \eqref{apdx_eq:residual-bound-weak} gives
 \begin{equation}
\|\R(w_N)\|_{L^2_tH^\alpha_x}\lesssim_u N^{\alpha-1-\beta}+N^{\alpha-1-2\beta}\to 0 \iff
\beta>\max\{\alpha-1,\tfrac{\alpha-1}2\}.
 \end{equation}
This proves part~(a).

\smallskip
\noindent\emph{Claim (b).} Choose instead $A_N=N^{2-\beta}$, normalizing
$\|z_N\|_{L^2_tH^\beta_x}\sim1$. Then 
\begin{equation}
\|z_N(0)\|_{H^r}\sim N^{r+2-\beta}\to0\iff r<\beta-2,
\end{equation}
and \eqref{apdx_eq:residual-bound-weak} becomes $\lesssim_u N^{\alpha+1-\beta}+N^{\alpha+3-2\beta}\to0
\iff \beta>\max\{\alpha+1,\tfrac{\alpha+3}2\}$. This proves part~(b).

\medskip\noindent\textbf{Step 3. Exact initial conditions.}
To enforce $w_N(0)=u(0)$ we replace the free decaying mode by a forced short pulse. Fix a smooth function
$\chi\in C^\infty_c((0,1))$ with $\|\chi\|_{L^2(0,1)}=1$ and $t_0\in(0,T)$, and set
$\chi_N(t)=N^2\chi(N^4(t-t_0))$, so $\|\chi_N\|_{L^2(0,T)}=1$ (change of variables
$s=N^4(t-t_0)$) and $\chi_N$ is supported on an interval of length $N^{-4}$. 

Define $q_N$ as the solution to the ODE 
\begin{equation}\label{apdx_ODE}
\frac{dq_N}{dt}=\lambda_Nq_N+B_N\chi_N, \qquad q_N(0)=0,
\end{equation}
where $B_N$ is a free amplitude parameter to be determined later. Then set $z_N=q_N(t)\phi_N(x)$, $w_N=u+z_N$; then $z_N(0)=0$ so $w_N(0)=u(0)$, and
\eqref{apdx_eq:residual-identity} acquires the extra forcing term $B_N\chi_N\phi_N$.

In fact, we can explicitly solve \eqref{apdx_ODE} to get $q_N=B_NV_{\lambda_N,T}\chi_N$. By
\Cref{lem:time-amplitude} (with $|\lambda_N|\sim\nu N^4$), we have
\begin{align}
  \norm{q_N}_{L^\infty_t} &\sim B_N N^{-2}, \label{apdx_eq:qLinf}\\
  \norm{q_N}_{L^2_t}      &\sim B_N N^{-4}, \label{apdx_eq:qL2}\\
  \norm{q_N^2}_{L^2_t}    &\lesssim \norm{q_N}_{L^\infty_t}   \norm{q_N}_{L^2_t}  \sim B_N^2 N^{-6}, \label{apdx_eq:qsq}
\end{align}
where the second estimate uses $\|e^{\lambda_Nt}\mathbf 1_{t\ge0}\|_{L^1}=|\lambda_N|^{-1}$ and Young's inequality. The matching lower bound follows because $\chi_N$ is supported on the time scale $|\lambda_N|^{-1}\sim N^{-4}$.

\medskip\noindent\textbf{Part A. The $L^2$ estimate for $q_N$.}

\smallskip\noindent\emph{Upper bound via Young's convolution inequality}\par

Write $V_{\lambda,T}g = h_\lambda * g\big|_{(0,T)}$ where
$h_\lambda(\tau) = e^{\lambda\tau}\mathbf{1}_{\tau\ge0}$.
For $\lambda = \lambda_N < 0$,
\begin{equation}
  \norm{h_{\lambda_N}}_{L^1(0,\infty)}
  = \int_0^\infty e^{\lambda_N \tau}\,d\tau
  = \frac{1}{|\lambda_N|}.
  \label{apdx_eq:h-L1}
\end{equation}
Young's convolution inequality $\norm{f*g}_{L^2}\le\norm{f}_{L^1}\norm{g}_{L^2}$ gives
\begin{equation}
  \norm{V_{\lambda_N,T}g}_{L^2_t}
  \;\le\; \norm{h_{\lambda_N}}_{L^1}\,\norm{g}_{L^2_t}
  \;=\; \frac{1}{|\lambda_N|}\,\norm{g}_{L^2_t}.
  \label{apdx_eq:Young}
\end{equation}
Applying \eqref{apdx_eq:Young} to $g = \chi_N$ and using
$\norm{\chi_N}_{L^2_t}=1$ and $|\lambda_N|\sim\nu N^4$:
\begin{equation}
  \norm{q_N}_{L^2_t}
  = B_N\norm{V_{\lambda_N,T}\chi_N}_{L^2_t}
  \;\le\; \frac{B_N}{|\lambda_N|}
  \;\lesssim\; B_N N^{-4}.
  \label{apdx_eq:qL2-upper}
\end{equation}

\smallskip\noindent\emph{Matching lower bound via concentration of $\chi_N$}\par

The key geometric fact is:

\smallskip\noindent\emph{Claim.} $e^{\lambda_N(t-s)}\ge e^{-\nu}$ whenever $t_0\le s\le t\le t_0+N^{-4}$.\par
\noindent\emph{Proof of the claim.}
  On $[t_0,t_0+N^{-4}]$ we have $0\le t-s\le N^{-4}$, so
  $\lambda_N(t-s)\ge\lambda_N\cdot N^{-4} = (N^2-\nu N^4)N^{-4}
  = N^{-2}-\nu \ge -\nu$ for $N\ge1$.
  Hence $e^{\lambda_N(t-s)}\ge e^{-\nu}$. \hfill$\square$

\noindent Now we obtain a lower bound for $q_N$. At $t_1:=t_0+N^{-4}$ (i.e.\ at the end of the support of $\chi_N$), since $e^{\lambda_N(t_1-s)}\ge e^{-\nu}$ on $\operatorname{supp}(\chi_N)$,
\begin{equation}
  q_N(t_1)
  = B_N\int_0^{t_1} e^{\lambda_N(t_1-s)}\chi_N(s)\,ds
  \;\ge\; B_N\,e^{-\nu}\int_{t_0}^{t_0+N^{-4}}\chi_N(s)\,ds.
  \label{apdx_eq:qN-lb-pointwise}
\end{equation}
Compute the integral of $\chi_N$: by the substitution $\sigma = N^4(s-t_0)$,
\begin{equation}
  \int_{t_0}^{t_0+N^{-4}}\chi_N(s)\,ds
  = N^2\int_0^1 \chi(\sigma)\,N^{-4}\,d\sigma
  = N^{-2}\norm{\chi}_{L^1(0,1)}
  =: c_\chi N^{-2},
  \label{apdx_eq:chi-L1}
\end{equation}
where $c_\chi = \norm{\chi}_{L^1} > 0$ is a fixed positive constant.
Thus at $t_1=t_0+N^{-4}$:
\begin{equation}
  q_N(t_1) \;\ge\; c_\chi\,e^{-\nu}\,B_N N^{-2}\,=:\,c_0 B_N N^{-2}.
  \label{apdx_eq:qN-amplitude}
\end{equation}
After the pulse, $\chi_N\equiv0$, so $q_N$ satisfies the
homogeneous ODE $q_N' = \lambda_N q_N$, giving
$q_N(t) = q_N(t_1)\,e^{\lambda_N(t-t_1)}$ for $t\ge t_1$.
The $L^2$ norm of this decaying tail:
\begin{equation}
  \norm{q_N}_{L^2(t_1,T)}^2
  \;\ge\; q_N(t_1)^2\int_{t_1}^T e^{2\lambda_N(t-t_1)}\,dt
  \;=\; q_N(t_1)^2\,\frac{1-e^{2\lambda_N(T-t_1)}}{2|\lambda_N|}
  \;\ge\; q_N(t_1)^2\,\frac{1-e^{-2}}{2|\lambda_N|},
  \label{apdx_eq:tail-L2}
\end{equation}
where the last step uses $|\lambda_N|(T-t_1)\ge|\lambda_N|(T-t_0-N^{-4})\to\infty$
(so $e^{2\lambda_N(T-t_1)}\to0$) and we used the crude bound $\ge1-e^{-2}$ for
large $N$. Substituting $q_N(t_1)\ge c_0 B_N N^{-2}$ and $|\lambda_N|\sim\nu N^4$:
\begin{equation}
  \norm{q_N}_{L^2_t}
  \;\ge\; \norm{q_N}_{L^2(t_1,T)}
  \;\gtrsim\; \frac{c_0 B_N N^{-2}}{|\lambda_N|^{1/2}}
  \;\sim\; B_N N^{-2}\cdot N^{-2}
  \;=\; B_N N^{-4}.
  \label{apdx_eq:qL2-lower}
\end{equation}
Combining \eqref{apdx_eq:qL2-upper} and \eqref{apdx_eq:qL2-lower} gives
$\norm{q_N}_{L^2_t}\sim B_N N^{-4}$, confirming \eqref{apdx_eq:qL2}. 

\medskip\noindent\textbf{Part B. Forced residual estimate.}

Since $q_N$ satisfies the \emph{forced} ODE \cref{apdx_ODE} , we have $\partial_t z_N - Lz_N = B_N\chi_N\phi_N$, and expanding
$\R(w_N) = \R(u+z_N)$ and using $\R(u)=0$ gives
\begin{equation}
  \R(w_N)
  = \underbrace{B_N\chi_N\phi_N}_{\text{(I) forcing}}
  + \underbrace{u\,\partial_x z_N}_{\text{(II) cross: }u\cdot z}
  + \underbrace{z_N\,u_x}_{\text{(III) cross: }z\cdot u}
  + \underbrace{z_N\,\partial_x z_N}_{\text{(IV) self-interaction}}.
  \label{apdx_eq:residual-decomp}
\end{equation}
We bound the $\ltH$ norm of each term.

\smallskip\noindent\emph{Term (I): forcing term $B_N\chi_N(t)\phi_N(x)$}\par

Since $\phi_N$ depends only on $x$ and $\chi_N$ depends only on $t$, the space and time
factors separate:
\begin{align}
  \norm{B_N\chi_N\phi_N}_{\ltH}^2
  &= B_N^2\int_0^T |\chi_N(t)|^2\,\norm{\phi_N}_{H^\alpha}^2\,dt
   = B_N^2\,\norm{\chi_N}_{L^2_t}^2\,\norm{\phi_N}_{H^\alpha}^2.
  \label{apdx_eq:term1-calc}
\end{align}
Using $\norm{\chi_N}_{L^2_t}=1$:
\begin{equation}
  \norm{B_N\chi_N\phi_N}_{\ltH}
  = B_N\,\norm{\phi_N}_{H^\alpha}
  \sim B_N N^\alpha.
  \label{apdx_eq:term1}
\end{equation}

\smallskip\noindent\emph{Term (II): cross term $u\,\partial_x z_N$}\par

Write $u\,\partial_x z_N = q_N(t)\cdot u(t,x)\,\partial_x\phi_N(x)$.
By the background regularity assumption, multiplication by $u(t,\cdot)$
is bounded on $H^\alpha$ with constant $C_u$, uniformly in $t$:
\begin{equation}
  \norm{u(t)\,\partial_x\phi_N}_{H^\alpha}
  \;\le\; C_u\,\norm{\partial_x\phi_N}_{H^\alpha}
  \; {\sim}\; C_u N^{\alpha+1}.
  \label{apdx_eq:term2-pointwise}
\end{equation}
Therefore, separating time and space:
\begin{equation}
  \norm{u\,\partial_x z_N}_{\ltH}^2
  = \int_0^T |q_N(t)|^2\,\norm{u(t)\,\partial_x\phi_N}_{H^\alpha}^2\,dt
  \;\le\; C_u^2 N^{2(\alpha+1)}\,\norm{q_N}_{L^2_t}^2.
  \label{apdx_eq:term2-calc}
\end{equation}
Taking the square root and using \eqref{apdx_eq:qL2}:
\begin{equation}
  \norm{u\,\partial_x z_N}_{\ltH}
  \;\lesssim_u\; N^{\alpha+1}\cdot B_N N^{-4}
  = B_N N^{\alpha-3}.
  \label{apdx_eq:term2}
\end{equation}

\smallskip\noindent\emph{Term (III): cross term $z_N\,u_x$}\par

Write $z_N u_x = q_N(t)\cdot\phi_N(x)\,u_x(t,x)$.
Since $u_x(t,\cdot)\in H^{m-1}$ with $m-1 > \tfrac{d}{2}+\alpha$ (from the
regularity assumption $m>\tfrac{d}{2}+\alpha+1$), $u_x(t,\cdot)$ is a Sobolev
multiplier on $H^\alpha$ with constant $C_u$:
\begin{equation}
  \norm{\phi_N\,u_x(t)}_{H^\alpha}
  \;\le\; C_u\,\norm{\phi_N}_{H^\alpha}
  \; {\sim}\; C_u N^\alpha.
  \label{apdx_eq:term3-pointwise}
\end{equation}
Then:
\begin{equation}
  \norm{z_N u_x}_{\ltH}^2
  = \int_0^T |q_N(t)|^2\,\norm{\phi_N u_x(t)}_{H^\alpha}^2\,dt
  \;\le\; C_u^2 N^{2\alpha}\,\norm{q_N}_{L^2_t}^2,
\end{equation}
so:
\begin{equation}
  \norm{z_N u_x}_{\ltH}
  \;\lesssim_u\; N^\alpha\cdot B_N N^{-4}
  = B_N N^{\alpha-4}.
  \label{apdx_eq:term3}
\end{equation}
This is smaller than term~(II) by a factor of $N^{-1}$.

\smallskip\noindent\emph{Term (IV): self-interaction $z_N\,\partial_x z_N$}\par

Write $z_N\,\partial_x z_N = q_N(t)^2\cdot\phi_N(x)\,\partial_x\phi_N(x)$.
The spatial factor is now $\phi_N\partial_x\phi_N$, estimated by \Cref{lem:probe}:
$\norm{\phi_N\partial_x\phi_N}_{H^\alpha}\sim N^{\alpha+1}$.
Separating:
\begin{equation}
  \norm{z_N\partial_x z_N}_{\ltH}^2
  = \int_0^T |q_N(t)|^4\,\norm{\phi_N\partial_x\phi_N}_{H^\alpha}^2\,dt
  = \norm{q_N^2}_{L^2_t}^2\cdot\norm{\phi_N\partial_x\phi_N}_{H^\alpha}^2.
  \label{apdx_eq:term4-calc}
\end{equation}
Taking the square root and using \eqref{apdx_eq:qsq} and \Cref{lem:probe}:
\begin{equation}
  \norm{z_N\partial_x z_N}_{\ltH}
  \;\le\; \norm{q_N^2}_{L^2_t}\cdot\norm{\phi_N\partial_x\phi_N}_{H^\alpha}
  \;\lesssim\; B_N^2 N^{-6}\cdot N^{\alpha+1}
  = B_N^2 N^{\alpha-5}.
  \label{apdx_eq:term4}
\end{equation}

\medskip\noindent\textbf{Part C. Combination of the residual terms.}

Collect \eqref{apdx_eq:term1}--\eqref{apdx_eq:term4}:

\begin{center}
\renewcommand{\arraystretch}{1.4}
\begin{tabular}{l|c|l}
\hline
Term & $\norm{\cdot}_{\ltH}$ & Rate \\
\hline
(I)  forcing $B_N\chi_N\phi_N$   & $\sim B_N N^\alpha$   & leading \\
(II) $u\,\partial_x z_N$       & $\lesssim_u B_N N^{\alpha-3}$ & lower by $N^{-3}$ \\
(III) $z_N u_x$               & $\lesssim_u B_N N^{\alpha-4}$ & lower by $N^{-4}$ \\
(IV) $z_N\partial_x z_N$       & $\lesssim B_N^2 N^{\alpha-5}$ & different $B_N$-power \\
\hline
\end{tabular}
\end{center}

\bigskip
By the triangle inequality applied to \eqref{apdx_eq:residual-decomp}:
\begin{align}
  \norm{\R(w_N)}_{\ltH}
  &\lesssim_u B_N N^\alpha
    + B_N N^{\alpha-3}
    + B_N N^{\alpha-4}
    + B_N^2 N^{\alpha-5}
  \label{apdx_eq:sum-four}\\
  &\lesssim_u B_N N^\alpha + B_N^2 N^{\alpha-5}.
  \label{apdx_eq:residual-bound-combined}
\end{align}

\smallskip
\noindent\emph{Case (c).} Note that
$\|z_N\|_{L^\infty_tH^\beta_x}\sim \|q_N\|_{L^\infty_t}\|\phi_N\|_{H^\beta}\sim B_N
N^{\beta-2}$, choosing $B_N=N^{2-\beta}$ normalizes this to $\sim1$. Then
\eqref{apdx_eq:residual-bound-combined} reads $\lesssim_u N^{\alpha+2-\beta}+N^{\alpha-1-2\beta}\to0
\iff\beta>\max\{\alpha+2,\tfrac{\alpha-1}2\}$. This proves part~(c).

\smallskip
\noindent\emph{Case (d).} Note that
$\|z_N\|_{L^2_tH^\beta_x}\sim \|q_N\|_{L^2_t}\|\phi_N\|_{H^\beta}\sim B_NN^{\beta-4}$;
choosing $B_N=N^{4-\beta}$ normalizes this to $\sim1$. Then
\eqref{apdx_eq:residual-bound-combined} reads $\lesssim_u N^{\alpha+4-\beta}+N^{\alpha+3-2\beta}\to0
\iff\beta>\max\{\alpha+4,\tfrac{\alpha+3}2\}$. This proves part~(d).

This concludes the proof of all four claims in \Cref{apdx_thm:ks-instability} (i).
\end{proof}

\subsubsection{Proof of part (ii)}

\begin{proof}

\paragraph{Notation.}
Throughout, $u$ is the fixed strong solution of the KS equation
on $[0,T]$ (\Cref{assum:regularity}), and
\[
  \kap \;:=\; \max_{k\in\mathbb{Z}}(|k|^2-\nu|k|^4)\;\approx\;\frac{1}{4\nu}
\]
is the maximal KS growth rate, attained at the resonant mode $k_*$ with $|k_*|^2\approx\tfrac{1}{2\nu}$,
hence $\jp{k_*}=(1+|k_*|^2)^{1/2}\approx\nu^{-1/2}$.
The operator $V_{\lambda,T}$ and the scalar functions $I_T(\lambda)$,
$J_T(\lambda)$ are as in \Cref{lem:time-amplitude}:
\[
  (V_{\lambda,T}g)(t) = \int_0^t e^{\lambda(t-s)}g(s)\,ds,\quad
  I_T(\lambda)=\Big(\int_0^T e^{2\lambda t}\,dt\Big)^{1/2},\quad
  J_T(\lambda)=\norm{V_{\lambda,T}}_{L^2\to L^\infty}.
\]
The background regularity constant $C_u$ bounds the $H^\alpha$-multiplier
norm of $u(t,\cdot)$ and $u_x(t,\cdot)$, uniformly in $t\in[0,T]$.
The probe mode is $\phi_{k_*}(x)=e^{ik_*x}$ (or its real part),
with $\norm{\phi_{k_*}}_{H^\beta}\sim\jp{k_*}^\beta\approx\nu^{-\beta/2}$.

\medskip\noindent\textbf{Time-integral estimates at the unstable mode}\par

We first compute the three quantities that appear in \eqref{apdx_eq:exp-lower}
when the time-amplitude formulas of \Cref{lem:time-amplitude} are evaluated at
$\lambda=\kap\approx\tfrac{1}{4\nu}>0$ with $\kap T\gg1$.

\smallskip\noindent\emph{(A) The $L^2$-time integral $I_T(\kap)$}\par

\begin{equation}
  I_T(\kap)
  = \Big(\int_0^T e^{2\kap t}\,dt\Big)^{1/2}
  = \bigg(\frac{e^{2\kap T}-1}{2\kap}\bigg)^{\!1/2}
  \sim \frac{e^{\kap T}}{\sqrt{2\kap}}
  \quad(\text{since }e^{2\kap T}\gg1).
  \label{apdx_eq:IT}
\end{equation}
Substituting $\kap\approx\tfrac{1}{4\nu}$:
\begin{equation}
  \frac{1}{\sqrt{2\kap}}\approx\sqrt{2\nu},
  \qquad
  e^{\kap T}\approx e^{T/(4\nu)},
  \qquad
  {I_T(\kap)\sim\sqrt{2\nu}\,e^{T/(4\nu)}.}
  \label{apdx_eq:IT-value}
\end{equation}

\smallskip\noindent\emph{(B) The $L^\infty$-operator norm $J_T(\kap)=\norm{V_{\kap,T}}_{L^2\to L^\infty}$}\par

For any $g\in L^2(0,T)$ and any $t\in[0,T]$, by the Cauchy--Schwarz inequality and the
substitution $s\mapsto t-s$:
\begin{align}
  |(V_{\kap,T}g)(t)|
  &= \Big|\int_0^t e^{\kap(t-s)}g(s)\,ds\Big|
  \;\le\; e^{\kap t}\int_0^t e^{-\kap s}|g(s)|\,ds \notag\\
  &\le\; e^{\kap t}\Big(\int_0^t e^{-2\kap s}\,ds\Big)^{1/2}\norm{g}_{L^2}
  \;\le\; e^{\kap T}\Big(\frac{1-e^{-2\kap T}}{2\kap}\Big)^{1/2}\norm{g}_{L^2}
  \;\sim\; \frac{e^{\kap T}}{\sqrt{2\kap}}\,\norm{g}_{L^2}.
  \label{apdx_eq:JT-upper}
\end{align}
This bound is achieved (up to a constant) by taking $g(s)\propto e^{-\kap s}\mathbf{1}_{[0,T/2]}$,
which satisfies $\norm{g}_{L^2}\sim 1$ (for $\kap T\gg1$)
and gives $(V_{\kap,T}g)(T)\sim e^{\kap T}/(2\kap)^{1/2}\norm{g}_{L^2}$.
Hence
\begin{equation}
  {J_T(\kap)
  = \norm{V_{\kap,T}}_{L^2\to L^\infty}
  \sim \frac{e^{\kap T}}{\sqrt{2\kap}}
  \approx \sqrt{2\nu}\,e^{T/(4\nu)}.}
  \label{apdx_eq:JT-value}
\end{equation}

\smallskip\noindent\emph{(C) The $L^2\to L^2$-operator norm $\norm{V_{\kap,T}}_{L^2\to L^2}$}\par

\paragraph{Lower bound.}
Take the test function $g_0(s)=c_0\,e^{-\kap s}\mathbf{1}_{[0,T/2]}$, normalised
so that $\norm{g_0}_{L^2}=1$:
\begin{equation}
  c_0^2\int_0^{T/2}e^{-2\kap s}\,ds = c_0^2\,\frac{1-e^{-\kap T}}{2\kap}\approx\frac{c_0^2}{2\kap},
  \quad\Rightarrow\quad c_0\approx\sqrt{2\kap}.
  \label{apdx_eq:g0-norm}
\end{equation}
For $t>T/2$, the integrand is entirely in the past of the support of $g_0$, so:
\begin{align}
  (V_{\kap,T}g_0)(t)
  &= c_0\int_0^{T/2}e^{\kap(t-s)}e^{-\kap s}\,ds
  = c_0\,e^{\kap t}\int_0^{T/2}e^{-2\kap s}\,ds \\
 & \approx c_0\,e^{\kap t}\cdot\frac{1}{2\kap}
  = \frac{c_0}{2\kap}\,e^{\kap t}
  \approx \frac{e^{\kap t}}{\sqrt{2\kap}}.
  \label{apdx_eq:Vg0}
\end{align}
Therefore:
\begin{equation}
  \norm{V_{\kap,T}g_0}_{L^2}^2
  \;\ge\; \int_{T/2}^T |(V_{\kap,T}g_0)(t)|^2\,dt
  \;\approx\; \frac{1}{2\kap}\int_{T/2}^T e^{2\kap t}\,dt
  \;\approx\; \frac{e^{2\kap T}}{4\kap^2}.
  \label{apdx_eq:V-L2-lb-sq}
\end{equation}
Since $\norm{g_0}_{L^2}=1$, this gives:
\begin{equation}
  \norm{V_{\kap,T}}_{L^2\to L^2}
  \;\ge\; \norm{V_{\kap,T}g_0}_{L^2}
  \;\gtrsim\; \frac{e^{\kap T}}{2\kap}
  \;\approx\; 2\nu\,e^{T/(4\nu)}.
  \label{apdx_eq:V-L2-lb}
\end{equation}

\paragraph{Upper bound (Hilbert--Schmidt).}
\begin{equation}
  \norm{V_{\kap,T}}_{L^2\to L^2}^2
  \le \norm{V_{\kap,T}}_{\mathrm{HS}}^2
  = \int_0^T\!\int_0^t e^{2\kap(t-s)}\,ds\,dt
  = \int_0^T\frac{e^{2\kap t}-1}{2\kap}\,dt
  \sim \frac{e^{2\kap T}}{4\kap^2},
  \label{apdx_eq:V-HS}
\end{equation}
giving $\norm{V_{\kap,T}}_{L^2\to L^2}\lesssim e^{\kap T}/(2\kap)\approx 2\nu\,e^{T/(4\nu)}$.

Combining the two sides:
\begin{equation}
  {\norm{V_{\kap,T}}_{L^2\to L^2}\;\sim\;\frac{e^{\kap T}}{2\kap}\;\approx\;2\nu\,e^{T/(4\nu)}.}
  \label{apdx_eq:V-L2-value}
\end{equation}

\medskip\noindent\textbf{Summary of the three estimates.}
\begin{equation}
  I_T(\kap)\sim\sqrt{2\nu}\,e^{T/(4\nu)},
  \quad
  J_T(\kap)\sim\sqrt{2\nu}\,e^{T/(4\nu)},
  \quad
  \norm{V_{\kap,T}}_{L^2\to L^2}\sim 2\nu\,e^{T/(4\nu)}.
  \label{apdx_eq:three-estimates}
\end{equation}

\medskip\noindent\textbf{Converting $\jp{k_*}$ to powers of $\nu$}\par

The resonant mode $k_*\in\mathbb{Z}$ (as specified in \Cref{apdx_thm:ks-instability}) maximises $\lambda_k=|k|^2-\nu|k|^4$,
which is maximised at $|k|^2=\tfrac{1}{2\nu}$.  Hence $\jp{k_*}^2=1+|k_*|^2\approx\nu^{-1}$
for $\nu$ small, and:
\begin{equation}
  \jp{k_*}\;\approx\;\nu^{-1/2},
  \qquad
  \jp{k_*}^\gamma\;\approx\;\nu^{-\gamma/2}
  \quad\text{for any }\gamma\in\mathbb{R}.
  \label{apdx_eq:kstar-power}
\end{equation}
The two particular combinations that appear are:
\begin{align}
  \jp{k_*}^{\beta-r} &\approx \nu^{(r-\beta)/2},
  \label{apdx_eq:kstar-IC}\\
  \jp{k_*}^{\beta-\alpha} &\approx \nu^{(\alpha-\beta)/2}.
  \label{apdx_eq:kstar-res}
\end{align}

\medskip\noindent\textbf{The IC-driven lower bounds}\par

\noindent\textbf{Construction of the perturbative solution.}
Take the initial condition $w(0) = u(0) + a\phi_{k_*}$ for a small amplitude $a>0$, and let
$w$ be the exact solution of the KS equation with this initial condition.
By construction $\mathcal{R}(w)=0$.  The difference $z:=w-u$ satisfies the
\emph{linearised} KS equation around~$u$:
\begin{equation}
  \partial_t z = Lz - u\partial_x z - z\partial_x u + O(z^2),
  \label{apdx_eq:lin-KS}
\end{equation}
where $L=-\nu\partial_x^4-\partial_x^2$ has symbol $\lambda_k=|k|^2-\nu|k|^4$.
The linearised operator $\mathcal{L}_u=L-u\partial_x-(\partial_x u)\cdot$
differs from $L$ by the bounded perturbation $P:=-u\partial_x-(\partial_x u)\cdot$,
whose operator norm on $H^\beta$ is bounded by $C_u$ uniformly in $t$.
By Duhamel's principle, the growth rate of the mode $\phi_{k_*}$ under $\mathcal{L}_u$
differs from $\lambda_{k_*}=\kap$ by at most $O(C_u)$; in particular, for
$\kap\gg C_u$ (i.e.\ $\nu$ sufficiently small), the solution satisfies
\begin{equation}
  \norm{z(t)}_{H^\beta}
  \;\ge\; a\jp{k_*}^\beta e^{(\kap-O(C_u))t},
  \qquad t\in[0,T].
  \label{apdx_eq:z-lower-pointwise}
\end{equation}

\noindent\textbf{Applying the stability inequality.}
Since $\mathcal{R}(w)=0$, the stability inequality reads:
\begin{equation}
  \norm{w-u}_X \;\le\; C\norm{w(0)-u(0)}_{H^r}
  = Ca\jp{k_*}^r.
  \label{apdx_eq:stab-IC}
\end{equation}

\paragraph{Case $X=L^\infty_t H^\beta_x$.}
From \eqref{apdx_eq:z-lower-pointwise} at $t=T$ (where $e^{\kap T}\approx e^{T/(4\nu)}$ dominates):
\begin{equation}
  \norm{w-u}_{L^\infty_t H^\beta}
  \;\ge\; \norm{z(T)}_{H^\beta}
  \;\gtrsim\; a\jp{k_*}^\beta e^{(\kap-O(C_u))T}.
  \label{apdx_eq:error-Linf}
\end{equation}
Substituting \eqref{apdx_eq:error-Linf} and \eqref{apdx_eq:stab-IC} and cancelling $a$:
\begin{equation}
  \jp{k_*}^\beta e^{\kap T}
  \;\lesssim\; C\jp{k_*}^r
  \quad\Longrightarrow\quad
  C \;\gtrsim\; \jp{k_*}^{\beta-r}e^{\kap T}
  \;\stackrel{\eqref{apdx_eq:kstar-IC},\,\eqref{apdx_eq:IT-value}}{\approx}\;
  \nu^{(r-\beta)/2}\,e^{T/(4\nu)}.
  \label{apdx_eq:IC-Linf}
\end{equation}

\paragraph{Case $X=L^2_t H^\beta_x$.}
Integrating \eqref{apdx_eq:z-lower-pointwise} in time:
\begin{align}
  \norm{w-u}_{L^2_t H^\beta}^2
  &\;\ge\; a^2\jp{k_*}^{2\beta}\int_0^T e^{2(\kap-O(C_u))t}\,dt
  \;=\; a^2\jp{k_*}^{2\beta}\,I_T(\kap-O(C_u))^2.
  \label{apdx_eq:error-L2}
\end{align}
Taking square roots and substituting into the stability inequality:
\begin{equation}
  C \;\gtrsim\; \frac{\jp{k_*}^\beta I_T(\kap-O(C_u))}{\jp{k_*}^r}
  = \jp{k_*}^{\beta-r}I_T(\kap)
  \;\stackrel{\eqref{apdx_eq:kstar-IC},\,\eqref{apdx_eq:IT-value}}{\approx}\;
  \nu^{(r-\beta)/2}\cdot\sqrt{2\nu}\,e^{T/(4\nu)}
  = \nu^{(r-\beta+1)/2}\,e^{T/(4\nu)}.
  \label{apdx_eq:IC-L2}
\end{equation}

\end{proof}

\endgroup

\section{Proof of \Cref{thm:pde-loss-variance}}
\label{app:pino-variance-proof}

\subsection{Formal statement for \Cref{thm:pde-loss-variance}}

Let $\{\phi_k\}_{k\in\mathbb Z^d}$ be the orthonormal Fourier basis of $L^2(\mathbb T^d)$, and write $\langle k\rangle=(1+|k|^2)^{1/2}$. For
\begin{equation}
    e_\theta(v):=\cg_\theta(v)-\cg(v),
    \qquad
    X_{\theta,k}(v):=
    \left|\left\langle e_\theta(v),\phi_k\right\rangle_{L^2}\right|^2,
    \label{apdx_eq:modal-squared-error}
\end{equation}
define the modal standard deviation
\begin{equation}
    s_{\theta,\nu}(k)
    :=
    \operatorname{Var}_{v\sim\nu}\!\left[X_{\theta,k}(v)\right]^{1/2}.
    \label{apdx_eq:modal-standard-deviation}
\end{equation}
We package these coefficients into the deterministic spectral profile function $\mathfrak{s}_{\theta,\nu}$ defined by
\begin{equation}
    \widehat{\mathfrak{s}}_{\theta,\nu}(k):=s_{\theta,\nu}(k).
    \label{apdx_eq:modal-variance-profile}
\end{equation}
This profile is not a characteristic function in the probabilistic sense. It records the standard deviation of the squared prediction error in each Fourier mode. We allow its Sobolev norm to take values in $[0,\infty]$; finiteness is precisely the summability condition used below.

\begin{theorem}[Variance amplification by residual regularity]
\label{apdx_thm:pde-loss-variance}
Let $\nu\in\spp(\cv)$, and suppose that the random variables $\{X_{\theta,k}\}_{k\in\mathbb Z^d}$ have finite second moments and are pairwise uncorrelated:
\begin{equation}
    \operatorname{Cov}_{v\sim\nu}
    \left(X_{\theta,k}(v),X_{\theta,j}(v)\right)=0,
    \qquad k\ne j.
    \label{apdx_eq:modal-decorrelation}
\end{equation}
Let $\cn=\cl$ be linear and translation invariant, with Fourier symbol $p_\cl(k)$ satisfying
\begin{equation}
    c_\cl\langle k\rangle^m
    \le |p_\cl(k)|
    \le C_\cl\langle k\rangle^m,
    \qquad k\in\mathbb Z^d,\quad |k|\ge K_0,
    \label{apdx_eq:multiplier-order}
\end{equation}
for some $m>0$, $K_0>0$, and constants $0<c_\cl\le C_\cl<\infty$.
Define
\begin{equation}
    \ell_{\mathrm{data}}^{(a)}(v)
    :=\|e_\theta(v)\|_{H^a}^2,
    \qquad
    \ell_{\mathrm{PDE}}^{(a)}(v)
    :=\|\cl e_\theta(v)\|_{H^a}^2.
    \label{apdx_eq:sample-loss-definitions}
\end{equation}
For $N\ge1$, define the truncated losses
\begin{align}
    \ell_{\mathrm{data},N}^{(a)}(v)
    &:=
    \sum_{|k|\le N}\langle k\rangle^{2a}X_{\theta,k}(v),
    \label{apdx_eq:truncated-data-loss}\\
    \ell_{\mathrm{PDE},N}^{(a)}(v)
    &:=
    \sum_{|k|\le N}\langle k\rangle^{2a}|p_\cl(k)|^2X_{\theta,k}(v).
    \label{apdx_eq:truncated-pde-loss}
\end{align}
We interpret the variance functionals in the extended sense
\begin{equation}
    \operatorname{Var}_\nu\!\left[\ell_\square^{(a)}\right]
    :=
    \lim_{N\to\infty}
    \operatorname{Var}_\nu\!\left[\ell_{\square,N}^{(a)}\right]
    \in[0,\infty],
    \qquad
    \square\in\{\mathrm{data},\mathrm{PDE}\}.
    \label{apdx_eq:extended-variance-functional}
\end{equation}
The limit exists because \Cref{apdx_eq:modal-decorrelation} makes the truncated variances nondecreasing. Whenever the full loss is the $L^2(\nu)$ limit of its truncations, \eqref{apdx_eq:extended-variance-functional} equals its ordinary variance. For nonnegative functionals $A$ and $B$, write $A\gg B$ if $B\le CA$ for a constant independent of the error distribution whenever $A<\infty$, and if there also exists an admissible error distribution for which $A=\infty$ but $B<\infty$.
Then, for every admissible $a$,
\begin{align}
    \operatorname{Var}_\nu\!\left[\ell_{\mathrm{data}}^{(a)}\right]
    &=\|\mathfrak{s}_{\theta,\nu}\|_{H^{2a}}^2,
    \label{apdx_eq:data-loss-variance-identity}\\
    \operatorname{Var}_\nu\!\left[\ell_{\mathrm{PDE}}^{(a)}\right]
    &=
    \sum_{k\in\mathbb Z^d}
    \langle k\rangle^{4a}|p_\cl(k)|^4s_{\theta,\nu}(k)^2.
    \label{apdx_eq:pde-loss-variance-identity}
\end{align}
Consequently,
\begin{equation}
    c_\cl^4\|\mathfrak{s}_{\theta,\nu}\|_{H^{2(a+m)}}^2
    \le
    \operatorname{Var}_\nu\!\left[\ell_{\mathrm{PDE}}^{(a)}\right]
    \le
    C_\cl^4\|\mathfrak{s}_{\theta,\nu}\|_{H^{2(a+m)}}^2.
    \label{apdx_eq:pde-data-variance-comparison}
\end{equation}
In particular, if $\alpha'>\alpha$, then
\begin{equation}
    \operatorname{Var}_\nu\!\left[\ell_{\mathrm{PDE}}^{(\alpha')}\right]
    \gg
    \operatorname{Var}_\nu\!\left[\ell_{\mathrm{PDE}}^{(\alpha)}\right]
    \asymp
    \operatorname{Var}_\nu\!\left[\ell_{\mathrm{data}}^{(\alpha+m)}\right] \gg
    \operatorname{Var}_\nu\!\left[\ell_{\mathrm{data}}^{(\alpha)}\right].
    \label{apdx_eq:formal-variance-chain}
\end{equation}
The comparison constants depend only on $c_\cl$ and $C_\cl$. As a concrete example, if $s_{\theta,\nu}$ is nonzero and supported on modes satisfying $K\le\langle k\rangle\le cK$ for a fixed $c\ge1$, then
\begin{equation}
    \frac{\operatorname{Var}_\nu[\ell_{\mathrm{PDE}}^{(\alpha')}]}
         {\operatorname{Var}_\nu[\ell_{\mathrm{PDE}}^{(\alpha)}]}
    \asymp K^{4(\alpha'-\alpha)},
    \qquad
    \frac{\operatorname{Var}_\nu[\ell_{\mathrm{data}}^{(\alpha+m)}]}
         {\operatorname{Var}_\nu[\ell_{\mathrm{data}}^{(\alpha)}]}
    \asymp K^{4m}.
    \label{apdx_eq:high-frequency-variance-separation}
\end{equation}
\end{theorem}

\subsection{Proof of \Cref{thm:pde-loss-variance}}

\begin{proof}
\medskip\noindent\textbf{Step 1. Variance identities.}
By Parseval's identity,
\begin{equation}
    \ell_{\mathrm{data}}^{(a)}(v)
    =
    \sum_{k\in\mathbb Z^d}
    \langle k\rangle^{2a}X_{\theta,k}(v).
    \label{apdx_eq:data-loss-modal-expansion}
\end{equation}
For each finite truncation, expanding the variance gives the sum of the individual variances plus pairwise covariance terms. The covariance terms vanish by \Cref{apdx_eq:modal-decorrelation}. Taking the monotone limit in \Cref{apdx_eq:extended-variance-functional} yields
\begin{equation}
    \operatorname{Var}_\nu\!\left[\ell_{\mathrm{data}}^{(a)}\right]
    =
    \sum_{k\in\mathbb Z^d}
    \langle k\rangle^{4a}
    \operatorname{Var}_\nu[X_{\theta,k}]
    =
    \|\mathfrak{s}_{\theta,\nu}\|_{H^{2a}}^2,
    \label{apdx_eq:data-loss-variance-proof}
\end{equation}
which proves \Cref{apdx_eq:data-loss-variance-identity}.

Because $v=\cl\cg(v)$ and $\cn=\cl$ is linear, the residual of the prediction is
\begin{equation}
    \cl\cg_\theta(v)-v
    =\cl\bigl(\cg_\theta(v)-\cg(v)\bigr)
    =\cl e_\theta(v).
    \label{apdx_eq:linear-residual-error}
\end{equation}
The multiplier representation therefore gives
\begin{equation}
    \ell_{\mathrm{PDE}}^{(a)}(v)
    =
    \sum_{k\in\mathbb Z^d}
    \langle k\rangle^{2a}|p_\cl(k)|^2X_{\theta,k}(v).
    \label{apdx_eq:pde-loss-modal-expansion}
\end{equation}
Applying the same covariance calculation to the truncated PDE losses and then taking the monotone limit proves \Cref{apdx_eq:pde-loss-variance-identity}. Raising \Cref{apdx_eq:multiplier-order} to the fourth power and substituting it into that identity proves \Cref{apdx_eq:pde-data-variance-comparison}.

For $\alpha'>\alpha$, each summand in \Cref{apdx_eq:pde-loss-variance-identity} is multiplied by $\langle k\rangle^{4(\alpha'-\alpha)}\ge1$. Hence finiteness of the stronger PDE variance bounds the weaker one with constant one. The comparison with the $H^{\alpha+m}$ data-loss variance follows from \Cref{apdx_eq:data-loss-variance-identity,apdx_eq:pde-data-variance-comparison}.

\medskip\noindent\textbf{Step 2. Strictness.}
Fix $s_2>s_1$, and choose distinct modes $k_n$ such that
\begin{equation}
    \langle k_n\rangle^{4(s_2-s_1)}\ge n^2,
    \qquad n\ge2.
    \label{apdx_eq:sparse-mode-growth}
\end{equation}
Define a modal standard-deviation profile by
\begin{equation}
    s(k_n):=\frac{\langle k_n\rangle^{-2s_1}}{n},
    \qquad
    s(k):=0\quad\text{otherwise}.
    \label{apdx_eq:strictness-profile}
\end{equation}
Then
\begin{equation}
    \sum_k\langle k\rangle^{4s_1}s(k)^2
    =\sum_{n\ge2}\frac{1}{n^2}<\infty,
    \qquad
    \sum_k\langle k\rangle^{4s_2}s(k)^2
    \ge\sum_{n\ge2}1=\infty.
    \label{apdx_eq:strict-sobolev-separation}
\end{equation}
This profile is realized by an admissible error distribution. Let $B_n$ be independent Bernoulli variables with $\mathbb P(B_n=1)=q_n:=n^{-4}$, and set
\begin{equation}
    X_{\theta,k_n}
    :=
    \frac{s(k_n)}{\sqrt{q_n(1-q_n)}}B_n,
    \qquad
    X_{\theta,k}:=0\quad\text{otherwise}.
    \label{apdx_eq:strictness-realization}
\end{equation}
The variables are independent and satisfy $\operatorname{Var}(X_{\theta,k})=s(k)^2$. Moreover,
\begin{equation}
    \sum_{n\ge2}
    \langle k_n\rangle^{2s_1}\eee X_{\theta,k_n}
    =
    \sum_{n\ge2}\frac{1}{n}\sqrt{\frac{q_n}{1-q_n}}
    <\infty,
    \label{apdx_eq:strictness-finite-mean}
\end{equation}
so the weaker loss is finite almost surely. Taking the Fourier coefficient of $e_\theta$ at $k_n$ to be $X_{\theta,k_n}^{1/2}$ realizes the required squared modal errors.

Apply \Cref{apdx_eq:strict-sobolev-separation} with $(s_1,s_2)=(\alpha+m,\alpha'+m)$. The multiplier comparison then gives an error distribution with finite $H^\alpha$ PDE-loss variance and infinite $H^{\alpha'}$ PDE-loss variance, proving the first strict relation in \Cref{apdx_eq:formal-variance-chain}. The same construction with $(s_1,s_2)=(\alpha,\alpha+m)$ proves the second strict relation.

\medskip\noindent\textbf{Step 3. High-frequency separation.}
If the spectral profile is supported on $K\le\langle k\rangle\le cK$, the factors $\langle k\rangle^{4(\alpha'-\alpha)}$ and $\langle k\rangle^{4m}$ are uniformly comparable to $K^{4(\alpha'-\alpha)}$ and $K^{4m}$, respectively. Taking the corresponding ratios proves \Cref{apdx_eq:high-frequency-variance-separation}.
\end{proof}

The decorrelation assumption isolates the regularity mechanism by removing cross-mode covariance. For a nonlinear    {input} operator, the residual contains products of modal coefficients; its variance then depends on cross-covariances and higher-order joint moments that cannot be ordered by differential order alone.

\section{Proof of \Cref{thm:manufactured-shift-informal}}
\label{app:synthetic-shift-proof}

Let $d_H^2(\rho,\nu):=1-\operatorname{Aff}(\rho,\nu)$, where
$\operatorname{Aff}(\rho,\nu):=\int\sqrt{d\rho\,d\nu}$ is the Hellinger affinity.
For a distribution $\rho$ on $L^2(\mathbb T^d)$, define its sharp Sobolev threshold by
\begin{equation}
    s_\rho
    :=
    \sup\left\{s\in\mathbb R:\rho(H^s(\mathbb T^d))=1\right\},
    \label{apdx_thm3:eq:sobolev-threshold}
\end{equation}
and assume $\rho(H^s)=0$ for every $s>s_\rho$.

\subsection{Formal statement of \Cref{thm:manufactured-shift-informal}}

\begin{theorem}[Structural distribution shift under manufactured sampling]
\label{apdx_thm3:thm:formal}
Let $\mu\in\spp(\cu)$ be a centered Gaussian distribution on $L^2(\mathbb T^d)$ with sharp threshold $s_\mu$, and let $\nu\in\spp(\cv)$ have sharp threshold $s_\nu$. We also assume $\nu$ is a Gaussian distribution.
\begin{enumerate}\renewcommand{\labelenumi}{\textup{(\roman{enumi})}}
    \item Suppose $\cn=\cl+\cb$, where $\cl$ is elliptic of order $K>0$ and $\cb$ has order $m<K$. Assume $\cb:H^q\to H^{q-m}$ is continuous for $q<s_\mu$ sufficiently close to $s_\mu$. Then $\cn_\#\mu$ has sharp threshold $s_\mu-K$. Consequently,
    \begin{equation}
        s_\nu\ne s_\mu-K
        \quad\Longrightarrow\quad
        \cn_\#\mu\perp\nu,
        \qquad
        d_H(\cn_\#\mu,\nu)=1.
        \label{apdx_thm3:eq:threshold-singularity}
    \end{equation}

    \item Suppose $\cb=0$ and there is an orthonormal basis $\{\phi_j\}_{j\ge1}$ satisfying
    \begin{equation}
        \cl\phi_j=\ell_j\phi_j,
        \qquad
        C_\mu\phi_j=\sigma_{\mu,j}^2\phi_j,
        \qquad
        C_\nu\phi_j=\sigma_{\nu,j}^2\phi_j.
        \label{apdx_thm3:eq:diagonal-data}
    \end{equation}
    Let $\rho:=\cl_\#\mu$, define
    \begin{equation}
        \lambda_j
        :=
        \frac{\ell_j^2\sigma_{\mu,j}^2}{\sigma_{\nu,j}^2},
        \qquad
        b_j:=\lambda_j-1,
        \label{apdx_thm3:eq:lambda-def}
    \end{equation}
    Let $P_N$ project onto the first $N$ modes, and set
    \begin{equation}
        \rho_N:=(P_N)_\#(\cl_\#\mu),
        \quad
        \nu_N:=(P_N)_\#\nu,
        \quad
        d_{H, N}=d_H(\rho_N, \nu_N),
        \label{apdx_thm3:eq:projected-distributions}
    \end{equation}
    where $d_{H, N}$ is the Hellinger distance between the first $N$ modes of $\rho$ and $\nu$. Then
    \begin{equation}
        d_{H, N}^2
        =
        1-
        \exp\!\left(-\frac12\sum_{j=1}^N\psi(b_j)\right),
        \qquad
        \psi(b):=-\log\!\left(\frac{2\sqrt{1+b}}{2+b}\right).
        \label{apdx_thm3:eq:hellinger-exponential}
    \end{equation}
    As $b\to0$,
    \begin{equation}
        \psi(b)=\frac{b^2}{8}+O(|b|^3).
        \label{apdx_thm3:eq:hellinger-log-expansion}
    \end{equation}
    Consequently,
    \begin{equation}
        d_H(\rho,\nu)<1
        \quad\Longleftrightarrow\quad
        \sum_{j\ge1}b_j^2<\infty.
        \label{apdx_thm3:eq:l2-dichotomy}
    \end{equation}

    \item For an arbitrary $\cn$, similarly define $\rho:=\cn_\#\mu$ and $\rho_N:=(P_N)_\#(\cn_\#\mu)$. Suppose there are measurable statistics $T_N$ with finite variances and nonzero mean gaps
    \begin{equation}
        \Delta_N
        :=
        \left|\eee_{\rho_N}T_N-\eee_{\nu_N}T_N\right|>0
        \label{apdx_thm3:eq:statistic-gap}
    \end{equation}
    such that
    \begin{equation}
        \frac{
            \sqrt{\operatorname{Var}_{\rho_N}(T_N)}
            +\sqrt{\operatorname{Var}_{\nu_N}(T_N)}
        }{\Delta_N}
        \longrightarrow0.
        \label{apdx_thm3:eq:statistic-separation}
    \end{equation}
    Then $d_{H, N}\to1$. This criterion applies in particular when nonlinear mode interactions create an accumulated statistic that is absent from the target distribution.
\end{enumerate}
\end{theorem}

\subsection{A Gaussian equivalence criterion}

\begin{lemma}[Feldman–Hájek theorem; see, e.g., \cite{da2014stochastic}]
\label{thm3:lem:gaussian-equivalence}
Let $\rho_1=\mathcal N(0,C_1)$ and $\rho_2=\mathcal N(0,C_2)$ be centered Gaussian measures on a separable Hilbert space. They are equivalent if and only if
\begin{equation}
    \operatorname{Ran}(C_1^{1/2})
    =
    \operatorname{Ran}(C_2^{1/2})
    \label{apdx_thm3:eq:CM-condition}
\end{equation}
and
\begin{equation}
    C_2^{-1/2}C_1C_2^{-1/2}-I
    \quad\text{is Hilbert--Schmidt}.
    \label{apdx_thm3:eq:HS-condition}
\end{equation}
If either condition fails, the measures are mutually singular.
\end{lemma}

\subsection{Proof of part \textup{(i)} of \Cref{apdx_thm3:thm:formal}}

\begin{proof}
Fix $r<s_\mu-K$. Then $r+K<s_\mu$, so $u\in H^{r+K}$ for $\mu$-almost every $u$. Elliptic regularity gives $\cl u\in H^r$. Since $m<K$, we also have $r+m<s_\mu$. The assumed mapping property of $\cb$ therefore gives $\cb(u)\in H^r$. Hence
\begin{equation}
    (\cn_\#\mu)(H^r)=1,
    \qquad r<s_\mu-K.
    \label{apdx_thm3:eq:threshold-upper}
\end{equation}

For the converse, first take $r\in(s_\mu-K,s_\mu-m)$. If $\cl u\in H^r$, elliptic regularity would imply $u\in H^{r+K}$. This event has $\mu$-probability zero because $r+K>s_\mu$. By contrast, $r+m<s_\mu$ gives $\cb(u)\in H^r$ almost surely. Therefore $\cn(u)=\cl u+\cb(u)$ cannot belong to $H^r$ almost surely. If $r\ge s_\mu-m$, choose $r_0\in(s_\mu-K,s_\mu-m)$ with $r_0<r$ and use the embedding $H^r\subset H^{r_0}$. Thus
\begin{equation}
    (\cn_\#\mu)(H^r)=0,
    \qquad r>s_\mu-K.
    \label{apdx_thm3:eq:threshold-lower}
\end{equation}
Equations~\eqref{apdx_thm3:eq:threshold-upper} and~\eqref{apdx_thm3:eq:threshold-lower} establish the threshold $s_\mu-K$. If $s_\nu\ne s_\mu-K$, a Sobolev space with index strictly between the thresholds has probability one under one distribution and zero under the other. This proves \eqref{apdx_thm3:eq:threshold-singularity}.
\end{proof}

\subsection{Proof of part \textup{(ii)} of \Cref{apdx_thm3:thm:formal}}

\begin{proof}
The $j$th coordinate of $\rho$ is Gaussian with variance $\ell_j^2\sigma_{\mu,j}^2$. For centered one-dimensional Gaussians with variances $a^2$ and $b^2$, direct integration gives the Hellinger affinity
\begin{equation}
    \int_{\mathbb R}\sqrt{p_a(x)p_b(x)}\,dx
    =
    \left(\frac{2ab}{a^2+b^2}\right)^{1/2}.
    \label{apdx_thm3:eq:one-dimensional-affinity}
\end{equation}
Independence across modes therefore yields
\begin{equation}
    \operatorname{Aff}(\rho_N,\nu_N)^2
    =
    \prod_{j=1}^N\frac{2\sqrt{\lambda_j}}{1+\lambda_j}
    =
    \exp\!\left(-\sum_{j=1}^N\psi(b_j)\right),
    \label{apdx_thm3:eq:affinity-product}
\end{equation}
therefore $\operatorname{Aff}(\rho_N,\nu_N) = \exp(-\frac12\sum_{j=1}^N\psi(b_j))$, which proves \eqref{apdx_thm3:eq:hellinger-exponential}.

Taylor expansion at $b=0$ gives
\begin{equation}
    \log\!\left(\frac{2\sqrt{1+b}}{2+b}\right)
    =
    -\frac{b^2}{8}+\frac{b^3}{8}+O(b^4),
    \label{apdx_thm3:eq:log-taylor}
\end{equation}
and hence \eqref{apdx_thm3:eq:hellinger-log-expansion}. If $b_j$ does not converge to zero, the summands $\psi(b_j)$ remain bounded away from zero along a subsequence. If $b_j\to0$, then $\psi(b_j)\asymp b_j^2$. These two cases prove \eqref{apdx_thm3:eq:l2-dichotomy}.

 In particular,
\begin{equation}
    b_j\asymp j^{-q}
    \quad\Longrightarrow\quad
    H(\rho,\nu)<1
    \ \text{if and only if}\ q>\frac12.
    \label{apdx_thm3:eq:modal-decay-rate}
\end{equation}
For Fourier modes indexed by $k\in\mathbb Z^d$, the analogous condition $b_k\asymp\langle k\rangle^{-q}$ is $q>d/2$.

The calibration condition can be written as
\begin{equation}
    \sigma_{\mu,j}^2
    =
    \frac{\sigma_{\nu,j}^2}{\ell_j^2}(1+b_j),
    \qquad
    (b_j)_{j\ge1}\in\ell^2.
    \label{apdx_thm3:eq:calibration}
\end{equation}
Thus the source distribution must match both the target covariance and the PDE multiplier at every mode.
\end{proof}

\begin{theorem}
\label{thm3:thm:elliptic-calibration}
Assume $|\ell_j|\asymp\rho_j^{K/2}$, $\sigma_{\mu,j}^2\asymp\rho_j^{-\alpha}$, and $\sigma_{\nu,j}^2\asymp\rho_j^{-\beta}$ as $j\to\infty$. If $\beta\ne\alpha-K$, then $H(\rho,\nu)=1$. If $\beta=\alpha-K$, non-singularity still requires the relative mismatch $b_j$ in \eqref{apdx_thm3:eq:lambda-def} to belong to $\ell^2$. If $\rho_j\asymp j^{2/d}$ and $|b_j|\asymp\rho_j^{-\gamma}$, this condition holds if and only if $\gamma>d/4$.
\end{theorem}

\begin{proof}
The assumptions give $\lambda_j\asymp\rho_j^{K-\alpha+\beta}$. If $\beta\ne\alpha-K$, then $\lambda_j$ cannot converge to one, so \eqref{apdx_thm3:eq:l2-dichotomy} gives singularity. When $\beta=\alpha-K$, the remaining condition is $\sum_jb_j^2<\infty$. Weyl scaling gives $b_j^2\asymp j^{-4\gamma/d}$, whose series converges exactly when $\gamma>d/4$.
\end{proof}

\begin{remark}[Implications for choosing the solution distribution]
\label{thm3:rem:calibration}
Equation~\eqref{apdx_thm3:eq:calibration} requires the covariance of $\mu$ to approximate $(\cl^*\cl)^{-1}C_\nu$ with square-summable relative error. Constructing such a distribution requires detailed information about both the target covariance and the inverse action of the PDE operator. This requirement limits the practical value of direct calibration.
\end{remark}

Outside the diagonal setting, \Cref{thm3:lem:gaussian-equivalence} gives the corresponding Gaussian criterion. Matching Sobolev thresholds alone is insufficient. The Cameron--Martin spaces must agree, and the relative covariance perturbation must be Hilbert--Schmidt.

\subsection{Proof of part \textup{(iii)} of \Cref{apdx_thm3:thm:formal}}

\begin{proof}
Assume without loss of generality that $\eee_{\rho_N}T_N>\eee_{\nu_N}T_N$, and define
\begin{equation}
    A_N
    :=
    \left\{
        T_N\ge
        \frac{\eee_{\rho_N}T_N+\eee_{\nu_N}T_N}{2}
    \right\}.
    \label{apdx_thm3:eq:separating-event}
\end{equation}
Chebyshev's inequality gives
\begin{equation}
    \rho_N(A_N^c)
    \le
    \frac{4\operatorname{Var}_{\rho_N}(T_N)}{\Delta_N^2},
    \qquad
    \nu_N(A_N)
    \le
    \frac{4\operatorname{Var}_{\nu_N}(T_N)}{\Delta_N^2}.
    \label{apdx_thm3:eq:separating-event-bounds}
\end{equation}
Condition~\eqref{apdx_thm3:eq:statistic-separation} makes both terms vanish. Splitting the affinity over $A_N$ and $A_N^c$ and applying Cauchy--Schwarz yields

\begin{align}
  \operatorname{Aff}(\rho_N,\nu_N)
  &=
  \int_{A_N} \sqrt{d\rho_N\,d\nu_N}
  +
  \int_{A_N^c} \sqrt{d\rho_N\,d\nu_N} \notag\\
  &\le
  \left(\int_{A_N} d\rho_N\right)^{1/2}
  \left(\int_{A_N} d\nu_N\right)^{1/2}
  +
  \left(\int_{A_N^c} d\rho_N\right)^{1/2}
  \left(\int_{A_N^c} d\nu_N\right)^{1/2} \notag\\
  &=
  \sqrt{\rho_N(A_N)\nu_N(A_N)}
  +
  \sqrt{\rho_N(A_N^c)\nu_N(A_N^c)}
  \longrightarrow 0 .
  \label{apdx_thm3:eq:affinity-separation}
\end{align}

Therefore $H(\rho_N,\nu_N)\to1$, which proves part~\textup{(iii)}.
\end{proof}

\subsection{A nonlinear example for part \textup{(iii)}}\label{apdx_thm3:sec:kirchhoff-example}

   {We construct a nonlinear operator for which parts~\textup{(i)} and~\textup{(ii)} of \Cref{apdx_thm3:thm:formal} do not detect the distribution shift, while part~\textup{(iii)} does.}

   {Let $\T=[0,2\pi)$, and consider mean-zero functions on $\T$. Let $\{\phi_j\}_{j\ge1}$ be the real Fourier basis
\begin{equation}
    \phi_{2k-1}=\pi^{-1/2}\cos(kx),
    \qquad
    \phi_{2k}=\pi^{-1/2}\sin(kx),
    \qquad k\ge1,
    \label{apdx_thm3:eq:fourier-basis}
\end{equation}
and let $k_j:=\lceil j/2\rceil$ denote the wavenumber of $\phi_j$. Then $\{\phi_j\}_{j\ge1}$ is an orthonormal basis of mean-zero $L^2(\T)$, and $\cl:=-\partial_x^2$ satisfies $\cl\phi_j=k_j^2\phi_j$. Powers of $\cl$ are defined spectrally. Consider the stationary Kirchhoff operator
\begin{equation}
    \cn(u):=-\bigl(1+\norm{\partial_x u}_{L^2}^2\bigr)\,\partial_x^2 u=M(u)\,\cl u,
    \qquad
    M(u):=1+\norm{\partial_x u}_{L^2}^2 .
    \label{apdx_thm3:eq:kirchhoff}
\end{equation}
Fix $a>5/2$ and $c>0$, and define
\begin{equation}
    \mu:=\mathcal N\bigl(0,\cl^{-a}\bigr),
    \qquad
    \nu:=\mathcal N\bigl(0,c^2\cl^{2-a}\bigr),
    \qquad
    \rho:=\cn_\#\mu .
    \label{apdx_thm3:eq:kirchhoff-measures}
\end{equation}
Both covariance operators are diagonal in $\{\phi_j\}_{j\ge1}$, with $\sigma_{\mu,j}^2=k_j^{-2a}$ and $\sigma_{\nu,j}^2=c^2k_j^{4-2a}$. In particular, $\nu=(c\cl)_\#\mu$. Let $(\xi_j)_{j\ge1}$ and $(\eta_j)_{j\ge1}$ be i.i.d.\ standard Gaussian random variables. By the Karhunen--Lo\`eve expansion, samples $u\sim\mu$ and $v\sim\nu$ can be written as
\begin{equation}
    u=\sum_{j\ge1}k_j^{-a}\,\xi_j\,\phi_j,
    \qquad
    v=c\sum_{j\ge1}k_j^{2-a}\,\eta_j\,\phi_j .
    \label{apdx_thm3:eq:kirchhoff-series}
\end{equation}
For $v$ in the support of $\rho$ or $\nu$, define
\begin{equation}
    R_N(v):=\frac1N\sum_{j=1}^N\frac{\jp{v,\phi_j}^2}{k_j^4\sigma_{\mu,j}^2},
    \qquad
    T_N(v):=\mathbf 1\Bigl\{\bigl|R_N(v)/c^2-1\bigr|>N^{-1/4}\Bigr\}.
    \label{apdx_thm3:eq:kirchhoff-statistic}
\end{equation}
Since $T_N$ depends only on the first $N$ coefficients, it is a measurable statistic of $\rho_N$ and $\nu_N$.}

\begin{proposition}
\label{apdx_thm3:prop:kirchhoff}
   {Let $\mu$, $\nu$, $\rho$ and $T_N$ be defined by \cref{apdx_thm3:eq:kirchhoff,apdx_thm3:eq:kirchhoff-measures,apdx_thm3:eq:kirchhoff-statistic} with $a>5/2$ and $c>0$. Then the following hold.}
\begin{enumerate}\renewcommand{\labelenumi}{\textup{(\alph{enumi})}}
    \item    {The distributions $\rho$ and $\nu$ are supported on $L^2(\T)$, and
    \begin{equation}
        s_\rho=s_\nu=s_\mu-2=a-\tfrac52 .
        \label{apdx_thm3:eq:kirchhoff-thresholds}
    \end{equation}}
    \item    {The statistics $T_N$ satisfy \cref{apdx_thm3:eq:statistic-separation}. Consequently, $d_{H,N}\to1$, $d_H(\rho,\nu)=1$, and $\rho\perp\nu$.}
\end{enumerate}
\end{proposition}

\begin{proof}
   {We first write $\cn(u)$ in the basis $\{\phi_j\}_{j\ge1}$, which reduces both statements to properties of the scalar $M$. We then compute the thresholds and verify \cref{apdx_thm3:eq:statistic-separation}.}

\textit{   {Step 1.}}    {We express the coefficients of $\cn(u)$ in terms of $(\xi_j)_{j\ge1}$. Since $\partial_x\phi_{2k-1}=-k\phi_{2k}$ and $\partial_x\phi_{2k}=k\phi_{2k-1}$, Parseval's identity and \cref{apdx_thm3:eq:kirchhoff-series} give $\norm{\partial_x u}_{L^2}^2=\sum_{j\ge1}k_j^{2-2a}\xi_j^2$. By Tonelli's theorem, this series has expectation $2\sum_{k\ge1}k^{2-2a}<\infty$, so it is finite almost surely. Therefore
\begin{equation}
    \jp{\cn(u),\phi_j}=M\,k_j^{2-a}\,\xi_j,
    \qquad
    M=1+\sum_{i\ge1}k_i^{2-2a}\xi_i^2,
    \label{apdx_thm3:eq:kirchhoff-modal}
\end{equation}
where $M\in[1,\infty)$ almost surely. We also show that $M$ has a continuous distribution, that is, $\mathbb P(M=m)=0$ for every $m\in\mathbb R$. Since $k_1=1$, we can write $M=\xi_1^2+A$, where $A$ is independent of $\xi_1$. Because $\xi_1^2$ has a continuous distribution, $\mathbb P(M=m)=\eee\bigl[\mathbb P(\xi_1^2=m-A\mid A)\bigr]=0$ for every $m\in\mathbb R$.}

\textit{   {Step 2.}}    {We compute the sharp Sobolev thresholds, which proves part~\textup{(a)}. Let $c_j\ge0$. If $\sum_jc_j<\infty$, then $\sum_jc_j\xi_j^2<\infty$ almost surely by Tonelli's theorem. If $\sum_jc_j=\infty$, then $\eee\exp(-\sum_jc_j\xi_j^2)=\prod_j(1+2c_j)^{-1/2}=0$, so $\sum_jc_j\xi_j^2=\infty$ almost surely. By \cref{apdx_thm3:eq:kirchhoff-series},
\[
    \norm{v}_{\dot H^s}^2=c^2\sum_{j\ge1}k_j^{2s+4-2a}\eta_j^2,
    \qquad
    \sum_{j\ge1}k_j^{2s+4-2a}=2\sum_{k\ge1}k^{2s+4-2a},
\]
and the last series converges if and only if $s<a-5/2$. Hence $s_\nu=a-5/2$. By \cref{apdx_thm3:eq:kirchhoff-modal}, $\norm{\cn(u)}_{\dot H^s}^2=M^2\sum_{j\ge1}k_j^{2s+4-2a}\xi_j^2$. Since $1\le M<\infty$ almost surely, the same dichotomy holds, so $s_\rho=a-5/2$. The same argument applied to $u$ in \cref{apdx_thm3:eq:kirchhoff-series} gives $s_\mu=a-1/2$, which proves \cref{apdx_thm3:eq:kirchhoff-thresholds}. Since $a>5/2$, both $\rho$ and $\nu$ are supported on $L^2(\T)$.} 

   {We now turn to part~\textup{(b)}. Denote $p_N:=\eee_{\nu_N}T_N$ and $q_N:=\eee_{\rho_N}T_N$. Since $T_N$ takes values in $\{0,1\}$, we have $\operatorname{Var}_{\nu_N}(T_N)=p_N(1-p_N)$ and $\operatorname{Var}_{\rho_N}(T_N)=q_N(1-q_N)$. It therefore suffices to show $p_N\to0$ and $q_N\to1$.}

\textit{   {Step 3.}}    {We bound $p_N$. By \cref{apdx_thm3:eq:kirchhoff-series}, $R_N(v)=c^2N^{-1}\sum_{j=1}^N\eta_j^2$ under $\nu$. Thus $R_N/c^2-1=N^{-1}\sum_{j=1}^N(\eta_j^2-1)$ has mean zero and variance $2/N$. Chebyshev's inequality gives
\begin{equation}
    p_N\le\frac{2/N}{N^{-1/2}}=2N^{-1/2}.
    \label{apdx_thm3:eq:kirchhoff-target-bound}
\end{equation}}

\textit{   {Step 4.}}    {We show $q_N\to1$. By \cref{apdx_thm3:eq:kirchhoff-modal}, $R_N(\cn(u))=M^2\,\Xi_N$ with $\Xi_N:=N^{-1}\sum_{j=1}^N\xi_j^2$. By the strong law of large numbers, $\Xi_N\to1$ almost surely, so $R_N/c^2-1\to M^2/c^2-1$ almost surely. Since $M\ge1$ and $M$ has no atoms by Step~1, $\mathbb P(M^2=c^2)=\mathbb P(M=c)=0$. On the event $\{M^2\ne c^2\}$, the limit $M^2/c^2-1$ is nonzero while $N^{-1/4}\to0$, so $T_N=1$ for all sufficiently large $N$. Hence $T_N\to1$ almost surely, and $q_N\to1$ by the bounded convergence theorem.}

\textit{   {Step 5.}}    {We combine the two estimates. By Steps~3 and~4, $\Delta_N=q_N-p_N>0$ for all sufficiently large $N$, and
\[
    \frac{\sqrt{q_N(1-q_N)}+\sqrt{p_N(1-p_N)}}{q_N-p_N}
    \le
    \frac{\sqrt{1-q_N}+\sqrt{p_N}}{q_N-p_N}
    \longrightarrow0 .
\]
Thus \cref{apdx_thm3:eq:statistic-separation} holds, and part~\textup{(iii)} of \Cref{apdx_thm3:thm:formal} gives $d_{H,N}\to1$. Since $\rho_N=(P_N)_\#\rho$ and $\nu_N=(P_N)_\#\nu$, the data-processing inequality for the Hellinger affinity gives $\operatorname{Aff}(\rho,\nu)\le\operatorname{Aff}(\rho_N,\nu_N)\to0$. Hence $d_H(\rho,\nu)=1$ and $\rho\perp\nu$. This completes the proof.}
\end{proof}

\section{Proof of \Cref{thm:wasp-informal}}
\label{app:wasp-step-dependent-proof}
\begingroup

This section formalizes the two mechanisms used in \Cref{sec:analysis}. We first prove
the pathwise error-transfer bound, and then control the mismatch between the intended
and realized    {input} distributions by a nonlinear recursion.

\paragraph{Notation.}
Let $\cu$ and $\cv$ be the solution and    {input} spaces, and let
$\cn:\cu\to\cv$ and $\cg:\cv\to\cu$ denote the PDE and solution operators. We assume
$\cn\circ\cg=\id$ on the distributions considered below. Let $\mu_0\in\spp(\cu)$,
$\nu_0=\cn_\#\mu_0$, and let $\nu_1\in\spp(\cv)$ be the target    {input} distribution.
If $\ct:\cv\to\cv$ transports $\nu_0$ to the target $\nu_1$, define the path distributions
\[
  \nu_{s}=((1-s)\id+s\ct)_\#\nu_0,
  \qquad
  \nu_n:=\nu_{s_n},
  \qquad 0=s_0\leq s_1\leq\cdots\leq s_N=1,
\]
following the subscript convention of \Cref{sec:transporting-distribution}: an integer subscript is a stage index, and the bare symbol $\nu_1$ always denotes the target of \Cref{sec:setup}. For $n\geq1$, set
\[
  \hat\mu_n=(\cg^{n-1})_\#\nu_n,
  \qquad
  \hat\nu_n=\cn_\#\hat\mu_n,
\]
with $\hat\mu_0=\mu_0$ and $\hat\nu_0=\nu_0$. Write $\ch^n=\cg^n-\cg$ and
\begin{equation}
  \epsilon_n:=\eee_{v\sim\hat\nu_n}\|\ch^n(v)\|_{\cu},
  \quad
  \delta_n:=W(\nu_n,\nu_{n+1}),
  \quad
  e_n:=\eee_{v\sim\nu_n}\|\ch^n(v)\|_{\cu},
  \quad
  d_n:=W(\nu_n,\hat\nu_n).
  \label{apdx_thm4:eq:quantities}
\end{equation}
Throughout this section, $W$ denotes the $1$-Wasserstein distance induced by
$\|\cdot\|_{\cv}$.

\subsection{Formal statement}
\label{thm4:sec:formal}

\begin{assumption}[Operator regularity]
\label{thm4:assum:reg}
$\cn$ is $K_\cn$-Lipschitz, and $\ch^n$ is $K_\ch$-Lipschitz for every
$0\leq n\leq N$.
\end{assumption}

\begin{assumption}[Local distributional generalization]
\label{thm4:assum:goodgen}
There exist $\eta>0$ and $C_n>0$ such that, for every $0\leq n\leq N$ and every
$\nu'\in\spp(\cv)$,
\begin{equation}
  \eee_{v\sim\nu'}\|\ch^n(v)\|_{\cu}
  \leq
  \epsilon_n+C_nW(\nu',\hat\nu_n)^{1+\eta}.
  \label{apdx_thm4:eq:goodgen}
\end{equation}
\end{assumption}

\begin{theorem}[WaSP error propagation]
\label{apdx_thm4:thm:formal}
Suppose \Cref{thm4:assum:reg} and \Cref{thm4:assum:goodgen} hold. Define
\begin{equation}
  B_n:=2^\eta K_\cn C_n,
  \qquad 0\leq n\leq N,
  \label{apdx_thm4:eq:Bn}
\end{equation}
and
\begin{equation}
  a_n:=K_\cn\epsilon_n+B_n\delta_n^{1+\eta},
  \qquad 0\leq n<N.
  \label{apdx_thm4:eq:an}
\end{equation}
Let $\alpha\in(1,2)$ and set
\begin{equation}
  K_\alpha:=\min_{0\leq n\leq N}\alpha^nB_n^{-1/\eta}.
  \label{apdx_thm4:eq:K-alpha}
\end{equation}
For any $0<K\leq K_\alpha$, define
\begin{equation}
  R_n
  :=\left(K\alpha^{-n}B_n^{1/\eta}\right)^{1/(1+\eta)}
  (2B_n)^{-1/\eta}
  =2^{-1/\eta}B_n^{-1/(1+\eta)}
  (K\alpha^{-n})^{1/(1+\eta)},
  \qquad
  0\leq n\leq N.
  \label{apdx_thm4:eq:explicit-radii}
\end{equation}
If
\begin{equation}
  a_n\leq\left(1-\frac{\alpha}{2}\right)R_{n+1},
  \qquad 0\leq n<N,
  \label{apdx_thm4:eq:smallness}
\end{equation}
then $d_n\leq R_n$ for every $0\leq n\leq N$, and
\begin{equation}
  e_N
  \leq
  \epsilon_N+2K_\ch\max_{0\leq k<N}
  \left[K_\cn\epsilon_k+2^\eta K_\cn C_k\delta_k^{1+\eta}\right].
  \label{apdx_thm4:eq:main-bound}
\end{equation}
\end{theorem}

\subsection{Pathwise error transfer}

\begin{lemma}[Pathwise transport perturbation]
\label{thm4:lem:transport}
Let $\ct_n$ be an optimal transport map from $\nu_n$ to $\nu_{n+1}$. For
$\ct_nv\neq v$, define
\[
  L_n(v):=
  \frac{\|\ch^n(\ct_nv)-\ch^n(v)\|_{\cu}}
       {\|\ct_nv-v\|_{\cv}},
\]
and set $L_n(v)=0$ when $\ct_nv=v$. If $\delta_n>0$, then
\begin{equation}
  \overline K_n:=
  \frac{\displaystyle\int L_n(v)\|\ct_nv-v\|_{\cv}\,\nu_n(\mathrm dv)}
       {\displaystyle\int\|\ct_nv-v\|_{\cv}\,\nu_n(\mathrm dv)}
  \label{apdx_thm4:eq:pathwise-factor}
\end{equation}
satisfies
\begin{equation}
  \eee_{v\sim\nu_{n+1}}\|\ch^n(v)\|_{\cu}
  \leq
  \eee_{v\sim\nu_n}\|\ch^n(v)\|_{\cu}
  +\overline K_n\delta_n.
  \label{apdx_thm4:eq:pathwise-transfer}
\end{equation}
Moreover, $\overline K_n\leq\operatorname*{ess\,sup}_{v\sim\nu_n}L_n(v)\leq K_\ch$.
If $\delta_n=0$, \eqref{apdx_thm4:eq:pathwise-transfer} holds with
$\overline K_n=0$.
\end{lemma}

\begin{proof}
Since $(\ct_n)_\#\nu_n=\nu_{n+1}$, the triangle inequality gives
\begin{align*}
  \eee_{v\sim\nu_{n+1}}\|\ch^n(v)\|_{\cu}
  &=\eee_{v\sim\nu_n}\|\ch^n(\ct_nv)\|_{\cu}\\
  &\leq\eee_{v\sim\nu_n}\|\ch^n(v)\|_{\cu}
  +\int L_n(v)\|\ct_nv-v\|_{\cv}\,\nu_n(\mathrm dv).
\end{align*}
Optimality of $\ct_n$ implies
$\int\|\ct_nv-v\|_{\cv}\,\nu_n(\mathrm dv)=\delta_n$, which proves
\eqref{apdx_thm4:eq:pathwise-transfer}. The remaining bounds follow from the definition
of a weighted average and \Cref{thm4:assum:reg}. When $\delta_n=0$, the two
distributions coincide, so the assertion is immediate.
\end{proof}

\begin{lemma}[Global Lipschitz transfer]
\label{thm4:lem:global-transport}
For any $\nu,\nu'\in\spp(\cv)$ and any $K_\ch$-Lipschitz map $\ch:\cv\to\cu$,
\begin{equation}
  \eee_{v\sim\nu}\|\ch(v)\|_{\cu}
  \leq
  \eee_{v\sim\nu'}\|\ch(v)\|_{\cu}+K_\ch W(\nu,\nu').
  \label{apdx_thm4:eq:transport-lip}
\end{equation}
\end{lemma}

\begin{proof}
Let $\pi$ be an optimal coupling of $\nu$ and $\nu'$. The triangle inequality and the
Lipschitz property imply
\begin{align}
  \int\|\ch(v)\|_{\cu}\,\nu(\mathrm dv)
  &=\int\|\ch(v)-\ch(v')+\ch(v')\|_{\cu}\,\nu'(\mathrm dv')\\
  &\leq
  \int\|\ch(v')\|_{\cu}\,\nu'(\mathrm dv')
  +K_\ch\int\|v-v'\|_{\cv}\,\pi(\mathrm dv,\mathrm dv').
\end{align}

The final integral equals $W(\nu,\nu')$.
\end{proof}

\subsection{Worst-case Lipschitz bound}

\begin{corollary}[Worst-case error propagation]
\label{thm4:cor:worstcase}
Under \Cref{thm4:assum:reg}, for every $0\leq n<N$,
\begin{equation}
  d_{n+1}
  \leq
  K_\cn K_\ch(d_n+\delta_n)+K_\cn\epsilon_n.
  \label{apdx_thm4:eq:worstcase-step}
\end{equation}
Consequently,
\begin{equation}
  e_N
  \leq
  \epsilon_N+
  \sum_{k=0}^{N-1}(K_\cn K_\ch)^{N-k}
  \left(K_\ch\delta_k+\epsilon_k\right).
  \label{apdx_thm4:eq:worstcase-total}
\end{equation}
\end{corollary}

\begin{proof}
The triangle inequality gives
$W(\nu_{n+1},\hat\nu_n)\leq\delta_n+d_n$. Applying
\Cref{thm4:lem:global-transport} to $\ch^n$, $\nu_{n+1}$, and $\hat\nu_n$ yields
\[
  \eee_{v\sim\nu_{n+1}}\|\ch^n(v)\|_{\cu}
  \leq
  \epsilon_n+K_\ch(\delta_n+d_n).
\]
The coupling $v\mapsto\cn(\cg^n(v))$ and the $K_\cn$-Lipschitz continuity of
$\cn$ then imply \eqref{apdx_thm4:eq:worstcase-step}. Iterating this recursion from
$d_0=0$ gives
\[
  d_N
  \leq
  \sum_{k=0}^{N-1}(K_\cn K_\ch)^{N-1-k}
  \left(K_\cn K_\ch\delta_k+K_\cn\epsilon_k\right).
\]
A final application of \Cref{thm4:lem:global-transport} gives
$e_N\leq\epsilon_N+K_\ch d_N$, which proves
\eqref{apdx_thm4:eq:worstcase-total}.
\end{proof}

\subsection{Proof of \Cref{apdx_thm4:thm:formal}}
We prove $d_n\leq R_n$ by induction on $n$. We first derive the mismatch recursion,
close the induction using the explicit radii in \eqref{apdx_thm4:eq:explicit-radii}, and
then sum the resulting contraction and convert the distributional mismatch into
prediction error.

\begin{proof}
\textit{Step 1. Nonlinear mismatch recursion.}
Since
$\hat\nu_{n+1}=(\cn\circ\cg^n)_\#\nu_{n+1}$, coupling each
$v\sim\nu_{n+1}$ with $\cn(\cg^n(v))\sim\hat\nu_{n+1}$ gives
\begin{equation}
  \begin{aligned}
  d_{n+1}
  &=W\left(\nu_{n+1},\hat\nu_{n+1}\right) \\
  &=W\left(\nu_{n+1},(\cn\circ\cg^n)_\#\nu_{n+1}\right) \\
  &\leq
  \eee_{v\sim\nu_{n+1}}
  \left\|v-\cn(\cg^n(v))\right\|_{\cv} \\
  &=
  \eee_{v\sim\nu_{n+1}}
  \left\|\cn(\cg(v))-\cn(\cg^n(v))\right\|_{\cv} \\
  &\leq
  K_\cn\eee_{v\sim\nu_{n+1}}
  \left\|\cg^n(v)-\cg(v)\right\|_{\cu} \\
  &=K_\cn\eee_{v\sim\nu_{n+1}}\|\ch^n(v)\|_{\cu}.
  \end{aligned}
  \label{apdx_thm4:eq:coupling-bound}
\end{equation}
Applying \Cref{thm4:assum:goodgen} with $\nu'=\nu_{n+1}$ and using
$W(\nu_{n+1},\hat\nu_n)\leq\delta_n+d_n$, we obtain
\begin{equation}
  d_{n+1}
  \leq K_\cn\epsilon_n+K_\cn C_n(\delta_n+d_n)^{1+\eta}.
  \label{apdx_thm4:eq:pre-recursion}
\end{equation}
Since $(x+y)^{1+\eta}\leq2^\eta(x^{1+\eta}+y^{1+\eta})$ for $x,y\geq0$,
\eqref{apdx_thm4:eq:pre-recursion} becomes
\begin{equation}
  d_{n+1}\leq a_n+B_nd_n^{1+\eta}.
  \label{apdx_thm4:eq:nonlinear-recursion}
\end{equation}

\textit{Step 2. Invariant shrinking radii.}
The definition of $R_n$ gives
\begin{equation}
  B_nR_n^{1+\eta}
  =2^{-(1+\eta)/\eta}K\alpha^{-n}.
  \label{apdx_thm4:eq:radius-identity}
\end{equation}
Moreover, $K\leq K_\alpha\leq\alpha^{n+1}B_{n+1}^{-1/\eta}$ implies
\begin{equation}
  \frac{B_nR_n^{1+\eta}}{R_{n+1}}
  =\frac{\alpha}{2}
  \left(\frac{K^\eta B_{n+1}}{\alpha^{(n+1)\eta}}\right)^{1/(1+\eta)}
  \leq\frac{\alpha}{2}.
  \label{apdx_thm4:eq:radius-transfer}
\end{equation}
We now prove $d_n\leq R_n$ by induction. The base case follows from $d_0=0$. If
$d_n\leq R_n$, then \eqref{apdx_thm4:eq:nonlinear-recursion},
\eqref{apdx_thm4:eq:radius-transfer}, and \eqref{apdx_thm4:eq:smallness} yield
\[
  d_{n+1}
  \leq a_n+B_nR_n^{1+\eta}
  \leq\left(1-\frac\alpha2\right)R_{n+1}
       +\frac\alpha2R_{n+1}
  =R_{n+1}.
\]
This closes the induction and explains the form of the radius sequence.

\textit{Step 3. Summation.}
For the sharper terminal estimate, the same definition also gives
\begin{equation}
  B_nR_n^\eta
  =\frac12
  \left(B_nK^\eta\alpha^{-n\eta}\right)^{1/(1+\eta)}
  \leq\frac12,
  \label{apdx_thm4:eq:radius-contraction}
\end{equation}
where the inequality uses $K\leq K_\alpha\leq\alpha^nB_n^{-1/\eta}$.
Since $d_n\leq R_n$, \eqref{apdx_thm4:eq:nonlinear-recursion} therefore implies
\begin{equation}
  d_{n+1}\leq a_n+\frac12d_n,
  \qquad 0\leq n<N.
  \label{apdx_thm4:eq:linear-recursion}
\end{equation}
Iterating \eqref{apdx_thm4:eq:linear-recursion} from $d_0=0$ gives
\begin{equation}
  d_N
  \leq\sum_{k=0}^{N-1}2^{-(N-1-k)}a_k
  \leq2\max_{0\leq k<N}a_k.
  \label{apdx_thm4:eq:dN-bound}
\end{equation}

\textit{Step 4. Prediction error.}
Applying \Cref{thm4:lem:global-transport} to $\ch^N$, $\nu_N$, and $\hat\nu_N$ yields
$e_N\leq\epsilon_N+K_\ch d_N$. The result now follows from
\eqref{apdx_thm4:eq:dN-bound} and the definition of $a_n$.
This completes the proof.
\end{proof}
\endgroup

\section{Hyperparameters}
\label{sec:hparams}
This appendix documents the full configuration used to produce the results in \Cref{sec:experiments}. All three benchmarks share the same training pipeline (\Cref{alg:wasp}) and differ only in the equation-specific parameters, the FNO width, and the transport schedule reported below. Unless a value is listed per equation, it is identical across Allen--Cahn (AC), Kuramoto--Sivashinsky (KS), and the complex Ginzburg--Landau equation (CGLE).

\subsection{Discretization and equation parameters}
\label{app:hparams-discretization}
Each    {input} $v=(u(0,x),f(t,x))$ and the corresponding solution are discretized on a uniform space--time grid of resolution $(n_t,n_x)$ over the domain $[0,T]_t\times[0,L]_x$. The temporal axis is zero-padded by $10\%$ to accommodate the non-periodic temporal boundary (\Cref{sec:experiments-setup}); this is the $[0.1,0]$ domain padding listed in \Cref{tab:hp-arch}. The CGLE field is complex, so it is stored in \texttt{complex64} and split into real and imaginary channels, whereas AC and KS use \texttt{float32}.

\begin{table}[h]
\centering
\caption{Discretization and equation parameters.}
\label{tab:hp-discretization}
\begin{tabular}{lccc}
\toprule
 & Allen--Cahn & KS & CGLE \\
\midrule
Temporal extent $T$            & $5$            & $0.128$        & $6.4$ \\
Spatial extent $L$             & $2\pi$         & $5$            & $50$ \\
Grid $(n_t,n_x)$               & $(128,128)$    & $(128,256)$    & $(128,256)$ \\
Data type                      & \texttt{float32} & \texttt{float32} & \texttt{complex64} \\
\midrule
Diffusion/Hyperviscosity $\nu$ & $0.01$         & $0.01$         & --- \\
$(\alpha,\beta)$               & ---            & ---            & $(2.0,-2.0)$ \\
IC amplitude                   & $10^{-3}$      & ---            & --- \\
IC spectral decay              & $1$            & ---            & --- \\
\midrule
Held-out test trajectories     & $1000$         & $10000$         & $10000$ \\
Spectral post-processing modes & ---            & $96$           & --- \\
\bottomrule
\end{tabular}
\end{table}

The spectral post-processing applies at the end of the forward pass: for KS, the predicted field is low-pass filtered to its leading $96$ spatial Fourier modes before any downstream computation, whereas AC and CGLE use no post-processing; this post-process is primarily for numerical stability since KS involves a $u_{xxxx}$ term that scales with $k^4$. The held-out test set consists of $1000$ ETD-RK4 solver snapshots for Allen--Cahn and $10000$ for KS and CGLE; the full evaluation protocol and its hyperparameters are described in \Cref{app:hparams-eval}.

\subsection{Neural operator architecture}
\label{app:hparams-arch}
We parametrize $\cg_\theta$ with a Fourier Neural Operator \citep{li2021fourier}. The input channels stack the initial condition, the forcing field, and a linear temporal encoding (five channels for the two-component CGLE field, three for the scalar AC and KS fields); the output channels carry the predicted solution (two for the complex CGLE field, one otherwise). All models use group normalization and a projection block whose width is twice the hidden width.

\begin{table}[h]
\centering
\caption{FNO architecture. Domain padding and normalization are shared.}
\label{tab:hp-arch}
\begin{tabular}{lccc}
\toprule
 & Allen--Cahn & KS & CGLE \\
\midrule
Input channels              & $3$        & $3$        & $5$ \\
Output channels             & $1$        & $1$        & $2$ \\
Fourier modes $(t,x)$       & $(64,128)$ & $(64,64)$  & $(64,64)$ \\
Hidden channels             & $32$       & $32$       & $64$ \\
Projection channel ratio    & $2$        & $2$        & $2$ \\
Domain padding $(t,x)$      & \multicolumn{3}{c}{$(0.1,\,0)$} \\
Normalization               & \multicolumn{3}{c}{group norm} \\
\bottomrule
\end{tabular}
\end{table}

The Allen-Cahn equation training requires higher number of spatial modes due to its phase-separation dynamics that causes sharp sptatial transition at large $t$.

\subsection{Optimization}
\label{app:hparams-optim}
All models are trained with AdamW \citep{loshchilov2018decoupled} under a StepLR schedule that multiplies the learning rate by $\gamma=0.9$ every fixed number of steps. Gradients are clipped to a per-equation maximum norm. The total step budget grows with the difficulty of the dynamics, from $2\times10^{4}$ steps for the relaxational AC equation to $2\times10^{5}$ for the chaotic CGLE.

\begin{table}[h]
\centering
\caption{Optimizer and learning-rate schedule. Optimizer, learning rate, weight decay, and momenta are shared across equations.}
\label{tab:hp-optim}
\begin{tabular}{lccc}
\toprule
 & Allen--Cahn & KS & CGLE \\
\midrule
Optimizer                  & \multicolumn{3}{c}{AdamW} \\
Learning rate              & \multicolumn{3}{c}{$4\times10^{-4}$} \\
Weight decay               & \multicolumn{3}{c}{$10^{-4}$} \\
$(\beta_1,\beta_2)$        & \multicolumn{3}{c}{$(0.9,\,0.999)$} \\
LR decay factor $\gamma$   & \multicolumn{3}{c}{$0.9$} \\
\midrule
Step-decay interval        & $2000$     & $10000$    & $20000$ \\
Gradient-norm clip         & $5000$     & $1000$     & $10000$ \\
Total optimizer steps      & $20000$    & $100000$   & $200000$ \\
\bottomrule
\end{tabular}
\end{table}

\subsection{Loss weights}
\label{app:hparams-loss}
The supervised objective \eqref{eq:loss_fn} combines a manufactured-data solution term (weight $\alpha_u$) with a input-consistency residual term (weight $\alpha_v$). Because the    {input} $v=(u(0,x),f(t,x))$ pairs an initial state with a forcing field, each term is further split into an interior component and an initial-condition component. The same values are used for all three equations; they are summarized in \Cref{tab:hp-loss}. They are also shared across the WaSP, MD, and MD+PINO configurations in \Cref{app:hparams-configs}; the exceptions are the PINO and PINO+OT, which raises the initial-condition residual weight from $0.1$ to $1.0$ so that the interior and initial-condition residuals are weighted equally; without the change, PINO fails to follow initial conditions and results in degenerate training.
 
\begin{table}[h]
\centering
\caption{Loss-term weights, common to all equations. The solution group realizes $\alpha_u$ in \eqref{eq:loss_fn} and the    {input} group realizes $\alpha_v$.}
\label{tab:hp-loss}
\begin{tabular}{llc}
\toprule
Group & Component & Weight \\
\midrule
Solution ($\alpha_u$)  & interior          & $10$ \\
                         & initial condition & $1.0$ \\
\midrule
   {Input} ($\alpha_v$) & interior residual         & $1.0$ \\
                         & initial-condition residual& $0.1$ \\
\bottomrule
\end{tabular}
\end{table}

\subsection{Replay buffer and dynamic normalization}
\label{app:hparams-buffer}
Self-generated supervision is stabilized with a replay buffer of synthesized pairs (\Cref{sec:implementation}). The buffer is initialized at full capacity and refilled at each step; each stored    {input} is reused for a fixed number of solution samples. Every optimizer step draws a batch composed of buffered (manufactured) pairs for the solution loss and freshly sampled    {inputs} for the residual loss; the two sub-batches sum to a total batch size of $64$ in every configuration (see \Cref{app:hparams-configs}). Input and output statistics are tracked online with an exponential moving average of momentum $10^{-3}$, which also sets the relative scaling between $\alpha_u$ and $\alpha_v$.

\begin{table}[h]
\centering
\caption{Replay-buffer, batch-size, and normalization settings. Generation chunk size, solution batch, evaluation batch, and normalizer momentum are shared.}
\label{tab:hp-buffer}
\begin{tabular}{lccc}
\toprule
 & Allen--Cahn & KS & CGLE \\
\midrule
Buffer capacity            & $5000$  & $20000$ & $20000$ \\
Initial buffer fill        & $5000$  & $20000$ & $20000$ \\
   {Inputs} reused per fill & $32$    & $16$    & $16$ \\
\midrule
Generation chunk size      & \multicolumn{3}{c}{$1000$} \\
Buffer (solution) batch    & \multicolumn{3}{c}{$32$} \\
Generated (PDE residual) batch & \multicolumn{3}{c}{$32$} \\
Evaluation batch           & \multicolumn{3}{c}{$64$} \\
Normalizer EMA momentum    & \multicolumn{3}{c}{$10^{-3}$} \\
\bottomrule
\end{tabular}
\end{table}

\subsection{Optimal transport and the interpolation schedule}
\label{app:hparams-transport}
For KS and CGLE the transport map $T$ from $\nu_0$ to $\nu_1$ is estimated with entropy-regularized Sinkhorn iterations \citep{cuturi2013sinkhorn, pooladian2021entropic} using a PCA-projected cost computation, fit on $5000$ samples. Allen--Cahn requires no transport map. Its manufactured initial condition distribution already matches the target, so no Sinkhorn problem is solved and the corresponding solver settings are absent from its configuration. Training then advances along the McCann interpolation $\nu_s$ \eqref{eq:mccann-path} starting from $s=0$. Rather than advancing $s$ on the fixed cosine base schedule, the adaptive rule increments $s$ by a fixed step only once the input-consistency residual falls below a threshold, so that harder portions of the path receive smaller effective Wasserstein steps (\Cref{sec:implementation}). $s$ is only advanced adaptively when it is in the range $[0.0, 0.9]$ and falls back to base cosine schedule otherwise. A short warm-up trains on the exact manufactured pair $(\mu_0,\nu_0)$ for the first $1\%$ of steps before transport begins.
 
\begin{table}[h]
\centering
\caption{Optimal-transport and interpolation-schedule settings. Allen--Cahn fits no transport map (its manufactured initial condition distribution already matches the target, the only transport is the zero map of forcing $f$), so the Sinkhorn solver settings do not apply; the schedule parameters are shared except where noted.}
\label{tab:hp-transport}
\begin{tabular}{lccc}
\toprule
 & Allen--Cahn & KS & CGLE \\
\midrule
\multicolumn{4}{l}{\emph{Sinkhorn transport map}}\\
Low-rank cost rank          & - & $64$      & $96$ \\
Fitting samples             & - & $5000$    & $5000$ \\
Entropic regularization     & - & $10^{-2}$ & $10^{-3}$ \\
Max iterations              & - & $1000$    & $1000$ \\
Stopping threshold          & - & $10^{-5}$ & $10^{-5}$ \\
\midrule
\multicolumn{4}{l}{\emph{Interpolation schedule}}\\
Base schedule               & \multicolumn{3}{c}{cosine} \\
Start time $s_0$            & \multicolumn{3}{c}{$0.0$} \\
Increment $\Delta s$        & $0.005$ & $0.002$   & $0.002$ \\
Advance threshold           & $0.2$   & $0.2$     & $0.3$ \\
Adaptive range of $s$       & \multicolumn{3}{c}{$[0.0,\,0.9]$} \\
Min.\ samples per step      & \multicolumn{3}{c}{$1000$} \\
Warm-up fraction            & \multicolumn{3}{c}{$0.01$} \\
\bottomrule
\end{tabular}
\end{table}

\subsection{Method and baseline configurations}
\label{app:hparams-configs}
WaSP and the two baselines and two ablations in \Cref{sec:experiments} share every setting above and are obtained by toggling the transport schedule and the two sub-batch sizes, so that all five optimize over the same total batch of $64$ and the same step budget. Manufactured Data (MD) keeps only the solution loss \eqref{eq:manufactured-loss} on the fixed synthetic distribution $\mu_0$; MD+PINO adds the residual term at the fixed initial pair $(\mu_0,\nu_0)$, which is exactly the WaSP objective \eqref{eq:loss_fn} frozen at $s=0$; and PINO keeps only the residual loss \eqref{eq:pino-residual-loss}, evaluated directly at the target distribution $\nu_1$ by fixing the transport time at $s=1$. The PINO and PINO+OT runs also raise the initial-condition residual weight from $0.1$ to $1.0$ so that interior and initial-condition residuals are weighted equally.

\begin{table}[h]
\centering
\caption{Configuration of WaSP, the two baselines, and the two ablations. Buffer and residual batches are the solution-loss and residual-loss sub-batches of \Cref{tab:hp-buffer}; they sum to $64$ in every case. PINO and PINO+OT additionally set the initial-condition residual weight to $1.0$ (equal to the interior residual weight).}
\label{tab:hp-configs}
\begin{tabular}{lcccl}
\toprule
Configuration & Transport & Buffer batch & Residual batch & Active objective \\
\midrule
WaSP (ours) & enabled  & $32$ & $32$ & \eqref{eq:loss_fn} along the path $\nu_s$ \\
PINO        & disabled & $0$  & $64$ & $\cl_{\mathrm{PDE}}$ at $\nu_1$ ($s=1$)\\
PINO+OT        & enabled & $0$  & $64$ & $\cl_{\mathrm{PDE}}$ at $\nu_s$ ($s\in[0, 1]$) \\
MD          & disabled & $64$ & $0$  & $\cl_{\mathrm{MD}}$ on fixed $\mu_0$ \\
MD+PINO     & disabled & $32$ & $32$ & \eqref{eq:loss_fn} frozen at $(\mu_0,\nu_0)$ \\
\bottomrule
\end{tabular}
\end{table}

\subsection{Evaluation protocol}
\label{app:hparams-eval}
All five configurations of each equation are evaluated from the final checkpoint ($2\times10^{4}$, $10^{5}$, and $2\times10^{5}$ steps for AC, KS, and CGLE) with the same protocol; the metrics correspond to those defined in \Cref{sec:evaluation}.
 
The relative solution error (metric~1) is a single-shot quantity: each held-out solver trajectory is mapped to its    {input} $v$ via $\cn$, the operator predicts $\cg_\theta(v)$ in one forward pass, and we report the aggregate relative error $\sum_i\|\cg_\theta(v_i)-u_i\|_2/\sum_i\|u_i\|_2$ over the test set, processed in batches of $64$. The same pass also logs the    {input} residual $\|\cn(\cg_\theta(v))-v\|_2$. This metric is the only one reported for the (non-chaotic) Allen--Cahn equation.
 
The four statistical metrics (metrics~2--5) require an attractor and are computed only for KS and CGLE, from autoregressive rollouts seeded by test initial conditions and compared against an independent ETD-RK4 solver rollout; the energy spectrum and the two autocorrelation errors are reported as relative RMSEs in the spectral and correlation domains. Three rollouts are collected, each burning in for $10000\,\delta t$ ($10$ time units for KS, $500$ for CGLE) before recording. A \emph{coarse} rollout, subsampled at physical interval $\Delta t$, supplies the invariant-measure distance, the energy spectrum, and the spatial autocorrelation (metrics~2, 5, 3). A \emph{fine} rollout at the native step $\delta t$ supplies the temporal autocorrelation (metric~4). The invariant-measure distance is the entropy-regularized Sinkhorn $W_2$ between the model and reference snapshot ensembles; the solver-only reference in \Cref{sec:evaluation} is obtained by running the same Sinkhorn solver between two independent solver rollouts that share the model's initial conditions. Trajectory and snapshot counts follow the defaults below, and the coarse-rollout interval $\Delta t$ and the Sinkhorn regularization are set per equation.
 
\begin{table}[h]
\centering
\caption{Evaluation-protocol settings. Trajectory and snapshot counts are shared across equations; every rollout burns in for $10000\,\delta t$ ($10$ time units for KS, $500$ for CGLE), and the coarse-rollout interval $\Delta t$ and Sinkhorn $\epsilon$ are set per equation. The four attractor-based metrics are disabled for Allen--Cahn.}
\label{tab:hp-eval}
\begin{tabular}{lccc}
\toprule
 & Allen--Cahn & KS & CGLE \\
\midrule
Solution $L^2$ (metric 1) & yes & yes & yes \\
Attractor metrics (2--5)     & no  & yes & yes \\
Test-set batch size          & \multicolumn{3}{c}{$64$} \\
\midrule
\multicolumn{4}{l}{\emph{Short-horizon eval (metric 1)}}\\
Snapshots                    & $1000$ & $10000$ & $10000$ \\
\midrule
\multicolumn{4}{l}{\emph{Coarse rollout (metrics 2, 3, 5)}}\\
Snapshots                    & --- & $10000$ & $10000$ \\
Parallel trajectories        & --- & $10$    & $10$ \\
Burn-in (time units)         & --- & $10$    & $500$ \\
Snapshot interval $\Delta t$ & --- & $0.5$   & $25$ \\
\midrule
\multicolumn{4}{l}{\emph{Fine rollout (metric 4)}}\\
Snapshots                    & --- & $20000$ & $20000$ \\
Parallel trajectories        & --- & $20$    & $20$ \\
Burn-in (time units)         & --- & $10$    & $500$ \\
Snapshot interval            & --- & native $\delta t$ & native $\delta t$ \\
\midrule
\multicolumn{4}{l}{\emph{Sinkhorn $W_2$ (metric 2)}}\\
Entropic regularization $\epsilon$ & --- & $10^{-2}$ & $10^{-3}$ \\
Max iterations                     & --- & $1000$    & $1000$ \\
Stopping threshold                 & --- & $10^{-5}$ & $10^{-5}$ \\
\bottomrule
\end{tabular}
\end{table}
 
Every rollout burns in for $10000\,\delta t$ before recording: $10$ time units for KS ($dt=T/n_t=10^{-3}$) and $500$ for CGLE ($dt=0.05$), corresponding to roughly $78$ autoregressive windows in each case.

\end{document}